\documentclass{article}
\usepackage{algorithm}
\usepackage{tikz}
\usetikzlibrary{arrows.meta, calc, decorations.pathreplacing, positioning, shapes.geometric}
\usepackage{subcaption}
\usepackage{amsmath,amssymb,amsfonts}
\usepackage{algorithm}
\usepackage{algpseudocode}
\usepackage{amsthm}
\usepackage{wrapfig}
\usepackage{multirow}
\usepackage{float}
\usepackage{booktabs}
\usepackage{graphicx} 
\usepackage[most]{tcolorbox}
\usepackage[table]{xcolor}
\usepackage{amsmath}
\usepackage{amsfonts}
\usepackage{subcaption}
\usepackage{pifont}
\newcommand{\pc}{\textbf{P}}

\newcommand{\cm}{\ding{51}} 
\newcommand{\xm}{\ding{55}} 
\usepackage{caption}

\usepackage[final]{corl_2026} 

\title{FedGuide: Diffusion Prior Alignment and Value Baseline Guidance for Heterogeneous Federated Reinforcement Learning}

\author{
  Zhilin He, Gauri Joshi\\
  Carnegie Mellon University \\
  \texttt{\{hectorh, gaurij\}@andrew.cmu.edu} \\
  Code: \href{https://github.com/hhhhzl/fedguide}{\texttt{https://github.com/hhhhzl/fedguide}}
}

\begin{document}
\maketitle
\vspace{-0.5cm}

 \newcommand{\GJ}[1]{{\leavevmode\color{magenta}[GJ: #1]}}

\begin{abstract}
Federated Reinforcement Learning (FRL) enables collaborative policy learning
across distributed agents with heterogeneous environments. While recent methods
based on variance reduction, divergence penalization, and momentum optimization
improve FRL under heterogeneous settings, they still primarily synchronize
policy or value-network parameters and do not explicitly address distributional
mismatch among heterogeneous clients. Therefore, we propose \textbf{FedGuide}, a FRL framework that uses diffusion priors as behavior models to provide personalized data supported distributions for heterogeneous
local policy learning. Instead of directly averaging local policies, FedGuide
aggregates those diffusion priors through Optimal-Transport Mixture-of-Experts
(OT-MoE), preserving heterogeneous behavior modes in distribution space. It
further develops a Distribution Correction Estimation (DICE) value baseline to provide low-variance, return-aware guidance for local policy improvement. Experiments across heterogeneous environments show that FedGuide outperforms representative FRL methods in client-average returns, final-round
performance, and worst-round robustness, while maintaining stable learning under
stronger heterogeneity.
\end{abstract}

\vspace{-0.3cm}
\section{Introduction}
\vspace{-0.3cm}
Robotic learning is increasingly moving toward distributed regimes in which data are collected across multiple robots, platforms, or deployment sites~\cite{wang2022distributed, halsted2021survey}. Federated Learning (FL) enables such systems to train collaboratively while sharing only model updates, but successful collaboration requires robustness to heterogeneous dynamics, environments, task instantiations, and data distributions~\cite{qi2021federated, yurdem2024federated, wang2024optimal}. The challenge, therefore, lies in learning across heterogeneous robotic clients, where mismatched local objectives and behavior distributions make naive model aggregation ineffective.
\vspace{-0.1cm}

Federated Reinforcement Learning (FRL) extends this collaborative paradigm to sequential decision-making by improving policies while keeping local trajectories decentralized~\cite{liu2019lifelong, liang2022federated, fan2021fault, woo2023blessing}. However, FRL faces harder data heterogeneity challenges because each client's data distribution is induced by its current policy interacting with a local Markov Decision Process (MDP), making both trajectories and update directions client-specific. Existing FRL methods mitigate this issue through policy regularization, distance constraint, or momentum optimization~\cite{xie2023fedkl, xiong2024personalized, wang2024momentum, yue2024momentum, jiang2025fedhpd}, but they fail to explicitly model or align the heterogeneous state-action distributions. As a result, aggregation combines policies whose underlying trajectory distributions remain mismatched across clients.

\vspace{-0.1cm}
This distribution level mismatch motivates a distribution alignment view of heterogeneous FRL. Diffusion models are a natural tool for this purpose because they explicitly model complex multimodal action or trajectory distributions and have been widely adopted in robotic policy learning, trajectory generation, and planning~\cite{chi2025diffusion, ren2025diffusion, he2026modelbaseddiffusionoptimalcontrol, he2026motionplanningmodelbaseddiffusion}. Meanwhile, Distribution Correction Estimation (DICE)~\cite{nachum2019dualdice, zhang2020gendice} provides a principled way to estimate offline value information for policy improvement~\cite{mao2024diffusion, jackson2024policy, ki2025prior}. However, existing diffusion and DICE-based methods typically assume a centralized and fixed behavior distribution, leaving open the federated case where heterogeneous clients induce distinct local trajectory distributions that must be aligned without sharing raw data.

\vspace{-0.1cm}
We propose \textbf{FedGuide}, a heterogeneous FRL framework for aligning client-specific behavior distributions without sharing raw trajectories. Instead of directly aggregating local policies, FedGuide uses diffusion behavior priors as the federated communication interface, allowing heterogeneous behavior modes to be shared in distribution space while policy parameters remain local. The resulting personalized priors regularize local reinforcement learning, and a state-only DICE value baseline further stabilizes return-aware policy improvement.

The contributions of this work are twofold: \textbf{C1:} We formulate heterogeneous FRL as local policy learning coupled by distribution-level diffusion behavior priors, and develop an OT-MoE aggregation mechanism that routes client prior heads to server experts and broadcasts personalized distribution-mixture priors without directly averaging local policies; \textbf{C2:} We introduce a prior-regularized local Proximal Policy Optimization (PPO)~\cite{schulman2017proximal} objective with a state-only DICE value baseline, and prove score-function unbiasedness, variance reduction under positive residual correlation, and convergence to an approximate stationary point of the federated surrogate with an explicit residual error floor.

\vspace{-0.3cm}
\section{Related Work}
\vspace{-0.3cm}
\paragraph{Heterogeneous Federated Reinforcement Learning.}
In FL, client heterogeneity has been addressed through dynamic parameter sharing, robust aggregation, and personalization mechanisms that preserve local specificity while enabling global collaboration~\cite{shen2023feddm, yang2023fedrep, tan2022towards, wang2020federated}. Recent FRL works extend this idea to sequential decision-making, but heterogeneity becomes policy-induced: local trajectories are generated under different MDPs, motivating methods based on knowledge transfer~\cite{liu2019lifelong, liang2022federated}, policy-divergence regularization~\cite{xie2023fedkl}, and stabilized policy-gradient aggregation~\cite{jin2022federated}. Building on these foundations, recent works move toward provable heterogeneous FRL, including momentum-based methods for interaction and communication efficient learning~\cite{wang2024momentum, yue2024momentum}, finite-time analysis of on-policy heterogeneous FRL~\cite{zhang2024finite}, and linear-speedup guarantees for federated Q-learning under heterogeneity~\cite{khodadadian2022federated, woo2024federated, woo2023blessing, woo2025blessing}. However, these advances resolve heterogeneity through policy-update optimization or shared policy aggregation, so heterogeneous behavior modes can still be collapsed when policy-induced state-action distributions are mismatched. Distribution-aware federated methods aggregate structured distributions rather than parameters, including personalized Mixture-of-Experts (MoE) models that share knowledge through client-specific experts~\cite{mei2024fedmoe, fang2026hfedmoe, yi2026pfedmoe, pereira2025heterogeneous} and Optimal-Transport (OT)-based aggregation~\cite{agueh2011barycenters, cuturi2013sinkhorn}. Nevertheless, these methods are designed for static supervised distributions rather than policy induced trajectory distributions.

\vspace{-0.3cm}
\paragraph{Diffusion Priors and Value-Guided Policy Learning.}
Diffusion models have been extensively studied in reinforcement learning as behavior models, where diffusion policies learn multi-modal action and trajectory distributions from demonstrations~\cite{chi2025diffusion, ren2025diffusion}. However, pure behavior modeling only imitates the data distribution and does not directly perform policy improvement. DICE-based methods use stationary distribution ratios to transform one state-action distribution into another in RL~\cite{mao2024odice, nachum2019algaedice, sikchi2024dual, lee2022coptidice}, and more recent works extend this view to guide diffusion policies toward high-value in-sample actions
~\cite{mao2024diffusion, jackson2024policy, ki2025prior}. However, these methods assume a centralized behavior distribution from a single dataset and do not address how multiple heterogeneous behavior priors should be communicated, routed, or personalized across clients. Since diffusion priors, OT, and DICE each already exist, our contribution is not a new component but the federated communication abstraction that combines them: behavior distributions, rather than policy parameters or gradients, become the exchanged object. OT-MoE then decides which represented behavior support is transferable to which client, while the DICE value function stays local as a state-only baseline and never guides diffusion sampling.


\vspace{-0.3cm}
\section{Preliminaries}
\vspace{-0.3cm}
In this section, we cover the relevant background. We use a left superscript $k$ for diffusion steps, a right superscript $t$ for environment time, and subscript $i$ for clients (e.g ${}^{k}a_i^{t}$ denotes client $i$'s action at time $t$ and $k$ diffusion step). We use $h=1,\ldots,H$ for communication rounds, $T$ for a trajectory horizon, $U$ for local updates per round, $N$ for clients, and $M$ for server side diffusion prior experts. Policy, prior, and DICE value parameters are denoted by $\theta_i$, $\psi_i$, and $\phi_i$, respectively.
\vspace{-0.2cm}
\label{sec:preliminaries}
\subsection{Problem Formulation}
\vspace{-0.2cm}
We consider a FRL system with $N$ heterogeneous
clients. Client $i \in \{1,\dots,N\}$ interacts with its MDP
\begin{math}
\mathcal{M}_i \;=\; \langle \mathcal{S},\,\mathcal{A},\,P_i,\,r_i,\,\rho_i,\,\gamma\rangle ,
\end{math}
where $\mathcal{S}$ and $\mathcal{A}$ are the shared state and action spaces, with $s \in \mathcal{S}$ and $a \in \mathcal{A}$, while the transition function $P_i$, reward $r_i$, and initial state distribution $\rho_i$ differ across clients. A trajectory of horizon $T$ generated by local policy $\pi_i$
follows
{%
\setlength{\abovedisplayskip}{2.5pt}
\setlength{\belowdisplayskip}{2.5pt}
\setlength{\abovedisplayshortskip}{1.5pt}
\setlength{\belowdisplayshortskip}{1.5pt}
\begin{equation}
p_i(\tau_i\mid\pi_i) \;=\; \rho_i(s_i^0)\prod_{t=0}^{T-1}\pi_i(a_i^t\mid s_i^t)\,P_i(s_i^{t+1}\mid s_i^t,a_i^t).
\end{equation}
}
The federated objective seeks a system of policies that maximizes the average
expected return
{%
\setlength{\abovedisplayskip}{2.5pt}
\setlength{\belowdisplayskip}{2.5pt}
\setlength{\abovedisplayshortskip}{1.5pt}
\setlength{\belowdisplayshortskip}{1.5pt}
\begin{equation}
\label{eq:frl-objective}
\max_{\{\pi_i\}_{i=1}^{N}}\; J(\{\pi_i\})\;=\;\frac{1}{N}\sum_{i=1}^{N} J_i(\pi_i),
\quad
J_i(\pi_i)\;=\;\mathbb{E}_{\tau\sim p_i(\tau_i\mid\pi_i)}\!\!\left[\sum_{t=0}^{T-1}\gamma r_i(s_i^t,a_i^t)\right].
\end{equation}
}with a discount factor $\gamma$. The heterogeneity of $\{P_i,r_i,\rho_i\}$ produces non-IID local data and inconsistent gradients. FedGuide therefore keeps policies local and couples clients through personalized distribution diffusion priors rather than a shared policy.

\vspace{-0.3cm}
\subsection{Diffusion Priors and DICE Value Functions}
\label{sec:prelim:dice-diffusion}
\vspace{-0.3cm}

\paragraph{Diffusion prior as behavior model.}
For each client $i$, given an offline dataset $\mathcal{D}_i$ of state--action
pairs, we train a state-conditioned diffusion model that defines a forward
noising process
{%
\setlength{\abovedisplayskip}{2.5pt}
\setlength{\belowdisplayskip}{2.5pt}
\setlength{\abovedisplayshortskip}{1.5pt}
\setlength{\belowdisplayshortskip}{1.5pt}
\begin{equation}
q_k({}^{k}a^{t}\mid {}^{0}a^{t},s^{t})
\;=\;\mathcal{N}\!\bigl({}^{k}a^{t};\,\sqrt{\bar{\alpha}_k}\,{}^{0}a^{t},\,(1-\bar{\alpha}_k)\mathbf{I}\bigr),
\quad k=0,\ldots,K,
\end{equation}
}where ${}^{0}a^{t}$ denotes the clean action at environment time step $t$,
${}^{k}a^{t}$ denotes its noisy version at diffusion step $k$, and
$\bar{\alpha}_k$ is the cumulative noise-schedule coefficient. The forward
process gradually corrupts data actions into Gaussian noise, while the reverse
process starts from ${}^{K}a^{t}\sim\mathcal{N}(0,\mathbf I)$ and iteratively
denoises it back to a data-supported action conditioned on $s^{t}$. We train
a score network $\mathcal{S}_{\psi,i}(s^{t},{}^{k}a^{t},k)$ to approximate the marginal
score $\nabla_{{}^{k}a^{t}}\log q_k({}^{k}a^{t}\mid s^{t})$, the gradient of the log-density of the noisy action distribution at diffusion step $k$, which gives the denoising direction for moving ${}^{k}a^{t}$ toward data-supported actions, through denoising score matching. We denote the learned behavior policy induced by this reverse diffusion process as
$\pi_{D,i}(\cdot\mid s^{t})$, which is a conditional probability density over clean
actions under state $s^{t}$:
\begin{math}
\pi_{D,i}(a^{t}\mid s^{t}) \;\equiv\; p_{\psi,i}({}^{0}a^{t}=a^{t}\mid s^{t}),
\end{math}
together with its score $\nabla_{a^{t}}\log \pi_{D,i}(a^{t}\mid s^{t})$. Crucially, $\pi_{D,i}$ is a behavior model: it captures which actions occur in $\mathcal{D}_i$, regardless of return, providing data-supported actions. 

\vspace{-0.3cm}
\paragraph{DICE value function.} While the diffusion prior provides a data-supported behavior distribution over actions, it does not by itself provide return information. Therefore, a value signal is needed for return-aware policy improvement within the behavior support. For a policy $\pi$ on client $i$, let $d_i^{\pi}(s,a)=(1-\gamma)\sum_{t\geq 0}\gamma^{t}\Pr(s_i^{t}{=}s,a_i^{t}{=}a\mid\pi,\mathcal{M}_i)$ denote its discounted state--action occupancy, and let $d_i^{\mathcal{D}_i}$ denote the empirical state--action distribution of the offline buffer $\mathcal{D}_i$. DICE methods~\cite{nachum2019dualdice} estimate the
distribution ratio
{%
\setlength{\abovedisplayskip}{2.5pt}
\setlength{\belowdisplayskip}{2.5pt}
\setlength{\abovedisplayshortskip}{1.5pt}
\setlength{\belowdisplayshortskip}{1.5pt}
\begin{equation}
\label{eq:dice-ratio}
w_i(s^{t},a^{t})\;=\;\frac{d_i^{\pi_i^{\star}}(s^{t},a^{t})}{d_i^{\mathcal{D}_i}(s^{t},a^{t})}
\end{equation}
}between an implicit improved local policy $\pi_i^{\star}$ and the data distribution $\mathcal{D}_i$, measuring how strongly an offline state--action pair should be reweighted toward the improved policy distribution, with a larger value $w_i$ indicating greater emphasis for policy improvement. Let $\phi$ denote the parameters of the DICE value estimators. In the
DualDICE~\cite{nachum2019algaedice} or OptiDICE~\cite{lee2021optidice} formulation this ratio is parameterized by value functions
$V_{\phi,i}:\mathcal{S}\to\mathbb{R}$ and $Q_{\phi,i}:\mathcal{S}\times\mathcal{A}\to\mathbb{R}$
satisfying
{%
\setlength{\abovedisplayskip}{2.5pt}
\setlength{\belowdisplayskip}{2.5pt}
\setlength{\abovedisplayshortskip}{1.5pt}
\setlength{\belowdisplayshortskip}{1.5pt}
\begin{equation}
w_i(s^{t},a^{t})\;\approx\;\exp\!\Big(\tfrac{1}{\omega}\bigl(Q_{\phi,i}(s^{t},a^{t})-V_{\phi,i}(s^{t})\bigr)\Big),
\end{equation}
}motivated by DualDICE objectives such as the following. Here, $f$ is the divergence generator used to regularize the distribution ratio, and $f^{*}$ is its corresponding dual function in the DICE objective
{%
\setlength{\abovedisplayskip}{2.5pt}
\setlength{\belowdisplayskip}{2.5pt}
\setlength{\abovedisplayshortskip}{1.5pt}
\setlength{\belowdisplayshortskip}{1.5pt}
\begin{equation}
\begin{aligned}
    \min_{V_{\phi,i}}\;
    \mathbb{E}_{(s^{t},a^{t})\sim d_i^{\mathcal{D}_i}}\!
    \Big[(1-\gamma)\,V_{\phi,i}(s^{t})
       + \omega\, f^{*}\!\!\Big(\tfrac{1}{\omega}\bigl(Q_{\phi,i}(s^{t},a^{t})-V_{\phi,i}(s^{t})\bigr)\Big)\Big],\\
    \min_{Q_{\phi,i}}\;
    \mathbb{E}_{(s^{t},a^{t},s^{t+1})\sim d_i^{\mathcal{D}_i}}\!\Big[(r_i(s^{t},a^{t})+\gamma V_{\phi,i}(s^{t+1})-Q_{\phi,i}(s^{t},a^{t}))^{2}\Big],
\label{eq:dice-Q}
\end{aligned}
\end{equation}
}which motivate a DICE-style offline estimator of the ratio $w_i(s^{t},a^{t})$ through $Q_{\phi,i}$ and $V_{\phi,i}$, and $\omega>0$ is the DICE regularization temperature. After pretraining, $V_{\phi,i}$ is a state-only value estimate trained on stationary offline data. Since it does not score actions, it avoids directly evaluating potentially out-of-distribution actions sampled by an evolving online policy~\cite{kumar2020conservative, mao2024diffusion}; since it is fixed before online improvement, it provides a stable baseline signal under small per-round rollouts~\cite{nachum2019dualdice,lee2021optidice}. We therefore use $V_{\phi,i}$ only to center advantage estimates for return-aware local policy improvement, not to reshape $\pi_{D,i}$ or enter as a direct policy-gradient term, which is the state-only baseline property used in our variance analysis.

\begin{figure}[tb!]
\centering
\begin{subfigure}[b]{0.30\textwidth}
\centering
\begin{tikzpicture}[>=latex, thick, scale=0.85]
    \definecolor{myred}{RGB}{180,50,50}
    \definecolor{myblue}{RGB}{50,90,200}
    \definecolor{mygreen}{RGB}{60,160,60}
    \definecolor{myblack}{RGB}{30,30,30}
    \definecolor{myyellow}{RGB}{200,160,0}
    \definecolor{myorange}{RGB}{230,110,40}
    \definecolor{mygray}{RGB}{140,140,140}

    \coordinate (x0) at (0,0);
    \coordinate (x1) at (2.5, -2);     
    \coordinate (x2) at (-1.2,-2.0);     
    \coordinate (x3) at (0.75, -2);     
    \coordinate (x4) at (0.7, -3);       

    \draw[dashed, myblue, line width=1pt] (x0) -- (x1);
    \draw[dashed, myblue, line width=1pt] (x0) -- (x2);
    \draw[dashed, myred, line width=1pt] (x1) -- (x2);

    \draw[->, myred, line width=2pt] (x0) -- (x3);
    \draw[->, myorange, line width=1.5pt] (x4) -- (x3)
        node[midway, right, text=myorange] {\small Avg Error};

    \draw[->, myblack, line width=1pt] (x0) -- (-0.55,-0.4);
    \draw[->, myblack, line width=1pt] (-0.55,-0.4) -- (-0.3,-0.95);
    \draw[->, myblack, line width=1pt] (-0.3,-0.95) -- (-1.05,-1.35);
    \draw[->, myblack, line width=1pt] (-1.05,-1.35) -- (x2);

    \draw[->, myblack, line width=1pt] (x0) -- (1.10,-0.30);
    \draw[->, myblack, line width=1pt] (1.10,-0.30) -- (1.30,-1.10);
    \draw[->, myblack, line width=1pt] (1.30,-1.10) -- (2.45,-1.40);
    \draw[->, myblack, line width=1pt] (2.45,-1.40) -- (x1);

    \fill (x0) circle (3pt) node[above left] {$\mathbf{x}^{(t,0)}$};
    \fill (x3) circle (3pt) node[above right=-1pt and -1pt] {$\mathbf{x}^{(t,H)}$};
    \fill[myblue] (x2) circle (3pt) node[myblack, left] {$\mathbf{x}_1^{*}$};
    \fill[myblue] (x1) circle (3pt) node[myblack, right] {$\mathbf{x}_2^{*}$};

    \node[rectangle, draw=mygreen, fill=mygreen,
          minimum size=5pt, inner sep=0pt]
          (xstar) at (0.7,-3) {};
    \node[below=1pt of xstar] {$\mathbf{x}^*$};
\end{tikzpicture}
\caption{Policy-based Aggregation}
\end{subfigure}
\hfill
\begin{subfigure}[b]{0.30\textwidth}
\centering
\begin{tikzpicture}[>=latex, thick, scale=0.85]
    \definecolor{myred}{RGB}{180,50,50}
    \definecolor{myblue}{RGB}{50,90,200}
    \definecolor{mygreen}{RGB}{60,160,60}
    \definecolor{myblack}{RGB}{30,30,30}
    \definecolor{myyellow}{RGB}{200,160,0}
    \definecolor{myorange}{RGB}{230,110,40}
    \definecolor{mypurple}{RGB}{125,80,180}

    \coordinate (x0) at (0,0);
    \coordinate (x1) at (2.5, -2);     
    \coordinate (x2) at (-1.2, -2.0);    
    \coordinate (x4) at (0.7, -3);       
    \coordinate (x5) at (0.65, -2.3);   

    \coordinate (x6) at (-0.6, -2.5);   
    \coordinate (x7) at (1.9, -2.5);   

    \coordinate (x8) at (0.05, -2.8);    
    \coordinate (x9) at (1.3, -2.9);     
    \coordinate (x10) at (0.65, -2.8); 

    \draw[dashed, myblue, line width=1pt] (x0) -- (x1);
    \draw[dashed, myblue, line width=1pt] (x0) -- (x2);

    \draw[->, myblack, line width=1pt] (x0) -- (-0.55,-0.4);
    \draw[->, myblack, line width=1pt] (-0.55,-0.4) -- (-0.3,-0.95);
    \draw[->, myblack, line width=1pt] (-0.3,-0.95) -- (-1.05,-1.35);
    \draw[->, myblack, line width=1pt] (-1.05,-1.35) -- (x2);
    \draw[->, myblue, line width=2pt] (x0) -- (x6);

    \draw[->, myblack, line width=1pt] (x0) -- (1.10,-0.30);
    \draw[->, myblack, line width=1pt] (1.10,-0.30) -- (1.30,-1.10);
    \draw[->, myblack, line width=1pt] (1.30,-1.10) -- (2.45,-1.40);
    \draw[->, myblack, line width=1pt] (2.45,-1.40) -- (x1);
    \draw[->, myblue, line width=2pt] (x0) -- (x7);

    \begin{scope}
        \fill[myred, opacity=0.07] (x10) ellipse (1.05 and 0.42);
        \draw[myred, opacity=0.35, line width=0.9pt] (x10) ellipse (1.05 and 0.42);
        \node[text=myred, opacity=0.9] at ($(x10)+(-0.02,0.2)$) {\scriptsize $\bar{\pi}_{D}$};

        \fill[myred, opacity=0.14] (x8) ellipse (0.33 and 0.22);
        \draw[myred, opacity=0.55, line width=0.8pt] (x8) ellipse (0.33 and 0.22);
        \fill[myred, opacity=0.18] ($(x8)+(-0.10,0.05)$) circle (0.07);
        \fill[myred, opacity=0.18] ($(x8)+(0.08,-0.04)$) circle (0.08);
        \node[text=myred] at ($(x8)+(-0.35,-0.4)$) {\scriptsize $\bar{\pi}_{D,1}$};

        \fill[myred, opacity=0.14] (x9) ellipse (0.33 and 0.22);
        \draw[myred, opacity=0.55, line width=0.8pt] (x9) ellipse (0.33 and 0.22);
        \fill[myred, opacity=0.18] ($(x9)+(-0.08,0.04)$) circle (0.07);
        \fill[myred, opacity=0.18] ($(x9)+(0.10,-0.04)$) circle (0.08);
        \node[text=myred] at ($(x9)+(0.30,-0.4)$) {\scriptsize $\bar{\pi}_{D,2}$};
    \end{scope}

    \draw[->, myyellow, line width=1.4pt] (x2) -- (x6);
    \draw[->, myyellow, line width=1.4pt] (x1) -- (x7);

    \draw[myblue, dashed, opacity=0.35, line width=1pt] (x6) ellipse (0.42 and 0.28);
    \draw[myblue, dashed, opacity=0.35, line width=1pt] (x7) ellipse (0.42 and 0.28);

    \fill (x0) circle (3pt) node[above left] {$\mathbf{x}^{(t,0)}$};
    
    
    \fill[myblue] (x6) circle (3pt) node[below left=-1pt and -10pt] {$\mathbf{x}_1^{(t,H)}$};
    \fill[myblue] (x7) circle (3pt) node[below right=0pt and -8pt] {$\mathbf{x}_2^{(t,H)}$};
    \fill[myblue] (x2) circle (3pt) node[myblack, left] {$\mathbf{x}_1^{*}$};
    \fill[myblue] (x1) circle (3pt) node[myblack, right] {$\mathbf{x}_2^{*}$};

    \node[rectangle, draw=mygreen, fill=mygreen,
          minimum size=5pt, inner sep=0pt]
          (xstar) at (0.7,-3) {};
    \node[below=1pt of xstar] {$\mathbf{x}^*$};
\end{tikzpicture}
\caption{Priors Only (FedGuide-P)}
\end{subfigure}
\hfill
\begin{subfigure}[b]{0.38\textwidth}
\centering
\begin{tikzpicture}[>=latex, thick, scale=0.85]
    \definecolor{myred}{RGB}{180,50,50}
    \definecolor{myblue}{RGB}{50,90,200}
    \definecolor{mygreen}{RGB}{60,160,60}
    \definecolor{myblack}{RGB}{30,30,30}
    \definecolor{myyellow}{RGB}{200,160,0}
    \definecolor{myorange}{RGB}{230,110,40}
    \definecolor{mypurple}{RGB}{125,80,180}

    \coordinate (x0) at (0,0);
    \coordinate (x1) at (2.5, -2);     
    \coordinate (x2) at (-1.2, -2);    
    \coordinate (x4) at (0.7, -3);       
    \coordinate (x5) at (0.65, -2.3);    

    \coordinate (x6) at (-0.6, -2.5);   
    \coordinate (x7) at (1.9, -2.5);   

    \coordinate (x8) at (0.05, -2.8);    
    \coordinate (x9) at (1.3, -2.9);     
    \coordinate (x10) at (0.65, -2.8); 

    \draw[dashed, myblue, line width=1pt] (x0) -- (x1);
    \draw[dashed, myblue, line width=1pt] (x0) -- (x2);
    
    \draw[->, myblack, opacity=0.22, line width=0.9pt] (x0) -- (-0.55,-0.4);
    \draw[->, myblack, opacity=0.22, line width=0.9pt] (-0.55,-0.4) -- (-0.3,-0.95);
    \draw[->, myblack, opacity=0.22, line width=0.9pt] (-0.3,-0.95) -- (-1.05,-1.35);
    \draw[->, myblack, opacity=0.22, line width=0.9pt] (-1.05,-1.35) -- (x2);
    
    \draw[->, myblack, opacity=0.22, line width=0.9pt] (x0) -- (1.10,-0.30);
    \draw[->, myblack, opacity=0.22, line width=0.9pt] (1.10,-0.30) -- (1.30,-1.10);
    \draw[->, myblack, opacity=0.22, line width=0.9pt] (1.30,-1.10) -- (2.45,-1.40);
    \draw[->, myblack, opacity=0.22, line width=0.9pt] (2.45,-1.40) -- (x1);
    
    \draw[->, mygreen, line width=1.5pt] (-0.55,-0.4) -- (-0.30,-0.55);
    \draw[->, mygreen, line width=1.5pt] (-0.3,-0.95) -- (-0.48,-1.05);
    \draw[->, mygreen, line width=1.5pt] (-1.05,-1.35) -- (-0.72,-1.52)
        node[midway, left, text=mygreen] {\scriptsize $V_{\phi,i}$};
    
    \draw[->, mygreen, line width=1.5pt] (1.10,-0.30) -- (0.62,-0.58);
    \draw[->, mygreen, line width=1.5pt] (1.30,-1.10) -- (1.15,-1.10);
    \draw[->, mygreen, line width=1.5pt] (2.45,-1.40) -- (1.88,-1.68)
        node[midway, right, text=mygreen] {\scriptsize $V_{\phi,i}$};
    
    \draw[->, myblack, line width=1pt] (x0) -- (-0.30,-0.55);
    \draw[->, myblack, line width=1pt] (-0.30,-0.55) -- (-0.48,-1.05);
    \draw[->, myblack, line width=1pt] (-0.48,-1.05) -- (-0.72,-1.52);
    \draw[->, myblack, line width=1pt] (-0.72,-1.52) -- (x2);
    
    \draw[->, myblack, line width=1pt] (x0) -- (0.62,-0.58);
    \draw[->, myblack, line width=1pt] (0.62,-0.58) -- (1.15,-1.10);
    \draw[->, myblack, line width=1pt] (1.15,-1.10) -- (1.88,-1.68);
    \draw[->, myblack, line width=1pt] (1.88,-1.68) -- (x1);
    
    \draw[->, myblue, line width=2pt] (x0) -- (x6);
    \draw[->, myblue, line width=2pt] (x0) -- (x7);

    \begin{scope}
        \fill[myred, opacity=0.07] (x10) ellipse (1.05 and 0.42);
        \draw[myred, opacity=0.35, line width=0.9pt] (x10) ellipse (1.05 and 0.42);
        \node[text=myred, opacity=0.9] at ($(x10)+(-0.02,0.2)$) {\scriptsize $\bar{\pi}_{D}$};

        \fill[myred, opacity=0.14] (x8) ellipse (0.33 and 0.22);
        \draw[myred, opacity=0.55, line width=0.8pt] (x8) ellipse (0.33 and 0.22);
        \fill[myred, opacity=0.18] ($(x8)+(-0.10,0.05)$) circle (0.07);
        \fill[myred, opacity=0.18] ($(x8)+(0.08,-0.04)$) circle (0.08);
        \node[text=myred] at ($(x8)+(-0.35,-0.4)$) {\scriptsize $\bar{\pi}_{D,1}$};

        \fill[myred, opacity=0.14] (x9) ellipse (0.33 and 0.22);
        \draw[myred, opacity=0.55, line width=0.8pt] (x9) ellipse (0.33 and 0.22);
        \fill[myred, opacity=0.18] ($(x9)+(-0.08,0.04)$) circle (0.07);
        \fill[myred, opacity=0.18] ($(x9)+(0.10,-0.04)$) circle (0.08);
        \node[text=myred] at ($(x9)+(0.30,-0.4)$) {\scriptsize $\bar{\pi}_{D,2}$};
    \end{scope}

    \draw[->, myyellow, line width=1.4pt] (x2) -- (x6);
    \draw[->, myyellow, line width=1.4pt] (x1) -- (x7);

    \draw[mygreen, opacity=0.9, line width=1.2pt] (x6) ellipse (0.20 and 0.13);
    \draw[mygreen, opacity=0.9, line width=1.2pt] (x7) ellipse (0.20 and 0.13);

    \node[text=mygreen] at ($(x6)+(0.55,0.32)$) {\scriptsize $\mathrm{Var}\downarrow$};
    \node[text=mygreen] at ($(x7)+(0.6,0.13)$) {\scriptsize $\mathrm{Var}\downarrow$};

    \fill (x0) circle (3pt) node[above left] {$\mathbf{x}^{(t,0)}$};

    \fill[myblue] (x6) circle (3pt)
        node[below left=-1pt and -10pt] {$\mathbf{x}_1^{(t,H)}$};
    \fill[myblue] (x7) circle (3pt)
        node[below right=0pt and -8pt] {$\mathbf{x}_2^{(t,H)}$};

    \fill[myblue] (x2) circle (3pt) node[myblack, left] {$\mathbf{x}_1^{*}$};
    \fill[myblue] (x1) circle (3pt) node[myblack, right] {$\mathbf{x}_2^{*}$};

    \node[rectangle, draw=mygreen, fill=mygreen,
          minimum size=5pt, inner sep=0pt]
          (xstar) at (0.7,-3) {};
    \node[below=1pt of xstar] {$\mathbf{x}^*$};
    \end{tikzpicture}
    \caption{Priors and Value Baseline (FedGuide)}
    \end{subfigure}
    \caption{
        \textbf{Method Visualization with 2 Clients Heterogeneous Settings.}
        (a) Policy aggregation averages local optima $\mathbf{x}_1^{*}, \mathbf{x}_2^{*}$ into global policy $\mathbf{x}^{(t,H)}$, away from the global optimum $\mathbf{x}^{*}$. (b) OT-MoE aggregates diffusion priors in distribution space, forming shared prior support $\bar{\pi}_{D}$ and personalized priors $\bar{\pi}_{D,1},\bar{\pi}_{D,2}$ (pink). The prior regularizer $\widehat{\mathcal R}$ (yellow) pulls local policies toward personalized prior supports. (c) FedGuide further uses the DICE value baseline (green) for policy improvement, shown as straighter policies updates and smaller endpoint dispersion ($\mathrm{Variance}\downarrow$).
    }
    \vspace{-0.5cm}
\end{figure}

\vspace{-0.2cm}
\section{Methodology}
\label{sec:methods}
\vspace{-0.2cm}
We now present our method \textbf{FedGuide}. Given each client's the diffusion prior $\pi_{D,i}$ and value function $V_{\phi,i}$, \textbf{FedGuide} \textbf{(1)} aggregates those diffusion priors through OT-MoE to align each local policy with a personalized distribution mixture prior $\bar{\pi}_{D,i}$, mitigating the distribution mismatch caused by federated heterogeneity, and \textbf{(2)} introduces a state-only DICE value baseline from $V_{\phi,i}$ to provide low-variance local policy improvement guidance.

\vspace{-0.3cm}
\subsection{OT-MoE Aggregation of the Diffusion Priors}
\label{sec:methods:otmoe}
\vspace{-0.2cm}

The server maintains $M$ diffusion prior experts
$\{\pi_{D,\psi^{(m)}}\}_{m=1}^{M}$, where $M$ denotes the number of global
distributional modes rather than the number of clients. Since the diffusion
prior is a distribution over actions, directly averaging heterogeneous priors can
collapse multimodal behaviors or place probability mass between client modes,
leading to unsupported actions. After local round $h$, each client uploads its
locally adapted prior head $\psi_i^{(h)}$. Heterogeneous clients should not be
forced into a single mode, so we use an OT-based plan to
soft-route clients to experts. Let $C^{(h)}\in\mathbb{R}_{+}^{N\times M}$ be a
cost matrix with
{%
\setlength{\abovedisplayskip}{2.5pt}
\setlength{\belowdisplayskip}{2.5pt}
\setlength{\abovedisplayshortskip}{1.5pt}
\setlength{\belowdisplayshortskip}{1.5pt}
\begin{equation}
\label{eq:ot-cost}
C^{(h)}_{i,m} \;=\; \mathbb{E}_{(s^{t},a^{t})\sim d_i^{\mathcal{D}_i}}\!\Bigl[
\bigl\|\nabla_{a^{t}}\log \pi_{D,\psi_i^{(h)}}(a^{t}\mid s^{t})
       -\nabla_{a^{t}}\log \pi_{D,\psi^{(m,h)}}(a^{t}\mid s^{t})\bigr\|_2^{2}\Bigr],
\end{equation}
}a score-MSE distance between client $i$'s prior and that of expert $m$. In implementation, we use a parameter-space proxy for this score distance; the analysis treats this proxy mismatch as a bounded cost-approximation error $\epsilon_C$. Let $Y\in\mathbb{R}_{+}^{N\times M}$ denote the OT-based plan, where
$Y_{i,m}$ is the soft assignment mass from client $i$ to expert $m$. The feasible
set is
$\Pi(\mu,\nu)=\{Y\ge 0 \mid \sum_{m=1}^{M}Y_{i,m}=\mu_i,\;
\sum_{i=1}^{N}Y_{i,m}=\nu_m\}$, and the
server solves a Sinkhorn-regularized OT problem~\cite{cuturi2013sinkhorn, sinkhorn1967concerning}
{%
\setlength{\abovedisplayskip}{2.5pt}
\setlength{\belowdisplayskip}{2.5pt}
\setlength{\abovedisplayshortskip}{1.5pt}
\setlength{\belowdisplayshortskip}{1.5pt}
\begin{equation}
\label{eq:ot-plan}
Y^{\star}\;=\;\arg\min_{Y\in\Pi(\mu,\nu)}\;
\sum_{i=1}^{N}\sum_{m=1}^{M}Y_{i,m}C^{(h)}_{i,m}
\;+\;\eta\sum_{i=1}^{N}\sum_{m=1}^{M}Y_{i,m}\log Y_{i,m},
\end{equation}
}where $\eta>0$ is the regularization coefficient, with
marginals $\mu_i=1/N$, $\nu_m=1/M$. Each expert head is updated as a
column-normalized convex combination of client prior heads,
{%
\setlength{\abovedisplayskip}{2.5pt}
\setlength{\belowdisplayskip}{2.5pt}
\setlength{\abovedisplayshortskip}{1.5pt}
\setlength{\belowdisplayshortskip}{1.5pt}
\begin{equation}
\label{eq:expert-update}
\psi^{(m,h+1)}\;\leftarrow\;
\sum_{i=1}^{N}\frac{Y_{i,m}^{\star}}{\sum_{j=1}^{N}Y_{j,m}^{\star}}\,\psi_{i}^{(h)},
\quad m=1,\dots,M.
\end{equation}
}The server then forms the personalized prior as a distribution mixture
{%
\setlength{\abovedisplayskip}{2.5pt}
\setlength{\belowdisplayskip}{2.5pt}
\setlength{\abovedisplayshortskip}{1.5pt}
\setlength{\belowdisplayshortskip}{1.5pt}
\begin{equation}
\label{eq:personalized-prior}
\bar{\pi}_{D,i}^{(h+1)}(\cdot\mid s^{t})
\;=\;
\sum_{m=1}^{M}
\frac{Y_{i,m}^{\star}}{\sum_{j=1}^{M}Y_{i,j}^{\star}}\,
\pi_{D,\psi^{(m,h+1)}}(\cdot\mid s^{t}),
\end{equation}
}

so that each client receives an explicit row-weighted mixture of expert
distributions rather than a single collapsed prior. The prior regularizer is defined with respect to this mixture density, while only the prior is aggregated so that the policy $\pi_{\theta,i}$ stay local to client $i$.


Given the personalized prior $\bar{\pi}_{D,i}$ from the server at round $h$, each client
maintains a parametric policy $\pi_{\theta,i}$. We write the local regularized problem as
{%
\setlength{\abovedisplayskip}{2.5pt}
\setlength{\belowdisplayskip}{2.5pt}
\setlength{\abovedisplayshortskip}{1.5pt}
\setlength{\belowdisplayshortskip}{1.5pt}
\begin{equation}
\label{eq:client-objective}
\min_{\theta_i}\;
\mathcal{J}_i(\theta_i)
\;=\;
-J_i(\pi_{\theta,i})
\;+\;\lambda_{1}\,\widehat D_{\mathrm{KL}}\!\bigl(\pi_{\theta,i}^{\,\text{old}}\,\big\|\,\pi_{\theta,i}\bigr)
\;+\;\lambda_{2}\,\widehat{\mathcal R}_{\mathrm{prior}}(\theta_i;\bar{\pi}_{D,i}).
\end{equation}
}The first regularizer is a local trust-region approximation~\cite{xie2023fedkl}. The prior regularizer $\widehat{\mathcal R}$ estimates
$\mathbb{E}_{s\sim d_i^{\mathrm{old}},\,a\sim\bar{\pi}_{D,i}(\cdot|s)}[-\log \pi_{\theta,i}(a|s)]$,
implemented with self-normalized importance weights over rollout samples, and keeps each client's policy close to its personalized offline-supported diffusion prior. $\lambda_{1}$ and $\lambda_{2}$ are the corresponding weights.

\vspace{-0.5cm}
\subsection{DICE Value Baseline Advantage Blending}
\label{sec:methods:vblend}
\vspace{-0.2cm}

The prior regularizer provides distributional alignment, while the DICE value function $V_{\phi,i}$ enters through the advantage estimator rather than directly reshaping the diffusion prior. 

Given a rollout $\{(s^t,a^t,r^t,s^{t+1})\}_{t=0}^{T-1}$ collected from
$\pi_{\theta,i}$, define a blended value baseline
{%
\setlength{\abovedisplayskip}{2.5pt}
\setlength{\belowdisplayskip}{2.5pt}
\setlength{\abovedisplayshortskip}{1.5pt}
\setlength{\belowdisplayshortskip}{1.5pt}
\begin{equation}
\label{eq:Vblend}
\mathbf{V}_{i}(s) \;=\; \beta\, V_{\theta,i}(s) \;+\; (1-\beta)\, V_{\phi,i}(s),
\quad \beta\in[0,1],
\end{equation}
}where $\beta$ is the blending coefficient with online critic $V_{\theta,i}$ . The
Generalized Advantage Estimator (GAE)~\cite{schulman2015high} is then formed with $V_{i}$
{%
\setlength{\abovedisplayskip}{2.5pt}
\setlength{\belowdisplayskip}{2.5pt}
\setlength{\abovedisplayshortskip}{1.5pt}
\setlength{\belowdisplayshortskip}{1.5pt}
\begin{equation}
\label{eq:GAE}
\hat{A}^{t}
\;=\;
\sum_{\ell=0}^{T-t-1}(\gamma\lambda_{\mathrm{GAE}})^{\ell}\,
\Bigl(r^{t+\ell}+\gamma \mathbf{V}_{i}(s^{t+\ell+1})-\mathbf{V}_{i}(s^{t+\ell})\Bigr),
\end{equation}
}where $\ell$ denotes a future temporal offsets and $\lambda_{\mathrm{GAE}}\in[0,1]$ is the GAE trace coefficient.

\begin{figure}[t]
\vspace{-0.6em}
\begin{minipage}[t]{0.49\textwidth}
\begin{algorithm}[H]
\small
\caption{FedGuide (Server)}
\label{alg:server}
\footnotesize
\begin{algorithmic}[1]
\Require Experts $\{\psi^{(m)}\}_{m=1}^{M}$; Sinkhorn temperature $\eta$; marginals $\mu_i\!=\!1/N$, $\nu_m\!=\!1/M$.
\For{round $h=1,\dots,H$}
   \State If $h=1$, initialize $\bar{\pi}_{D,i}$ from seeded experts; \\
   $\qquad$ otherwise form $\bar{\pi}_{D,i}$ from previous $Y^\star$ \eqref{eq:personalized-prior}.
   \State \textbf{Broadcast} $\bar{\pi}_{D,i}$ {\color{orange}(and $\bar{\theta}$)} to client $i$.
   \State // Server side update via Alg.~\ref{alg:client}.
   \vspace{0.05cm}
   \State \textbf{Receive} $\{\psi_i\}_{i=1}^{N}$ {\color{orange}(and $\{\theta_i\}_{i=1}^{N}$)} from clients.
   \State Build cost matrix $C_{i,m}$ via score-MSE \eqref{eq:ot-cost}.
   \State Solve Sinkhorn OT-based plan $Y^{\star}$ \eqref{eq:ot-plan}.
   \State Refresh expert heads $\psi^{(m)}$ via \eqref{eq:expert-update}.
   \State Store $Y^\star$ for next-round personalized routing.
   \State {\color{orange} Averaging received policies into $\bar{\theta}$.}
\EndFor
\end{algorithmic}
\end{algorithm}
\end{minipage}
\hfill
\begin{minipage}[t]{0.49\textwidth}
\begin{algorithm}[H]
\small
\caption{FedGuide (Client $i$)}
\label{alg:client}
\footnotesize
\begin{algorithmic}[1]
\Require Personalized prior $\bar{\pi}_{D,i}$ from server; local params $\theta_i$ {\color{orange}(or received $\bar{\theta}$)}, online critic $V_{\theta,i}$, DICE value baseline $V_{\phi,i}$, prior head $\psi_i$.
\State Collect rollout $\{(s^t,a^t,r^t,s^{t+1})\}_{t=0}^{T-1}$ with $\pi_{\theta,i}$.
\State Form blended $\mathbf{V}_{i}$ with $\beta=0.5$ {\color{red}(or $\beta=1$)} \eqref{eq:Vblend}.
\State Compute advantage $\hat A^t$ from $\mathbf{V}_{i}$ \eqref{eq:GAE}.
\For{each epoch}
   \State Update $\theta_i$ on PPO surrogate $\mathcal{L}_i$ \eqref{eq:client-loss}.
   \State Update online critic $V_{\theta,i}$ on $(s^t,\hat R^t)$.
\EndFor
\State Optionally adapt prior head $\psi_i$ locally, or keep the pretrained head.
\State \textbf{Upload} $\psi_i$ {\color{orange}(and $\theta_i$)} to server.
\State // Uncolored lines are standard FedGuide; colored parts apply only to {\color{orange} FedGuide-A} and {\color{red} FedGuide-P}.
\end{algorithmic}
\end{algorithm}
\end{minipage}
\vspace{-0.5cm}
\end{figure}

We use PPO~\cite{schulman2017proximal} for local policy optimization at each client. For client $i$, the standard clipped-ratio surrogate loss uses $\hat{A}^{t}$ in place of the usual
single-baseline advantage, and is augmented by a discrete approximation of the
prior regularizer in Eq.\eqref{eq:client-objective} with $\varepsilon>0$ as the PPO clipping radius
{%
\setlength{\abovedisplayskip}{2.5pt}
\setlength{\belowdisplayskip}{2.5pt}
\setlength{\abovedisplayshortskip}{1.5pt}
\setlength{\belowdisplayshortskip}{1.5pt}
\begin{equation}
\label{eq:client-loss}
\mathcal{L}_i(\theta_i)
=
\mathbb{E}\!\Biggl[
-\min\!\Biggl(
\frac{\pi_{\theta_i}}
{\pi_{\theta_i}^{\mathrm{old}}}
\hat{A}^{t},
\mathrm{clip}\!\Biggl(
\frac{\pi_{\theta_i}}
{\pi_{\theta_i}^{\mathrm{old}}},
1{-}\varepsilon,1{+}\varepsilon
\Biggr)\hat{A}^{t}
\Biggr)
\Biggr]
+\lambda_{2}\!\sum_{t=0}^{T-1}
\widehat{\mathcal R}_{\mathrm{prior}}(\theta_i;\bar{\pi}_{D,i})(s^{t}).
\end{equation}
}

\vspace{-0.2cm}
\subsection{Putting Everything Together: FedGuide}
\label{sec:methods:algo}
\vspace{-0.2cm}

We now summarize our proposed method, FedGuide. The server
(Alg.~\ref{alg:server}) maintains $M$ diffusion-prior experts and, after
receiving client prior heads, routes clients to experts via the Sinkhorn OT-based plan
from Eq.\eqref{eq:ot-plan}. It then refreshes each expert head by the column-normalized
combination in Eq.\eqref{eq:expert-update} and broadcasts a personalized distribution
mixture prior $\bar{\pi}_{D,i}$ to each client. By receiving
prior, every client $i$ (Alg.~\ref{alg:client}) performs $U$ local updates
on the prior-regularized loss from Eq.\eqref{eq:client-loss}, using the
Eq.\eqref{eq:Vblend}. In the implementation, each client uploads only its current
diffusion prior head $\psi_i$; the policy $\pi_{\theta,i}$, online critic $V_{\theta,i}$, and DICE value baseline
$V_{\phi,i}$ stay local, and the server never broadcasts or averages policy parameters. Policy averaging occurs only in {\color{orange}FedGuide-A}, and dropping the DICE baseline ($\beta=1$) only in {\color{red}FedGuide-P}; the colored parts of Alg.~\ref{alg:server} and Alg.~\ref{alg:client} mark exactly where each variant departs from standard FedGuide. The blend weight $\beta$ can be fixed or annealed from stronger offline guidance at early
rounds toward more online-critic weight later as $V_{\theta,i}$ becomes more reliable, while
the trust-region term in Eq.\eqref{eq:client-objective} stabilizes local policy updates.

\vspace{-0.2cm}
\section{Theoretical Analysis}
\label{sec:theory}
\vspace{-0.3cm}

We summarize the main theoretical properties of FedGuide and defer detailed
proofs to Appendix A. The analysis isolates the effects of the
DICE value baseline, the OT-MoE prior aggregation, and the local KL trust
region. Let $\tilde F(\psi,\theta)$ denote the federated surrogate that combines
the client objectives and the prior-aggregation penalty. We assume (\textbf{1}) standard
$L$-smoothness of $\tilde F$, (\textbf{2}) bounded stochastic-gradient variance
$\varsigma^{2}$, (\textbf{3}) bounded local KL drift
$D_{\mathrm{KL}}(\pi_{\theta,i}^{(h+1)}\|\pi_{\theta,i}^{(h)})\le
\delta$, (\textbf{4}) bounded approximation errors from the practical prior
surrogate, DICE value baseline, and OT cost proxy, denoted by
$\epsilon_{\mathrm{SNIS}}$, $\epsilon_V$, and $\epsilon_C$, respectively, and
(\textbf{5}) a bounded Sinkhorn approximation error $\epsilon_{\mathrm{OT}}$.

\vspace{-0.2cm}
\vspace{-0.2cm}

\paragraph{Lemma 1 (State-only baseline does not add score function bias).}
For any baseline $\mathbf{V}_i(s^t)$,
\begin{equation}
\label{eq:lemma1}
\mathbb{E}_{a^t\sim\pi_{\theta,i}(\cdot\mid s^t)}
\!\left[
\nabla_{\theta_i}\log \pi_{\theta,i}(a^t\mid s^t)\mathbf{V}_i(s^t)
\right]
=0.
\end{equation}
Therefore, using the baseline
$\mathbf{V}_i(s)=\beta V_{\theta,i}(s)+(1-\beta)V_{\phi,i}(s)$ does not introduce
direct action-dependent score-function bias into the policy-gradient estimator. The DICE
value baseline $V_{\phi,i}$ only centers the advantage estimate, and it does not directly
reshape the diffusion prior or evaluate generated actions; any bootstrapping
error from GAE is accounted for separately.

\vspace{-0.3cm}
\paragraph{Lemma 2 (Variance reduction).}
Let $G_i^t$ be the return target. Define $X_i^t=G_i^t-V_{\theta,i}(s^t)$ and
$Z_i^t=V_{\phi,i}(s^t)-V_{\theta,i}(s^t)$, with
$\kappa_i=\mathrm{Corr}(X_i^t,Z_i^t)$. If
$\kappa_i>0$, then the $\mathbf{V}_{i}(s)$ acts as a control
variate for a coefficient $\beta$. In particular,
\begin{math}
\varsigma_{\mathrm{eff}}^{2}
\le
\varsigma^{2}\bigl(1-c_\beta\bar{\kappa}^{2}\bigr)
\end{math} and 
\begin{math}
\bar{\kappa}^{2}=\sum_{i=1}^{N}\zeta_i\kappa_i^2,
\end{math}
where $c_\beta>0$ depends on the $\beta$ and $\zeta_i$ is the client
aggregation weight with $\sum_i\zeta_i=1$. Thus, the DICE value baseline improves the stochastic constant whenever it is positively correlated with the online-baseline residual.

\vspace{-0.3cm}

\paragraph{Theorem 1 (Convergence of FedGuide).}
Under the assumptions above, with policy-optimization stepsize $\eta_{\mathrm{pg}}\le 1/(4L)$, after $H$
federated rounds and $U$ local updates per round, FedGuide satisfies
\begin{equation}
\label{eq:thm1}
\frac{1}{H}\sum_{h=0}^{H-1}
\mathbb{E}\bigl\|\nabla \tilde F(\psi^{(h)},\theta^{(h)})\bigr\|^{2}
\;\le\;
\mathcal{O}\!\left(\frac{\tilde F^{(0)}-\tilde F^\star}{\eta_{\mathrm{pg}} H}\right)
+
\mathcal{O}\!\left(\eta_{\mathrm{pg}} L\frac{\varsigma_{\mathrm{eff}}^{2}}{NU}\right)
+
\mathcal{O}\!\left(\Delta\right).
\end{equation}
where
\begin{math}
\Delta=\epsilon_{\mathrm{OT}}+\epsilon_C+\epsilon_{\mathrm{SNIS}}+\epsilon_V+\delta.
\end{math}
By Lemma~1, the state-only DICE value baseline does not introduce direct score-function bias, so its approximation error is absorbed into $\epsilon_V$ rather than appearing as an additional policy-gradient bias. By Lemma~2, positive residual correlation reduces the stochastic-gradient constant from $\varsigma^{2}$ to $\varsigma_{\mathrm{eff}}^{2}$. Since the policy $\pi_{\theta,i}$ and the DICE value baseline $V_{\phi,i}$ remain local to each client, no policy-averaging heterogeneity term appears in the bound.

\newpage
\section{Experimental Analysis}
\label{sec:Exp}
\vspace{-0.3cm}
\begin{wraptable}{r}{0.53\textwidth}
    \centering
    \small
    \setlength{\tabcolsep}{1.3pt} 
    \renewcommand{\arraystretch}{0.45}
    \begin{tabular}{l|c|c|c}
        \toprule
        Variants & Diffusion Prior & DICE Value & Client Policy \\
        \midrule
        FG-A & OT-MoE & Local & Avg. \\
        FG-P & OT-MoE & N/A   & Local \\
        FG   & OT-MoE & Local & Local \\
        \bottomrule
    \end{tabular}
    \vspace{-0.1cm}
    \caption{Comparison of FedGuide (FG) variants: FedGuide keeps the DICE value baseline $\mathbf{V}_i$ local, while FedGuide-P does not use $V_{\phi,i}$ ($\beta=1$). FedGuide-A uses $V_{\phi,i}$ but periodically averages client policies at the server. All variants use OT-MoE aggregation for diffusion prior.}
    \label{tab:baselines}
    \vspace{-0.4cm}
\end{wraptable}

We aim to explore the performance of our method through various environments to address the following three questions: \textbf{RQ1:} How does FedGuide compare to SOTA FRL methods in terms of evaluation return of clients, across different tasks and heterogeneous reinforcement learning environments? \textbf{RQ2:} Is FedGuide robust with respect to number of clients and harder heterogeneous setting within different environments? \textbf{RQ3:} What role do the diffusion prior and value baseline play, and why are they important in the heterogeneous federated settings?

\vspace{-0.3cm}
\begin{wraptable}{r}{0.43\linewidth}
  \vspace{-0.55cm}
  \centering
  \small
  \setlength{\tabcolsep}{3pt}
  \renewcommand{\arraystretch}{0.8}
  \begin{tabular}{l ccc}
    \toprule
    & \multicolumn{3}{c}{$\mathrm{CV}_{\sigma_T}$ $\downarrow$} \\
    \cmidrule(lr){2-4}
    Env & FG-A & FG-P & FG \\
    \midrule
    Reacher & 0.380 & \underline{0.068} & \textbf{0.053} \\
    Hopper & 0.402 & \underline{0.164} & \textbf{0.155} \\
    Walker2D & 0.208 & \underline{0.179} & \textbf{0.127} \\
    HalfCheetah & 0.628 & \textbf{0.235} & \underline{0.241} \\
    MetaWorld10 & 0.364 & \underline{0.272} & \textbf{0.260} \\
    \midrule
    Reacher-Hard & 2.097 & \textbf{0.229} & \underline{0.342} \\
    Hopper-Hard & 0.377 & \underline{0.185} & \textbf{0.133} \\
    Walker2D-Hard & 0.178 & \underline{0.155} & \textbf{0.088} \\
    HalfCheetah-Hard & 0.591 & \underline{0.231} & \textbf{0.201} \\
    \bottomrule
  \end{tabular}
    \caption{Normalized temporal volatility $\mathrm{CV}_{\sigma_T}$ for FedGuide (FG) and variants, computed over the last 20 rounds as the across-seed mean of return standard deviation normalized by the absolute mean return. Smaller values indicate smoother late-stage returns and are consistent with, but do not directly measure, the variance-reduction effect in Lemma~2.}
  \label{tab:fg_family_ablation}
  \vspace{-0.5cm}
\end{wraptable}

\paragraph{Experimental Setup.}
We evaluate \textbf{FedGuide} in simulation on heterogeneous FRL benchmarks where clients face different MDP variants in rewards, goals, dynamics, morphology, objects, or scenarios. (\textbf{1}) Bandit2D, a 2D continuous bandit with multimodal reward landscapes, used to visualize the significance of diffusion priors. (\textbf{2}) Reacher~\cite{brockman2016openai}, a continuous goal-reaching task where heterogeneity comes from different target positions and environment parameters. (\textbf{3}) MuJoCo locomotion~\cite{todorov2012mujoco} includes Hopper, Walker2D, and HalfCheetah, testing transfer across clients with different morphologies, contact dynamics, and gait objectives. (\textbf{4}) MetaWorld10~\cite{yu2020meta} covers manipulation tasks such as reaching, pushing, and grasping, with heterogeneity from object layouts, contacts, and rewards. Demonstrations of environments are shown in Fig.~\ref{fig:placeholder}, and settings are detailed in Appendix B.

\begin{figure}[t]
    \centering

    \setlength{\fboxsep}{2pt}
    \setlength{\fboxrule}{0.6pt}

    \begin{minipage}[t]{0.49\linewidth}
        \vspace{0pt}
        \centering

        \newlength{\cellw}
        \setlength{\cellw}{0.25\linewidth}

        \newlength{\topboxh}
        \setlength{\topboxh}{1.12\cellw}

        \newlength{\topimg}
        \setlength{\topimg}{0.78\cellw}

        \newcommand{\singleenvbox}[2]{%
            \begin{minipage}[t]{\cellw}
                \centering
                \fbox{%
                    \begin{minipage}[c][\topboxh][c]{\dimexpr\cellw-2\fboxsep-2\fboxrule\relax}
                        \centering
                        \IfFileExists{#2}{%
                            \includegraphics[
                                width=\topimg,
                                height=\topimg,
                                keepaspectratio
                            ]{#2}%
                        }{%
                            \phantom{\rule{\topimg}{\topimg}}%
                        }\\[-0.15em]
                        {\scriptsize\textbf{#1}}
                    \end{minipage}%
                }%
            \end{minipage}%
        }

        \newcommand{\groupenvimg}[2]{%
            \begin{minipage}[c]{0.333\linewidth}
                \centering
                \IfFileExists{#2}{%
                    \includegraphics[
                        width=\topimg,
                        height=\topimg,
                        keepaspectratio
                    ]{#2}%
                }{%
                    \phantom{\rule{\topimg}{\topimg}}%
                }\\[-0.15em]
                {\scriptsize\textbf{#1}}
            \end{minipage}%
        }

        \newcommand{\groupenvbox}{%
            \fbox{%
                \begin{minipage}[c][\topboxh][c]{\dimexpr 3\cellw-2\fboxsep-2\fboxrule\relax}
                    \centering
                    \groupenvimg{Hopper}{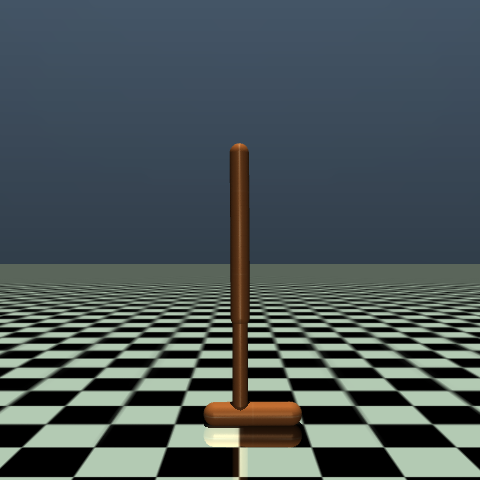}%
                    \groupenvimg{Walker2D}{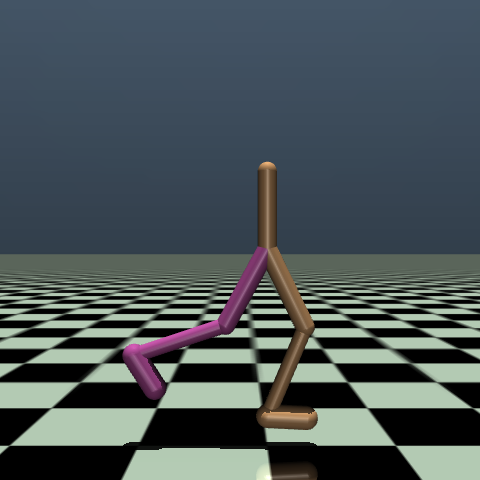}%
                    \groupenvimg{HalfCheetah}{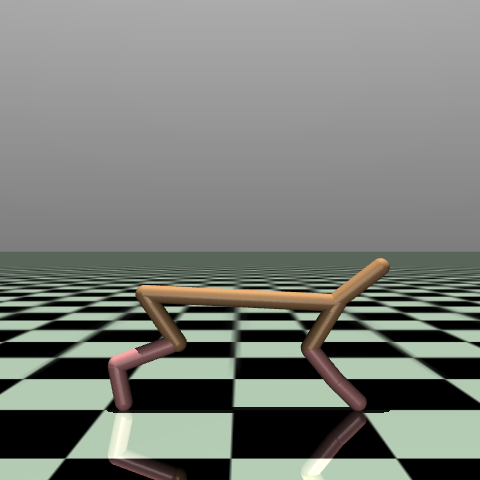}%
                \end{minipage}%
            }%
        }

        \noindent
        \makebox[\linewidth][c]{%
            \singleenvbox{Reacher}{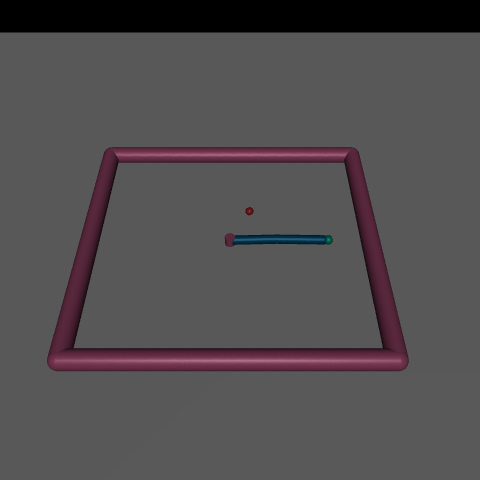}%
            \groupenvbox%
        }

        \fbox{%
            \begin{minipage}[t]{\dimexpr\linewidth-2\fboxsep-2\fboxrule\relax}
                \centering
                \includegraphics[
                    width=\linewidth
                ]{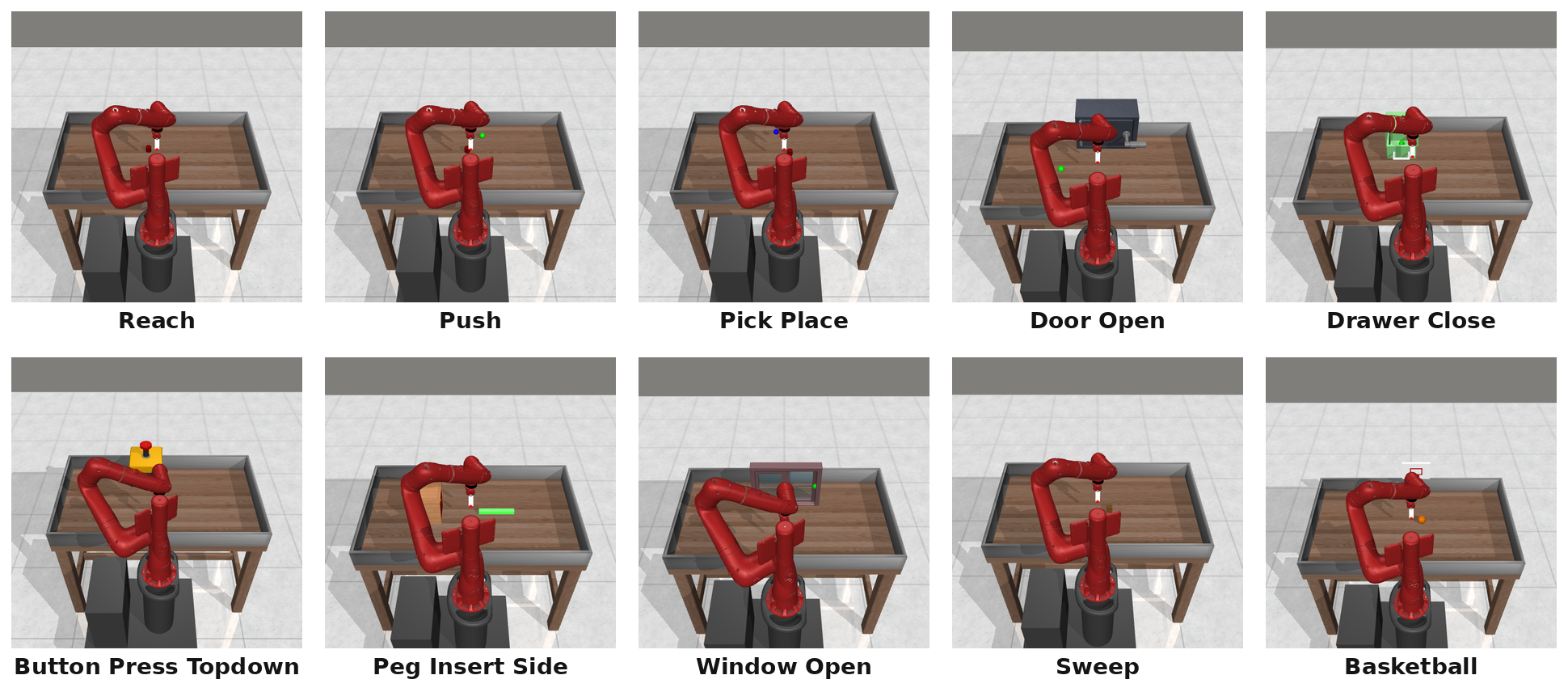}\\[-0.15em]
                {\scriptsize\textbf{MetaWorld10}}
            \end{minipage}%
        }

    \end{minipage}
    \hfill
    \begin{minipage}[t]{0.5\linewidth}
    \vspace{0pt}
    \centering

    \fbox{%
        \begin{minipage}[t]{\dimexpr0.63\linewidth-2\fboxsep-2\fboxrule\relax}
            \centering
    
            \begin{minipage}[t]{0.48\linewidth}
                \centering
                \includegraphics[
                    width=\linewidth,
                    height=0.95\linewidth
                ]{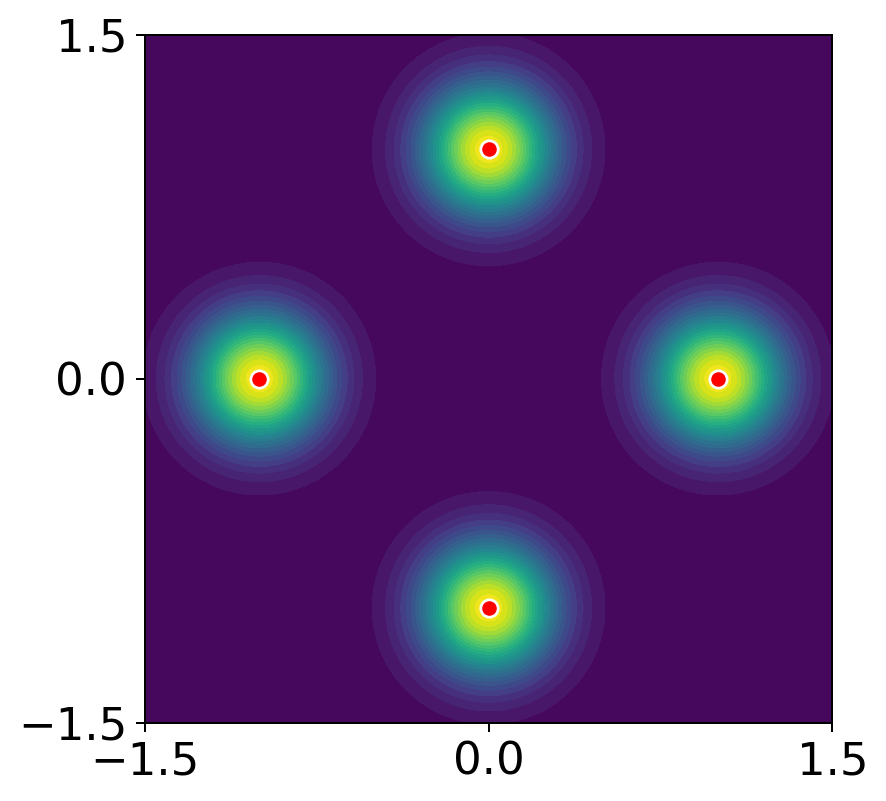}%
            \end{minipage}
            \hfill
            \begin{minipage}[t]{0.48\linewidth}
                \centering
                \includegraphics[
                    width=\linewidth,
                    height=0.95\linewidth
                ]{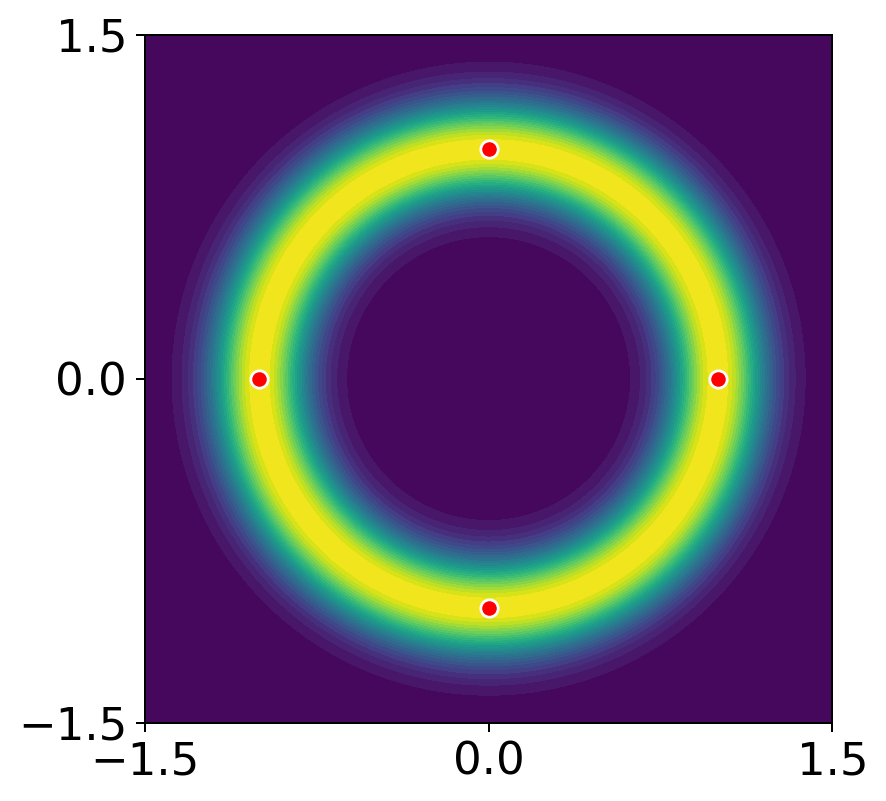}%
            \end{minipage}
    
            \par\vspace{-0.1em}
            {\scriptsize\textbf{Bandit2D Ground Truth}}
        \end{minipage}%
    }
    \hfill
    \fbox{%
        \begin{minipage}[t]{\dimexpr0.35\linewidth-2\fboxsep-2\fboxrule\relax}
            \centering
            \includegraphics[
                width=\linewidth,
                height=0.85\linewidth
            ]{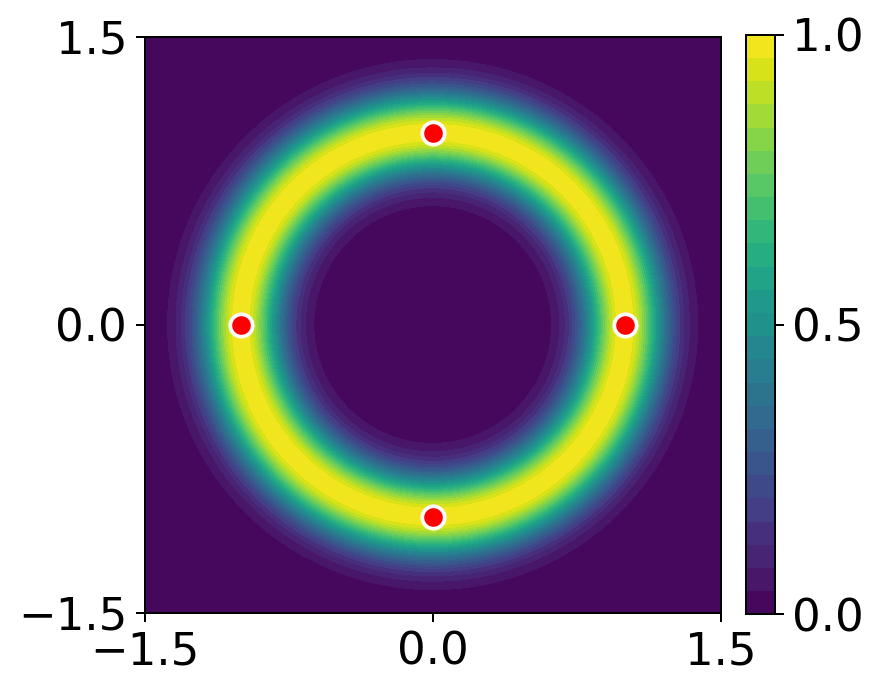}%
    
            \par\vspace{-0.1em}
            {\scriptsize\textbf{OTMoE Global Prior}}
        \end{minipage}%
    }
    
    \vspace{0.45em}

    \begin{minipage}[c]{\dimexpr\linewidth-2\fboxsep-2\fboxrule\relax}
        \centering
        \begin{minipage}[c]{0.94\linewidth}
            \centering
            \includegraphics[
                width=\linewidth
            ]{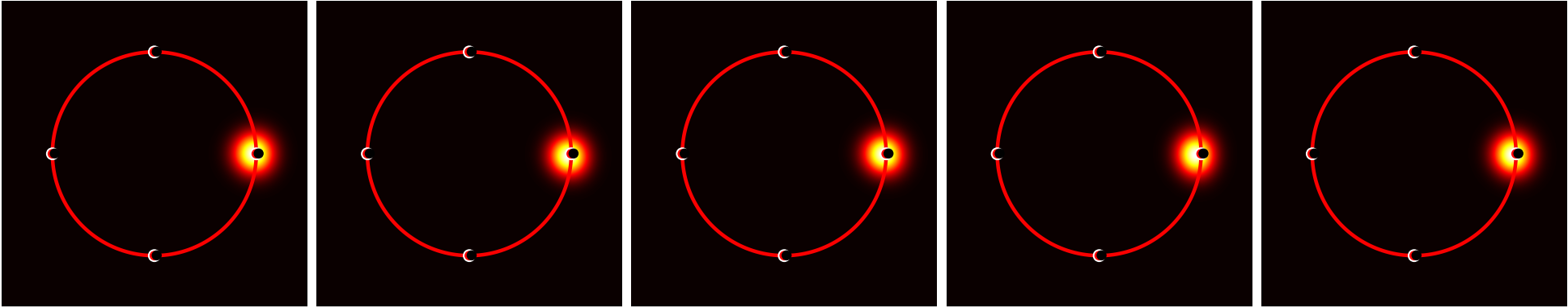}\\
        \end{minipage}%
        \hspace{0.01\linewidth}%
        \begin{minipage}[c]{0.05\linewidth}
            \centering
            \rotatebox[origin=c]{270}{\textbf{FedAvg}}
        \end{minipage}%
    \end{minipage}%
    \hfill
    \begin{minipage}[c]{\dimexpr\linewidth-2\fboxsep-2\fboxrule\relax}
        \centering
        \begin{minipage}[c]{0.94\linewidth}
            \centering
            \includegraphics[
                width=\linewidth
            ]{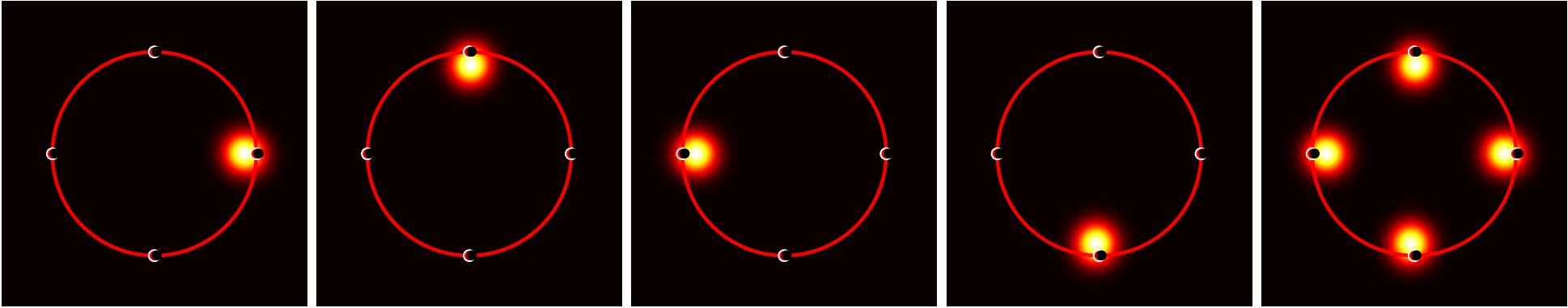}\\
        \end{minipage}%
        \hspace{0.01\linewidth}%
        \begin{minipage}[c]{0.05\linewidth}
            \centering
            \rotatebox[origin=c]{270}{\textbf{FedGuide}}
        \end{minipage}%
    \end{minipage}%

    \vspace{0em}
    {\scriptsize
    \par\noindent
    \makebox[\linewidth][l]{%
        \begin{minipage}[t]{0.2\linewidth}
            \centering
            \textbf{Client 1}
        \end{minipage}%
        \begin{minipage}[t]{0.18\linewidth}
            \centering
            \textbf{Client 2}
        \end{minipage}%
        \begin{minipage}[t]{0.18\linewidth}
            \centering
            \textbf{Client 3}
        \end{minipage}%
        \begin{minipage}[t]{0.18\linewidth}
            \centering
            \textbf{Client 4}
        \end{minipage}%
        \begin{minipage}[t]{0.2\linewidth}
            \centering
            \textbf{Global}
        \end{minipage}%
    }
    }
    \end{minipage}
    \vspace{-0.55cm}
    \caption{Left: Demonstrations of different heterogeneous FRL environments: Reacher, Walker2D, Hopper, HalfCheetah, and MetaWorld10 with 10 heterogeneous manipulation tasks. Right: Visualization of toy-case Bandit2D with 4 isotropic Gaussian reward peaks ($\sigma{=}0.2$) evenly spaced on the unit circle, each client sampled from an overlapping $120^\circ$ sector of the annulus $r\in[0.7,1.3]$ centered on one peak. All clients in FedAvg collapse to one peak, while FedGuide (including its two variants) captures each peak; its global diffusion prior recovers the ring-shaped union of client supports through OT-MoE aggregation.}
    \label{fig:placeholder}
    \vspace{-0.6cm}
\end{figure}

\begin{figure*}[tb!]
  \centering
  \caption{Client average return for 100 rounds at different heterogeneous FRL environments. We set client numbers to 10 in MetaWorld10 and keep 8 clients in other environments, evaluating on 5 random seeds with Behavior Cloning warm-up. Color regions indicate one standard deviation.}
  \label{fig:results_curves}
  \includegraphics[width=\linewidth]{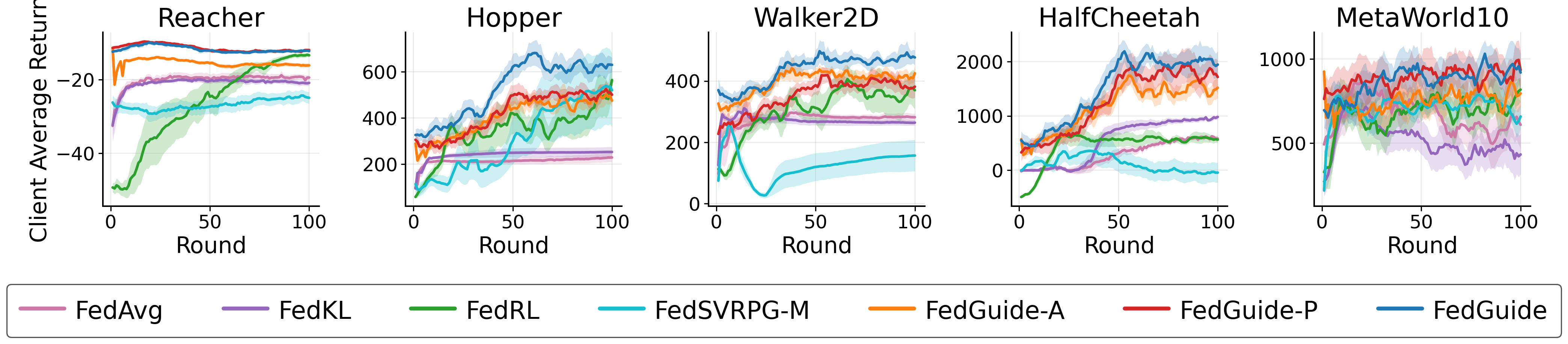}
  \begin{minipage}{\linewidth}
  \centering
  \scriptsize
  \setlength{\tabcolsep}{4pt}
  \renewcommand{\arraystretch}{0}
  \begin{tabular}{l l ccccccc}
    \toprule
    Environment & Metric & FedAvg & FedKL & FedRL & FedSVRPG-M\textsuperscript{\textdagger} & FedGuide-A & FedGuide-P & FedGuide \\
    \midrule
    \multirow{2}{*}{Reacher} & $\mathbf{F}\uparrow$ & -19.4{\scriptsize$\pm$1.82} & -20.9{\scriptsize$\pm$1.63} & -13.4{\scriptsize$\pm$1.16} & -24.9{\scriptsize$\pm$3.46} & -16.1{\scriptsize$\pm$0.48} & \textbf{-12.2}{\scriptsize$\pm$0.35} & \underline{-12.3}{\scriptsize$\pm$0.67} \\
     & $\mathbf{W}\uparrow$ & -21.4 & -22.3 & -14.4 & -29.8 & -16.8 & \textbf{-12.6} & \underline{-13.0} \\
    \cmidrule(lr){1-9}
    \multirow{2}{*}{Hopper} & $\mathbf{F}\uparrow$ & 227{\scriptsize$\pm$42.5} & 253{\scriptsize$\pm$48.1} & \underline{527}{\scriptsize$\pm$120} & 525{\scriptsize$\pm$360} & 481{\scriptsize$\pm$26.5} & 507{\scriptsize$\pm$19.3} & \textbf{632}{\scriptsize$\pm$19.0} \\
     & $\mathbf{W}\uparrow$ & 180 & 207 & 380 & 166 & 451 & \underline{481} & \textbf{614} \\
    \cmidrule(lr){1-9}
    \multirow{2}{*}{Walker2D} & $\mathbf{F}\uparrow$ & 283{\scriptsize$\pm$27.4} & 265{\scriptsize$\pm$21.4} & 360{\scriptsize$\pm$113} & 156{\scriptsize$\pm$113} & \underline{422}{\scriptsize$\pm$31.5} & 383{\scriptsize$\pm$18.7} & \textbf{480}{\scriptsize$\pm$18.6} \\
     & $\mathbf{W}\uparrow$ & 252 & 250 & 219 & 21.3 & \underline{377} & 353 & \textbf{461} \\
    \cmidrule(lr){1-9}
    \multirow{2}{*}{HalfCheetah} & $\mathbf{F}\uparrow$ & 600{\scriptsize$\pm$92.1} & 950{\scriptsize$\pm$90.7} & 579{\scriptsize$\pm$145} & -42.2{\scriptsize$\pm$393} & 1439{\scriptsize$\pm$176} & \underline{1768}{\scriptsize$\pm$500} & \textbf{1994}{\scriptsize$\pm$304} \\
     & $\mathbf{W}\uparrow$ & 486 & 821 & 421 & -409 & \underline{1246} & 978 & \textbf{1681} \\
    \cmidrule(lr){1-9}
    \multirow{2}{*}{MetaWorld10} & $\mathbf{F}\uparrow$ & 631{\scriptsize$\pm$238} & 445{\scriptsize$\pm$213} & 767{\scriptsize$\pm$53.6} & 668{\scriptsize$\pm$102} & 829{\scriptsize$\pm$82.5} & \underline{914}{\scriptsize$\pm$58.2} & \textbf{939}{\scriptsize$\pm$141} \\
     & $\mathbf{W}\uparrow$ & 248 & 313 & 724 & 547 & \underline{744} & \textbf{816} & 716 \\
    \specialrule{1pt}{2pt}{3pt}
    \multirow{2}{*}{Reacher-Hard} & $\mathbf{F}\uparrow$ & -2531{\scriptsize$\pm$1823} & -4274{\scriptsize$\pm$2070} & -90.2{\scriptsize$\pm$4.97} & -8339{\scriptsize$\pm$6580} & -1995{\scriptsize$\pm$128} & \textbf{90.7}{\scriptsize$\pm$15.7} & \underline{86.5}{\scriptsize$\pm$37.6} \\
     & $\mathbf{W}\uparrow$ & -5210 & -7118 & -94.8 & -16456 & -2148 & \textbf{67.4} & \underline{37.6} \\
    \cmidrule(lr){1-9}
    \multirow{2}{*}{Hopper-Hard} & $\mathbf{F}\uparrow$ & 216{\scriptsize$\pm$20.1} & 242{\scriptsize$\pm$41.0} & \textbf{699}{\scriptsize$\pm$256} & 387{\scriptsize$\pm$285} & 398{\scriptsize$\pm$10.9} & 437{\scriptsize$\pm$51.6} & \underline{510}{\scriptsize$\pm$44.0} \\
     & $\mathbf{W}\uparrow$ & 187 & 197 & \underline{404} & 169 & 384 & 383 & \textbf{445} \\
    \cmidrule(lr){1-9}
    \multirow{2}{*}{Walker2D-Hard} & $\mathbf{F}\uparrow$ & 256{\scriptsize$\pm$13.2} & 286{\scriptsize$\pm$24.3} & 321{\scriptsize$\pm$57.8} & 57.8{\scriptsize$\pm$24.5} & \underline{380}{\scriptsize$\pm$10.9} & 361{\scriptsize$\pm$25.3} & \textbf{415}{\scriptsize$\pm$14.5} \\
     & $\mathbf{W}\uparrow$ & 241 & 258 & 258 & 27.3 & \underline{368} & 327 & \textbf{396} \\
    \cmidrule(lr){1-9}
    \multirow{2}{*}{HalfCheetah-Hard} & $\mathbf{F}\uparrow$ & 231{\scriptsize$\pm$451} & 373{\scriptsize$\pm$110} & 832{\scriptsize$\pm$349} & 96.1{\scriptsize$\pm$251} & 1272{\scriptsize$\pm$203} & \underline{1519}{\scriptsize$\pm$327} & \textbf{2146}{\scriptsize$\pm$304} \\
     & $\mathbf{W}\uparrow$ & -544 & 217 & 328 & -282 & 928 & \underline{1168} & \textbf{1893} \\
    \bottomrule
  \end{tabular}
  \vspace{-0.1cm}
  \captionof{table}{Quantitative results of $\mathbf{F}$inal-round and $\mathbf{W}$orst-round client average return performance across different heterogeneous FRL environments and seeds, along with results for stronger heterogeneity robustness (“Hard” ) experiments. We use $M=8$ for all environments. \textsuperscript{\textdagger} denotes the method is implemented by us.}
  \label{tab:results_metrics}
  \vspace{-0.6cm}
\end{minipage}

\end{figure*}

\vspace{-0.2cm}
\textbf{Baselines.} We compare our method, \textbf{FedGuide}, with 
(\textbf{1}) FedAvg~\cite{mcmahan2017communication}, a standard federated baseline that directly averages local policy parameters; (\textbf{2}) FedKL~\cite{xie2023fedkl}, a FRL method that constrains local policy updates by KL regularization; (\textbf{3}) FedRL~\cite{jin2022federated}, a federated actor-critic baseline that aggregates local Deep Deterministic Policy Gradient (DDPG) agents; (\textbf{4}) FedSVRPG-M~\cite{wang2024momentum}, a momentum-based method that aggregates variance-reduced policy-gradient updates; and our three variants: \textbf{FedGuide-A}, \textbf{FedGuide-P}, and \textbf{FedGuide}. Main difference of our variants are summarized in Table~\ref{tab:baselines}. Baselines are described in detail in Appendix C.

\vspace{-0.2cm}
\subsection{Experimental Results}
\vspace{-0.2cm}
\paragraph{Performance for Heterogeneous FRL.}
As illustrated in Fig.~\ref{fig:results_curves}, we showcase the client average
return across all federated clients for tested environments. FedGuide and its
variants generally achieve higher returns over the training horizon. FedGuide also
remains robust in terms of final-round and worst-round performance
(Table~\ref{tab:results_metrics}), indicating that the diffusion prior and the
DICE value baseline improve not only peak performance but also training
stability under heterogeneous clients. In contrast, FedAvg and FedKL, which rely
on direct policy averaging or shared-policy KL regularization, either stagnate at low returns on Reacher and
the three locomotion tasks or degrade after around 30 rounds in MetaWorld10.
FedRL obtains more stable increasing returns but shows lower early-stage
returns, likely because its off-policy DDPG backbone depends on sufficient
replay-buffer coverage for stable critic learning, and it fails in HalfCheetah. FedSVRPG-M is competitive with FedRL on Hopper, but shows larger variance and weak worst-seed returns on others, suggesting that momentum-based update correction alone does not remove distribution mismatch.
We notice that all methods only obtain slight improvement with higher variance in MetaWorld10, where FedGuide family increases only around 200$\sim$300, indicating that FRL methods still struggle with heterogeneous task geometry, beyond the environment-parameter heterogeneity mostly studied so far. The cause is that MetaWorld10 clients differ in task semantics rather than in environment parameters: objects, contact sequences, reward geometries, and success conditions all change across clients, so behavior-support overlap is weak and each personalized prior carries less transferable support. This is consistent with the FedKL~\cite{xie2023fedkl} analysis, where stronger heterogeneity increases disagreement among local policy advantages. FedGuide still attains the best final return ($767\!\rightarrow\!939$), showing that distributional guidance remains beneficial, while task-aware prior representations are needed to close the remaining gap.

\vspace{-0.2cm}
\begin{wrapfigure}{r}{0.45\linewidth}
  \centering
  \includegraphics[width=\linewidth]{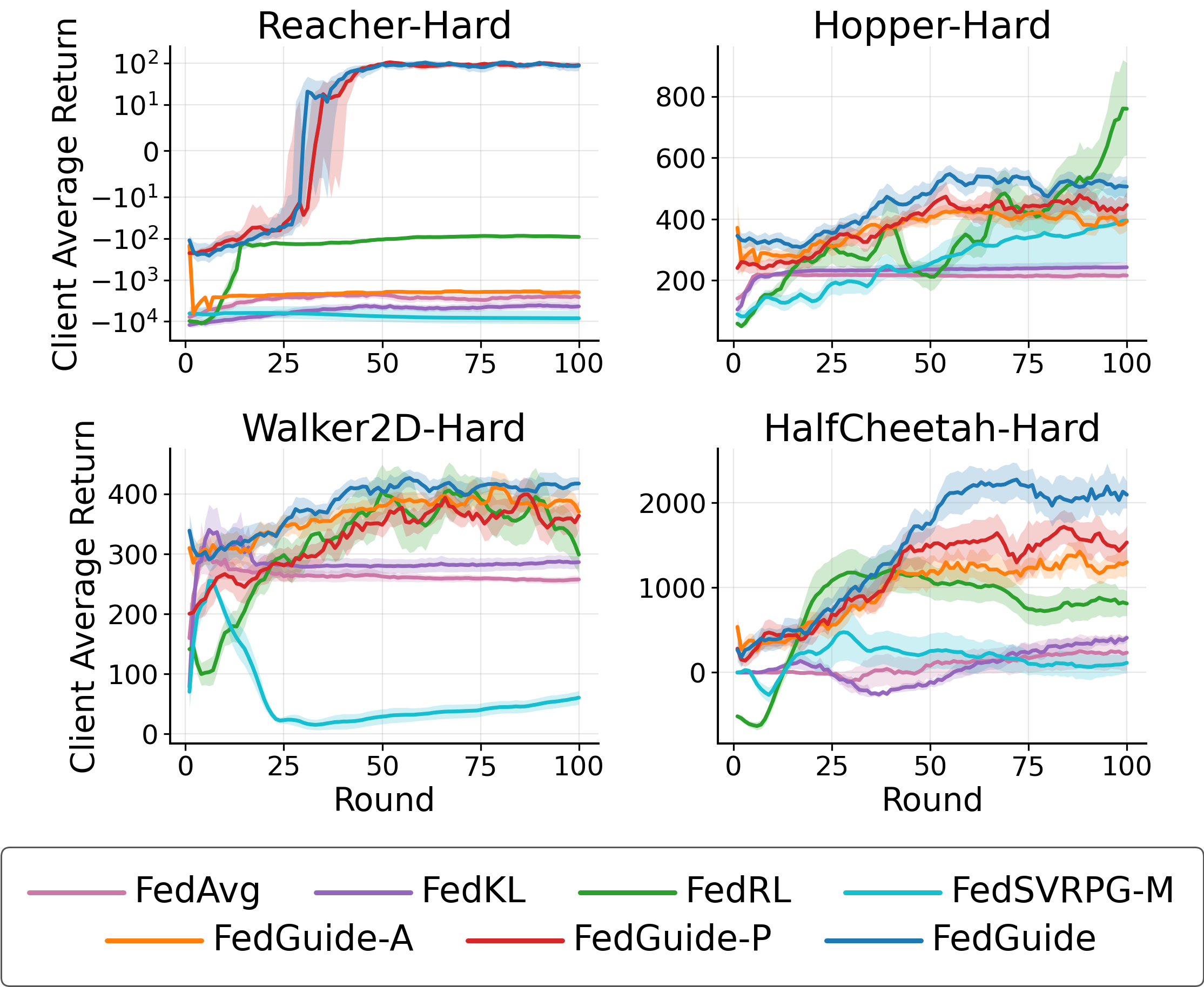}
  \vspace{-0.55cm}
  \caption{Client average return for FRL environments with stronger heterogeneity (``Hard"). We fixed client numbers at 8 and the same experimental settings with Fig.~\ref{fig:results_curves}.}
  \label{fig:ablation_hard}
  \vspace{-0.3cm}
\end{wrapfigure}

\paragraph{Discussion of Robustness.}
We continue to evaluate robustness of our method within different environmental setups with increased heterogeneity, which we called ``Hard" environments (Detailed in Appendix B). Fig.~\ref{fig:ablation_hard} shows the FedGuide family attains the best final on Walker2D, HalfCheetah, Reacher, with FedRL leading on Hopper-Hard. In Reacher-Hard, FedAvg and FedKL collapse to $-10^{3}$ returns and FedRL stalls at $-100$, while the FedGuide variants converge to around $100$, indicating the OT-MoE prior keeps each policy in its local behavior support even under MDP-level heterogeneity. Together with MetaWorld10's 10-client setting, this suggests robustness of FedGuide to both stronger heterogeneity and varying client counts.
FedSVRPG-M also degrades sharply on Reacher-Hard and remains below the FedGuide family on Walker2D-Hard and HalfCheetah-Hard, reinforcing the need for distribution alignment under stronger heterogeneity.

\vspace{-0.2cm}
\paragraph{Why Diffusion Prior and DICE Value Baseline?}
We visualize toy-case Bandit2D to better interpret the importance of diffusion prior and DICE value baseline. In Fig.~\ref{fig:placeholder}, the FedGuide family captures all four reward peaks, while FedAvg collapses to one peak. This indicates that OT-MoE aggregation over diffusion priors preserves heterogeneous behavior distributions in distribution space, and even helps FedGuide-A avoid mode collapse. 
Table~\ref{tab:fg_family_ablation} further shows the role of the value baseline: using it usually lowers late-stage return volatility, indicating that the prior provides support while the DICE value baseline improves return-aware local updates. Overall, FedGuide often improves more stably because it combines multimodal behavioral support with local DICE-based policy improvement. 


\vspace{-0.2cm}
\section{Limitations}
\label{sec:limitations}
\vspace{-0.3cm}
All results are obtained in simulation; we do not evaluate on physical robots, so whether distribution-level alignment survives real perception noise, actuation latency, and contact uncertainty remains open. Communicating only diffusion-prior heads keeps policies and trajectories local, but this is not a formal privacy guarantee and we do not claim privacy-preserving FRL. FedGuide further assumes shared state-action spaces and meaningful overlap among client behavior priors, and may fail under severe task-geometry or reward shifts. Finally, OT-MoE cost grows with client and expert counts, and setting $M$ adaptively rather than from $N$ remains unresolved.

\vspace{-0.2cm}
\section{Conclusions}
\label{sec:conclusion}
\vspace{-0.3cm}
We presented FedGuide, a heterogeneous FRL framework that aggregates diffusion behavior priors through OT-MoE and broadcasts personalized distribution-mixture priors to local clients, while a DICE value baseline provides return-aware local policy improvement. Across heterogeneous benchmarks FedGuide improves client-average, final-round, and worst-round return over representative FRL methods, and the Bandit2D study shows that OT-MoE combines represented behavior supports rather than extrapolating unsupported modes. Extending distribution-level alignment to task-semantic heterogeneity and to physical multi-robot deployment are the natural next steps.

\section*{Acknowledgments}
\vspace{-0.3cm}
This work was partially supported by NSF grants CCF 2045694, CNS-2112471, CPS-2111751, CCF-2428569 and ONR grant N00014-23-1-2149. This work used Bridges-2 GPU at the Pittsburgh Supercomputing Center from the Advanced Cyberinfrastructure Coordination Ecosystem: Services \& Support (ACCESS) program, which is supported by NSF grants \#2138259, \#2138286, \#2138307, \#2137603, and \#2138296 \citep{access}.

\bibliography{reference}
\newpage
\appendix
\section*{Appendix Contents}

\begingroup
\small
\setcounter{tocdepth}{2}
\makeatletter
\@dottedtocline{1}{0em}{2.4em}{\hyperref[app:proofs]{\ref*{app:proofs}\quad Theoretical Proofs}}{\pageref*{app:proofs}}
\@dottedtocline{2}{1.6em}{3.2em}{\hyperref[app:assumptions]{\ref*{app:assumptions}\quad Assumptions}}{\pageref*{app:assumptions}}
\@dottedtocline{2}{1.6em}{3.2em}{\hyperref[app:variance_reduction]{\ref*{app:variance_reduction}\quad Variance Reduction from the Value Baseline}}{\pageref*{app:variance_reduction}}
\@dottedtocline{2}{1.6em}{3.2em}{\hyperref[app:federated_convergence]{\ref*{app:federated_convergence}\quad Federated Convergence}}{\pageref*{app:federated_convergence}}
\@dottedtocline{2}{1.6em}{3.2em}{\hyperref[app:policy-avg]{\ref*{app:policy-avg}\quad Policy Averaging under Prior-Regularized Alignment}}{\pageref*{app:policy-avg}}
\@dottedtocline{1}{0em}{2.4em}{\hyperref[app:env_setup]{\ref*{app:env_setup}\quad Environmental Setup}}{\pageref*{app:env_setup}}
\@dottedtocline{2}{1.6em}{3.2em}{\hyperref[app:env_detail]{\ref*{app:env_detail}\quad Environment Description}}{\pageref*{app:env_detail}}
\@dottedtocline{2}{1.6em}{3.2em}{\hyperref[app:heterogeneous_setup]{\ref*{app:heterogeneous_setup}\quad Heterogeneous Setup}}{\pageref*{app:heterogeneous_setup}}
\@dottedtocline{2}{1.6em}{3.2em}{\hyperref[app:hard_heterogeneous_setup]{\ref*{app:hard_heterogeneous_setup}\quad Heterogeneous Setup for Stronger Environments}}{\pageref*{app:hard_heterogeneous_setup}}
\@dottedtocline{1}{0em}{2.4em}{\hyperref[app:baseline_details]{\ref*{app:baseline_details}\quad Baseline Details}}{\pageref*{app:baseline_details}}
\@dottedtocline{1}{0em}{2.4em}{\hyperref[app:hyperparameters]{\ref*{app:hyperparameters}\quad Hyperparameters}}{\pageref*{app:hyperparameters}}
\@dottedtocline{2}{1.6em}{3.2em}{\hyperref[app:pretrain_hparams]{\ref*{app:pretrain_hparams}\quad Offline Pretraining}}{\pageref*{app:pretrain_hparams}}
\@dottedtocline{2}{1.6em}{3.2em}{\hyperref[app:online_hparams]{\ref*{app:online_hparams}\quad Online FedGuide Settings}}{\pageref*{app:online_hparams}}
\@dottedtocline{2}{1.6em}{3.2em}{\hyperref[app:variant_hparams]{\ref*{app:variant_hparams}\quad FedGuide Variants}}{\pageref*{app:variant_hparams}}
\@dottedtocline{2}{1.6em}{3.2em}{\hyperref[app:baseline_hparams]{\ref*{app:baseline_hparams}\quad Baseline Hyperparameters}}{\pageref*{app:baseline_hparams}}
\@dottedtocline{1}{0em}{2.4em}{\hyperref[app:additional]{\ref*{app:additional}\quad Additional Analysis}}{\pageref*{app:additional}}
\@dottedtocline{2}{1.6em}{3.2em}{\hyperref[app:sensitivity]{\ref*{app:sensitivity}\quad Sensitivity to $M$, $\eta$, and $\lambda_2$}}{\pageref*{app:sensitivity}}
\@dottedtocline{2}{1.6em}{3.2em}{\hyperref[app:weak_overlap]{\ref*{app:weak_overlap}\quad Behavior under Weak Overlap}}{\pageref*{app:weak_overlap}}
\@dottedtocline{1}{0em}{2.4em}{\hyperref[app:visualizations]{\ref*{app:visualizations}\quad Visualizations}}{\pageref*{app:visualizations}}
\makeatother
\endgroup

\section{Theoretical Proofs}
\label{app:proofs}

We provide the detailed assumptions and proofs for the theoretical
claims in Section~\ref{sec:theory}. The goal is not to establish global
optimality, but to justify three design choices in FedGuide: using the DICE
value as a state-only baseline, aggregating only diffusion priors through
OT-MoE, and keeping policies local under client heterogeneity.

\subsection{Assumptions}
\label{app:assumptions}

\begin{description}
\item[A1 (Smoothness).]
The federated surrogate $\tilde F(\psi,\theta)$ is $L$-smooth in the prior
parameters $\psi$ and local policy parameters $\theta=\{\theta_i\}_{i=1}^{N}$. The diffusion-prior
parametrization $\psi\mapsto \pi_{D,\psi}$ is $L_{\psi}$-Lipschitz in score.

\item[A2 (Bounded stochastic gradient noise).]
The stochastic policy gradient noise has variance bounded by $\varsigma^2$.
After applying the DICE value baseline, the per-update effective variance is
denoted by $\varsigma_{\mathrm{eff}}^2$; averaging over $N$ clients and $U$
local updates yields the $1/(NU)$ scaling in the convergence bound. Let
$g^{(h)}$ denote the aggregated stochastic descent direction for
$x^{(h)}=(\psi^{(h)},\theta^{(h)})$. Conditional on the history
$\mathcal{F}_h$, write
\begin{equation}
g^{(h)}
=
\nabla \tilde F(x^{(h)})+\xi^{(h)}+b^{(h)},
\quad
\mathbb{E}[\xi^{(h)}\mid\mathcal{F}h]=0,
\end{equation}
where $\xi^{(h)}$ is the zero-mean stochastic gradient noise after client and
local update averaging, and $b^{(h)}$ is the bias induced by the practical
surrogate approximation. We then assume
\begin{equation}
\mathbb{E}\left[|\xi^{(h)}|^2\mid\mathcal{F}h\right]
\le
\frac{\varsigma{\mathrm{eff}}^2}{NU}.
\end{equation}
The practical self-normalized importance-sampling prior surrogate has bounded bias $\epsilon_{\mathrm{SNIS}}$.

\item[A3 (Bounded local KL drift).]
Each local policy update satisfies
\begin{equation}
D_{\mathrm{KL}}
\!\left(
\pi_{\theta,i}^{(h+1)}
\;\middle\|\;
\pi_{\theta,i}^{(h)}
\right)
\le \delta .
\end{equation}

\item[A4 (Informative bounded DICE baseline).]
The DICE value baseline is bounded, $|V_{\phi,i}(s)|\le B_V$. Let $G_i^t$
denote the return target used in advantage estimation. The residual correlation
between the online-baseline error and the DICE-baseline correction is
\begin{equation}
\kappa_i
=
\mathrm{Corr}
\!\left(
G_i^t - V_{\theta,i}(s^t),
\;
V_{\phi,i}(s^t)-V_{\theta,i}(s^t)
\right).
\end{equation}
Any additional bias from using $V_{\phi,i}$ inside bootstrapped GAE is bounded
by $\epsilon_V$.

\item[A5 (Bounded OT-MoE approximation errors).]
The Sinkhorn solution $\hat Y$ is $\epsilon_{\mathrm{OT}}$-suboptimal in the
entropic OT objective relative to the exact optimizer $Y^\star$. The
implementation's parameter-space cost proxy differs from the ideal score-MSE
cost by at most $\epsilon_C$. Together with A2--A4, these approximation terms
bound the squared descent-direction bias as
\begin{equation}
\|b^{(h)}\|^2
\le
\left(
\epsilon_{\mathrm{OT}}+\epsilon_C+\epsilon_{\mathrm{SNIS}}+\epsilon_V+\delta
\right).
\end{equation}
\end{description}

\paragraph{On the role of assumptions.}
The assumptions above are regularity conditions rather than claims that require
proof inside the paper. A1--A2 are standard smoothness and stochastic-approximation
conditions for nonconvex policy gradient analysis; A3 is enforced in practice by
the local trust-region KL control; A4 states when the DICE value is informative
as a control variate; and A5 records the controlled numerical and proxy errors
introduced by Sinkhorn OT, the parameter-space cost proxy, and the practical
prior/value surrogates. The lemmas and theorems below prove what follows once
these conditions hold.

\subsection{Variance Reduction from the Value Baseline}
\label{app:variance_reduction}

\paragraph{Lemma 1 (State-only baseline does not add score-function bias).}
For any state-only baseline $\mathbf{V}_i(s^t)$,
\begin{equation}
\label{eq:app-lemma1}
\mathbb{E}_{a^t\sim\pi_{\theta,i}(\cdot\mid s^t)}
\!\left[
\nabla_{\theta_i}\log \pi_{\theta,i}(a^t\mid s^t)\mathbf{V}_i(s^t)
\right]
=0.
\end{equation}
Therefore, using the blended baseline
\begin{equation}
b_i(s)=\beta V_{\theta,i}(s)+(1-\beta)V_{\phi,i}(s)
\end{equation}
does not introduce additional action-dependent bias into the policy-gradient
estimator. This statement concerns the score-function baseline term; finite
horizon GAE and imperfect value estimates are handled as approximation errors.

\paragraph{Proof.}
Since $b_i(s^t)$ does not depend on $a^t$,
$\int
\pi_{\theta,i}(a^t\mid s^t)
\,da^t$ = 1
\begin{align}
\mathbb{E}_{a^t\sim\pi_{\theta,i}}
\!\left[
\nabla_{\theta_i}\log \pi_{\theta,i}(a^t\mid s^t)\mathbf{V}_i(s^t)
\right]
&=
\mathbf{V}_i(s^t)
\int
\pi_{\theta,i}(a^t\mid s^t)
\nabla_{\theta_i}\log \pi_{\theta,i}(a^t\mid s^t)
\,da^t
\\
&=
\mathbf{V}_i(s^t)
\int
\nabla_{\theta_i}\pi_{\theta,i}(a^t\mid s^t)
\,da^t
\\
&=
\mathbf{V}_i(s^t)
\nabla_{\theta_i}
\int
\pi_{\theta,i}(a^t\mid s^t)
\,da^t
=\mathbf{V}_i(s^t)
\nabla_{\theta_i} 1 = 0.
\end{align}
Thus $V_{\phi,i}$ enters only as a baseline for centering the advantage, not as
an action-dependent policy-gradient term.
\hfill$\square$

\paragraph{Lemma 2 (Variance reduction).}
Let $G_i^t$ be the return target and define
\begin{equation}
X_i^t = G_i^t - V_{\theta,i}(s^t),
\quad
Z_i^t = V_{\phi,i}(s^t)-V_{\theta,i}(s^t).
\end{equation}
The blended advantage can be written as
\begin{equation}
\hat A_{\beta,i}^t
=
G_i^t-\beta V_{\theta,i}(s^t)-(1-\beta)V_{\phi,i}(s^t)
=
X_i^t-(1-\beta)Z_i^t .
\end{equation}
If $\mathrm{Cov}(X_i^t,Z_i^t)>0$, then there exists
$\beta^\star\in(0,1)$ such that
\begin{equation}
\label{eq:app-variance-reduction}
\mathrm{Var}\!\left[\hat A_{\beta^\star,i}^t\right]
<
\mathrm{Var}\!\left[\hat A_{\theta,i}^t\right],
\end{equation}
where $\hat A_{\theta,i}^t=G_i^t-V_{\theta,i}(s^t)$ is the online-baseline
advantage.

\paragraph{Proof.}
Let $\alpha=1-\beta$. Then
\begin{align}
\mathrm{Var}\!\left[\hat A_{\beta,i}^t\right]
&=
\mathrm{Var}\!\left[X_i^t-\alpha Z_i^t\right]
\\
&=
\mathrm{Var}[X_i^t]
-2\alpha\,\mathrm{Cov}(X_i^t,Z_i^t)
+\alpha^2\mathrm{Var}[Z_i^t].
\end{align}
If $\mathrm{Var}(Z_i^t)=0$, then
$\mathrm{Cov}(X_i^t,Z_i^t)=0$ by Cauchy's inequality, contradicting the
assumption $\mathrm{Cov}(X_i^t,Z_i^t)>0$. Hence $\mathrm{Var}(Z_i^t)>0$. The
variance difference relative to the online-baseline advantage is
\begin{equation}
\mathrm{Var}\!\left[\hat A_{\beta,i}^t\right]
-
\mathrm{Var}[X_i^t]
=
-2\alpha\,\mathrm{Cov}(X_i^t,Z_i^t)
+
\alpha^2\mathrm{Var}[Z_i^t].
\end{equation}
For any
\begin{equation}
0<\alpha<
\min
\left\{
1,\,
\frac{2\mathrm{Cov}(X_i^t,Z_i^t)}
{\mathrm{Var}(Z_i^t)}
\right\},
\end{equation}
this difference is strictly negative. Therefore there exists
$\beta=1-\alpha\in(0,1)$ such that
\eqref{eq:app-variance-reduction} holds. The variance is minimized at
\begin{equation}
\alpha^\star
=
\frac{\mathrm{Cov}(X_i^t,Z_i^t)}
{\mathrm{Var}(Z_i^t)}.
\end{equation}
If $\alpha^\star\in(0,1)$, then $\beta^\star=1-\alpha^\star\in(0,1)$ and
\begin{equation}
\mathrm{Var}\!\left[\hat A_{\beta^\star,i}^t\right]
=
\mathrm{Var}[X_i^t]
-
\frac{\mathrm{Cov}(X_i^t,Z_i^t)^2}
{\mathrm{Var}(Z_i^t)}
<
\mathrm{Var}[X_i^t].
\end{equation}
Using
\begin{equation}
\mathrm{Cov}(X_i^t,Z_i^t)^2
=
\kappa_i^2\,
\mathrm{Var}(X_i^t)\,
\mathrm{Var}(Z_i^t),
\end{equation}
the clientwise optimal correction gives
\begin{equation}
\mathrm{Var}\!\left[\hat A_{\beta^\star,i}^t\right]
=
\mathrm{Var}(X_i^t)(1-\kappa_i^2).
\end{equation}
For a fixed blending coefficient, the same form holds with a coefficient
$c_{\beta,i}>0$ whenever the chosen $\beta$ lies in the variance reducing
interval above:
\begin{equation}
\mathrm{Var}\!\left[\hat A_{\beta,i}^t\right]
\le
\mathrm{Var}(X_i^t)(1-c_{\beta,i}\kappa_i^2).
\end{equation}
Taking $c_\beta=\min_i c_{\beta,i}$ over the finite client set and averaging
with client aggregation weights $\zeta_i$ satisfying $\sum_i\zeta_i=1$ yields the effective variance form used in the main text:
\begin{equation}
\label{eq:app-sigma-eff}
\varsigma_{\mathrm{eff}}^2
\le
\varsigma^2
\left(1-c_\beta\bar{\kappa}^2\right),
\quad
\bar{\kappa}^2=\sum_{i=1}^{N}\zeta_i\kappa_i^2,
\end{equation}
where $c_\beta>0$ depends on the chosen blending coefficient. Therefore, an
informative DICE value baseline acts as a control variate for the return target.
\hfill$\square$

\subsection{Federated Convergence}
\label{app:federated_convergence}

\paragraph{Theorem 1 (Convergence of FedGuide).}
Under A1--A5, with policy-optimization stepsize $\eta_{\mathrm{pg}}\le 1/(4L)$, after $H$ communication rounds and
$U$ local updates per round, FedGuide satisfies
\begin{equation}
\label{eq:app-thm1}
\frac{1}{H}\sum_{h=0}^{H-1}
\mathbb{E}
\bigl\|
\nabla \tilde F(\psi^{(h)},\theta^{(h)})
\bigr\|^2
\le
\mathcal{O}
\!\left(
\frac{\tilde F^{(0)}-\tilde F^\star}{\eta_{\mathrm{pg}} H}
\right)
+
\mathcal{O}
\!\left(
\eta_{\mathrm{pg}} L\frac{\varsigma_{\mathrm{eff}}^2}{NU}
\right)
+
\mathcal{O}
\!\left(
\Delta
\right).
\end{equation}
where 
\begin{math}
\Delta=\epsilon_{\mathrm{OT}}+\epsilon_C+\epsilon_{\mathrm{SNIS}}+\epsilon_V+\delta.
\end{math}
With a standard diminishing choice such as $\eta_{\mathrm{pg}}=\Theta(H^{-1/2})$, this gives
the usual nonconvex stationarity rate plus the controlled approximation terms.

\paragraph{Proof.}
Write $x^{(h)}=(\psi^{(h)},\theta^{(h)})$ and the update as
\begin{equation}
x^{(h+1)}
=
x^{(h)}-\eta_{\mathrm{pg}}g^{(h)} .
\end{equation}
By A1, $\tilde F$ is $L$-smooth, hence
\begin{align}
\tilde F(x^{(h+1)})
&\le
\tilde F(x^{(h)})
-
\eta_{\mathrm{pg}}
\left\langle
\nabla \tilde F(x^{(h)}),g^{(h)}
\right\rangle
+
\frac{L\eta_{\mathrm{pg}}^2}{2}
\|g^{(h)}\|^2 .
\end{align}
Using the decomposition in A2,
$g^{(h)}=\nabla\tilde F(x^{(h)})+\xi^{(h)}+b^{(h)}$, and conditioning on
$\mathcal F_h$, the zero-mean property
$\mathbb{E}[\xi^{(h)}\mid\mathcal F_h]=0$ gives
\begin{align}
\mathbb{E}\!\left[
\tilde F(x^{(h+1)})\mid\mathcal F_h
\right]
&\le
\tilde F(x^{(h)})
-
\eta_{\mathrm{pg}}
\|\nabla\tilde F(x^{(h)})\|^2
-
\eta_{\mathrm{pg}}
\left\langle
\nabla\tilde F(x^{(h)}),b^{(h)}
\right\rangle
\\
&\quad+
\frac{L\eta_{\mathrm{pg}}^2}{2}
\mathbb{E}\!\left[
\|\nabla\tilde F(x^{(h)})+\xi^{(h)}+b^{(h)}\|^2
\mid\mathcal F_h
\right].
\end{align}
We bound the bias inner product by Young's inequality,
\begin{equation}
-
\left\langle
\nabla\tilde F(x^{(h)}),b^{(h)}
\right\rangle
\le
\frac{1}{4}
\|\nabla\tilde F(x^{(h)})\|^2
+
\|b^{(h)}\|^2,
\end{equation}
and use
\begin{equation}
\|u+v+w\|^2
\le
3\|u\|^2+3\|v\|^2+3\|w\|^2 .
\end{equation}
Together with A2 and A5, this implies
\begin{align}
\mathbb{E}\!\left[
\tilde F(x^{(h+1)})\mid\mathcal F_h
\right]
&\le
\tilde F(x^{(h)})
-
\left(
\frac{3\eta_{\mathrm{pg}}}{4}
-
\frac{3L\eta_{\mathrm{pg}}^2}{2}
\right)
\|\nabla\tilde F(x^{(h)})\|^2
\\
&\quad+
C_1 L\eta_{\mathrm{pg}}^2
\frac{\varsigma_{\mathrm{eff}}^2}{NU}
+
C_2\eta_{\mathrm{pg}}\Delta,
\end{align}
for universal constants $C_1,C_2>0$. Since
$\eta_{\mathrm{pg}}\le 1/(4L)$, the coefficient of the gradient term is at
least $\eta_{\mathrm{pg}}/4$. Therefore,
\begin{equation}
\frac{\eta_{\mathrm{pg}}}{4}
\mathbb{E}
\|\nabla\tilde F(x^{(h)})\|^2
\le
\mathbb{E}\tilde F(x^{(h)})
-
\mathbb{E}\tilde F(x^{(h+1)})
+
C_1 L\eta_{\mathrm{pg}}^2
\frac{\varsigma_{\mathrm{eff}}^2}{NU}
+
C_2\eta_{\mathrm{pg}}\Delta .
\end{equation}
Summing this inequality from $h=0$ to $H-1$ telescopes the objective values:
\begin{align}
\frac{\eta_{\mathrm{pg}}}{4}
\sum_{h=0}^{H-1}
\mathbb{E}
\|\nabla\tilde F(x^{(h)})\|^2
&\le
\tilde F^{(0)}-\tilde F^\star
+
H C_1 L\eta_{\mathrm{pg}}^2
\frac{\varsigma_{\mathrm{eff}}^2}{NU}
+
H C_2\eta_{\mathrm{pg}}\Delta .
\end{align}
Dividing by $H\eta_{\mathrm{pg}}/4$ yields
\begin{equation}
\frac{1}{H}
\sum_{h=0}^{H-1}
\mathbb{E}
\|\nabla\tilde F(x^{(h)})\|^2
\le
\mathcal{O}
\!\left(
\frac{\tilde F^{(0)}-\tilde F^\star}{\eta_{\mathrm{pg}}H}
\right)
+
\mathcal{O}
\!\left(
\eta_{\mathrm{pg}}L\frac{\varsigma_{\mathrm{eff}}^2}{NU}
\right)
+
\mathcal{O}(\Delta),
\end{equation}
which is Eq.~\eqref{eq:app-thm1}.
\hfill$\square$

\paragraph{Theorem 2 (Linear convergence under PL).}
Suppose, in addition to A1--A5, that $\tilde F$ satisfies the
Polyak--{\L}ojasiewicz (PL) condition
\begin{equation}
\label{eq:app-pl}
\tilde F(x)-\tilde F^\star
\le
c_{\mathrm{PL}}
\|\nabla \tilde F(x)\|^2 .
\end{equation}
Then FedGuide converges linearly to a neighborhood of the optimum:
\begin{equation}
\label{eq:app-thm2}
\mathbb{E}
\!\left[
\tilde F(x^{(h)})-\tilde F^\star
\right]
\le
(1-\mu)^h
\left(
\tilde F^{(0)}-\tilde F^\star
\right)
+
\mathcal{O}
\!\left(
\frac{\eta_{\mathrm{pg}} L\varsigma_{\mathrm{eff}}^2}{NU}
+
\Delta
\right),
\end{equation}
where $\mu=\eta_{\mathrm{pg}}/(4c_{\mathrm{PL}})$. Thus the DICE value baseline reduces the
stochastic error floor through $\varsigma_{\mathrm{eff}}^2$, while OT-MoE,
the prior or value surrogates, and the local trust region contribute bounded
approximation terms.

\paragraph{Proof.}
Let
\begin{math}
R_h
=
\mathbb{E}
\left[
\tilde F(x^{(h)})-\tilde F^\star
\right].
\end{math}
The proof of Theorem~1 establishes the following one-step descent bound:
\begin{equation}
R_{h+1}
\le
R_h
-
\frac{\eta_{\mathrm{pg}}}{4}
\mathbb{E}
\|\nabla \tilde F(x^{(h)})\|^2
+
C_1L\eta_{\mathrm{pg}}^2
\frac{\varsigma_{\mathrm{eff}}^2}{NU}
+
C_2\eta_{\mathrm{pg}}\Delta .
\end{equation}
By the PL condition,
\begin{equation}
\mathbb{E}
\|\nabla \tilde F(x^{(h)})\|^2
\ge
\frac{1}{c_{\mathrm{PL}}}
\mathbb{E}
\left[
\tilde F(x^{(h)})-\tilde F^\star
\right]
=
\frac{1}{c_{\mathrm{PL}}}R_h .
\end{equation}
Substituting this into the one-step inequality gives
\begin{equation}
R_{h+1}
\le
\left(
1-\frac{\eta_{\mathrm{pg}}}{4c_{\mathrm{PL}}}
\right)R_h
+
C_1L\eta_{\mathrm{pg}}^2
\frac{\varsigma_{\mathrm{eff}}^2}{NU}
+
C_2\eta_{\mathrm{pg}}\Delta .
\end{equation}
Let $\mu=\eta_{\mathrm{pg}}/(4c_{\mathrm{PL}})$ and
\begin{equation}
A
=
C_1L\eta_{\mathrm{pg}}^2
\frac{\varsigma_{\mathrm{eff}}^2}{NU}
+
C_2\eta_{\mathrm{pg}}\Delta .
\end{equation}
The recursion becomes
\begin{equation}
R_{h+1}\le (1-\mu)R_h+A .
\end{equation}
Unrolling it yields
\begin{align}
R_h
&\le
(1-\mu)^hR_0
+
A\sum_{r=0}^{h-1}(1-\mu)^r
\\
&\le
(1-\mu)^hR_0+\frac{A}{\mu}.
\end{align}
Since $1/\mu=4c_{\mathrm{PL}}/\eta_{\mathrm{pg}}$, the residual term satisfies
\begin{equation}
\frac{A}{\mu}
=
\mathcal{O}
\!\left(
\frac{\eta_{\mathrm{pg}}L\varsigma_{\mathrm{eff}}^2}{NU}
+
\Delta
\right),
\end{equation}
where constants absorb $c_{\mathrm{PL}}$. Replacing $\Delta$ by its definition
gives Eq.~\eqref{eq:app-thm2}.
\hfill$\square$

\subsection{Policy Averaging under Prior-Regularized Alignment (FedGuide-A)}
\label{app:policy-avg}

\paragraph{Theorem 3 (Controlled policy-averaging error).}
Let $\bar{\pi}_{\theta}^{(h)}$ denote the averaged policy obtained from local
policies $\{\pi_{\theta_i}^{(h)}\}_{i=1}^{N}$ at communication round $h$. Define
the policy-dispersion term
\begin{equation}
\label{eq:policy-dispersion}
\Gamma_{\pi}^{(h)}
=
\sum_{i=1}^{N}\zeta_i\,
\mathbb{E}_{s\sim d_i^{(h)}}\!
\left[
D_{\mathrm{KL}}
\bigl(
\pi_{\theta_i}^{(h)}(\cdot\mid s)
\;\|\;
\bar{\pi}_{\theta}^{(h)}(\cdot\mid s)
\bigr)
\right],
\end{equation}
where $\zeta_i$ is the client aggregation weight and $d_i^{(h)}$ is the state distribution
induced by client $i$. Assume the expected return is Lipschitz with respect to
the action distribution: for any policies $\pi,\pi'$,
\begin{equation}
\label{eq:return-lipschitz-tv}
\left|
J_i(\pi)-J_i(\pi')
\right|
\le
L_J
\mathbb{E}_{s\sim d_i^{(h)}}\!
\left[
D_{\mathrm{TV}}
\bigl(
\pi(\cdot\mid s),\pi'(\cdot\mid s)
\bigr)
\right].
\end{equation}
where $D_{\mathrm{TV}}$ denotes the total variation distance between two
action distributions at a fixed state,
$
D_{\mathrm{TV}}\left(\pi(\cdot\mid s),\pi'(\cdot\mid s)\right)
:=
\frac{1}{2}\int_{\mathcal A}
\left|\pi(a\mid s)-\pi'(a\mid s)\right|da,
$
with the integral replaced by a summation for discrete action spaces. Then
\begin{equation}
\label{eq:policy-avg-bound}
\left|
\sum_{i=1}^{N}\zeta_i J_i(\pi_{\theta_i}^{(h)})
-
\sum_{i=1}^{N}\zeta_i J_i(\bar{\pi}_{\theta}^{(h)})
\right|
\le
C_{\pi}\sqrt{\Gamma_{\pi}^{(h)}},
\end{equation}
for a constant $C_{\pi}>0$ depending on the reward range and the horizon.
Therefore, policy averaging has limited impact when the local policies remain
aligned in distribution.

\paragraph{Proof.}
For each client $i$, apply the Lipschitz condition
\eqref{eq:return-lipschitz-tv} with
$\pi=\pi_{\theta_i}^{(h)}$ and
$\pi'=\bar{\pi}_{\theta}^{(h)}$:
\begin{equation}
\left|
J_i(\pi_{\theta_i}^{(h)})
-
J_i(\bar{\pi}_{\theta}^{(h)})
\right|
\le
L_J
\mathbb{E}_{s\sim d_i^{(h)}}\!
\left[
D_{\mathrm{TV}}
\bigl(
\pi_{\theta_i}^{(h)}(\cdot\mid s),
\bar{\pi}_{\theta}^{(h)}(\cdot\mid s)
\bigr)
\right].
\end{equation}
Pinsker's inequality gives, for every state $s$,
\begin{equation}
D_{\mathrm{TV}}
\bigl(
\pi_{\theta_i}^{(h)}(\cdot\mid s),
\bar{\pi}_{\theta}^{(h)}(\cdot\mid s)
\bigr)
\le
\sqrt{
\frac{1}{2}
D_{\mathrm{KL}}
\bigl(
\pi_{\theta_i}^{(h)}(\cdot\mid s)
\;\|\;
\bar{\pi}_{\theta}^{(h)}(\cdot\mid s)
\bigr)
}.
\end{equation}
Taking expectation over $s\sim d_i^{(h)}$ and using Jensen's inequality,
\begin{align}
\mathbb{E}_{s\sim d_i^{(h)}}\!
\left[
D_{\mathrm{TV}}
\bigl(
\pi_{\theta_i}^{(h)}(\cdot\mid s),
\bar{\pi}_{\theta}^{(h)}(\cdot\mid s)
\bigr)
\right]
&\le
\sqrt{
\frac{1}{2}
\mathbb{E}_{s\sim d_i^{(h)}}\!
\left[
D_{\mathrm{KL}}
\bigl(
\pi_{\theta_i}^{(h)}(\cdot\mid s)
\;\|\;
\bar{\pi}_{\theta}^{(h)}(\cdot\mid s)
\bigr)
\right]
}.
\end{align}
Now average the return gaps with weights $\zeta_i$:
\begin{align}
&\left|
\sum_{i=1}^{N}\zeta_iJ_i(\pi_{\theta_i}^{(h)})
-
\sum_{i=1}^{N}\zeta_iJ_i(\bar{\pi}_{\theta}^{(h)})
\right|
\\
&\quad\le
\sum_{i=1}^{N}\zeta_i
\left|
J_i(\pi_{\theta_i}^{(h)})
-
J_i(\bar{\pi}_{\theta}^{(h)})
\right|
\\
&\quad\le
\frac{L_J}{\sqrt{2}}
\sum_{i=1}^{N}\zeta_i
\sqrt{
\mathbb{E}_{s\sim d_i^{(h)}}\!
\left[
D_{\mathrm{KL}}
\bigl(
\pi_{\theta_i}^{(h)}(\cdot\mid s)
\;\|\;
\bar{\pi}_{\theta}^{(h)}(\cdot\mid s)
\bigr)
\right]
}.
\end{align}
Since the weights satisfy $\sum_i \zeta_i=1$, Jensen's inequality for the concave
square-root function gives
\begin{align}
\sum_{i=1}^{N}\zeta_i
\sqrt{A_i}
&\le
\sqrt{
\sum_{i=1}^{N}\zeta_iA_i
}
=
\sqrt{\Gamma_{\pi}^{(h)}},
\end{align}
where
\begin{equation}
A_i
=
\mathbb{E}_{s\sim d_i^{(h)}}\!
\left[
D_{\mathrm{KL}}
\bigl(
\pi_{\theta_i}^{(h)}(\cdot\mid s)
\;\|\;
\bar{\pi}_{\theta}^{(h)}(\cdot\mid s)
\bigr)
\right].
\end{equation}
Therefore,
\begin{equation}
\left|
\sum_{i=1}^{N}\zeta_iJ_i(\pi_{\theta_i}^{(h)})
-
\sum_{i=1}^{N}\zeta_iJ_i(\bar{\pi}_{\theta}^{(h)})
\right|
\le
\frac{L_J}{\sqrt{2}}
\sqrt{\Gamma_{\pi}^{(h)}}.
\end{equation}
Setting $C_{\pi}=L_J/\sqrt{2}$ proves Eq.~\eqref{eq:policy-avg-bound}.
\hfill$\square$

\paragraph{Implication for FedGuide.}
The bound shows that FedGuide does not require forbidding policy aggregation.
Instead, policy averaging (FedGuide-A) is safe when $\Gamma_{\pi}^{(h)}$ is small. In
FedGuide, the OT-MoE diffusion prior aligns clients in distribution space while
preserving multiple behavior modes, and the DICE value baseline centers
return-aware advantage estimates within this support. As a result, the local policies
are encouraged toward compatible behavior-support regions, so averaging the policy can
perform similarly to keeping policies fully local.

\section{Environmental Setup}
\label{app:env_setup}
We provide detailed descriptions of the experimental environments used in this work. This section serves three purposes: 
(\textbf{1}) to introduce each benchmark and motivate its relevance to heterogeneous FRL; 
(\textbf{2}) to describe how client-level heterogeneity is instantiated in each environment; and 
(\textbf{3}) to specify the construction of the ``Hard'' environments used for robustness evaluation under stronger heterogeneity.

\subsection{Environment Description}
\label{app:env_detail}

We design the environments to evaluate heterogeneous FRL under progressively stronger forms of client mismatch. Across all environments, clients share the same state and action spaces within each benchmark family, but differ in at least one component of the local MDP, including the initial-state distribution, reward function, transition dynamics, goal configuration, morphology, object layout, or task scenario. This setup directly matches the FedGuide problem setting: local clients induce different state-action distributions, so naively averaging policies or parameters can collapse behavior modes or produce unsupported actions. FedGuide is therefore evaluated on whether its diffusion-prior aggregation can preserve heterogeneous behavior support, while its local value baseline supports return-aware policy improvement within each client's feasible distribution.

\paragraph{Bandit2D.}
Bandit2D is a two-dimensional continuous bandit environment used as a diagnostic toy case for visualizing distribution mismatch and mode preservation. The policy outputs a continuous point $\mathbf{x}\in\mathbb{R}^2$, and the reward landscape contains four isotropic Gaussian modes placed uniformly on the unit circle:
\begin{equation}
    r(\mathbf{x})
    =
    \max_{m\in\{1,\ldots,4\}}
    \exp\left(
    -\frac{\|\mathbf{x}-\mathbf{c}_m\|_2^2}{2\sigma^2}
    \right),
    \quad
    \sigma=0.2,
\end{equation}
where $\mathbf{c}_m$ denotes the center of the $m$-th reward peak. Each client draws local samples from an annular sector aligned with one reward mode: the radius is sampled uniformly by area over the annulus $r\in[0.7,1.3]$, and the angle uniformly over a $120^\circ$ sector centered on that mode, giving $30\%$ overlap between adjacent clients. Each client is therefore concentrated on one mode, while the union of client supports forms a connected ring rather than four isolated disks. Therefore, Bandit2D isolates the key failure mode of policy-space aggregation: averaging heterogeneous clients can collapse multiple modes into a single mode or place mass in low-density regions. In contrast, FedGuide should preserve all modes through OT-MoE aggregation of diffusion priors and recover shared prior support that reflects the multi-modal data distribution. Because that support is ring-shaped by construction, the aggregated global prior is expected to recover the ring rather than four isolated peaks.

\paragraph{Reacher.}
Reacher is a low-dimensional continuous-control goal-reaching task based on the standard OpenAI Gym Reacher environment. The agent controls a two-link planar arm through continuous torques and receives reward based on the distance between the fingertip and a target, together with a control penalty. We introduce heterogeneity by assigning different clients to different target positions and environment configurations. As a result, each client has a different reward landscape and different local state-action occupancy, although the observation and action dimensions remain shared. Reacher therefore tests whether a federated method can transfer useful reaching behavior across clients without forcing all clients toward the same target-conditioned policy. This environment mainly stresses reward and goal heterogeneity, and serves as a controlled setting between the Bandit2D toy problem and more complex contact-rich locomotion tasks.

\paragraph{MuJoCo Locomotion.}
We evaluate Hopper, Walker2D, and HalfCheetah from MuJoCo to test FedGuide under dynamics, morphology, contact, and gait heterogeneity. These tasks are more challenging than Reacher because the policy-induced distribution depends strongly on contact timing, balance, body dynamics, and long-horizon coordination. For each locomotion benchmark, clients share the same nominal observation and action spaces, but differ through client-specific environment parameters such as body mass, link inertia, joint damping, actuator strength, ground friction, target velocity, or reward coefficients. These variations produce different transition functions $P_i$ and rewards $r_i$, leading to distinct local gait distributions.

Hopper evaluates single-leg hopping with strong sensitivity to contact stability and falling. Small changes in mass, damping, or actuation can substantially alter the feasible hopping rhythm. Walker2D evaluates bipedal locomotion, where heterogeneous clients may prefer different walking speeds, contact phases, and balance strategies. HalfCheetah evaluates high-speed planar locomotion with larger action dimensionality and smoother forward dynamics, but still requires coordinated multi-joint gait patterns. Together, these three environments test whether FedGuide can preserve transferable locomotion priors while allowing each client to adapt to its own dynamics and reward structure. This is especially relevant to heterogeneous robotic deployments, where robots may share a broad skill class, such as locomotion, but differ in morphology, calibration, terrain interaction, or actuator response.

\paragraph{MetaWorld10.}
MetaWorld10 evaluates stronger task-geometry heterogeneity through robotic manipulation. We use ten MetaWorld manipulation tasks, including reaching, pushing, pick-place, door opening, drawer closing, button pressing, peg insertion, sweeping, basketball, and related object-interaction tasks. Each client corresponds to one manipulation task, so the clients differ not only in reward functions, but also in object layouts, contact modes, task geometry, and successful behavior distributions. Compared with Reacher and MuJoCo locomotion, MetaWorld10 introduces weaker overlap between local behavior supports: a policy useful for pushing may not directly match the contact sequence needed for peg insertion or door opening.

This environment is important for evaluating the limitation of simple parameter aggregation under heterogeneous task semantics. Since different clients may require different manipulation modes, averaging local policies can erase task-specific behaviors. FedGuide instead aggregates diffusion priors in distribution space, preserving multiple manipulation modes while broadcasting personalized priors back to clients. The DICE value baseline then supports return-aware policy improvement within each client's task-relevant behavior support.

\paragraph{Client and Expert Configuration.}
Unless otherwise specified, we set the number of server-side diffusion-prior experts equal to the number of clients, i.e., $M=N$; MetaWorld10 uses $N=10$ clients and $M=8$ prior experts. For Reacher and MuJoCo locomotion, we use $N=8$ heterogeneous clients. For the Bandit2D visualization, we use four clients corresponding to the four reward modes. Each client maintains its own policy, online critic, and frozen DICE value baseline. Only the diffusion-prior head is uploaded for server-side OT-MoE aggregation. This configuration ensures that performance improvements come from distribution-space prior alignment and local DICE-baseline policy improvement, rather than from sharing raw data or directly averaging client policies.

\subsection{Heterogeneity Setup}
\label{app:heterogeneous_setup}

We instantiate client heterogeneity through fixed metadata files, so every
method is evaluated on the same client assignment and the same local MDPs. The
metadata specifies only client-side environment parameters; no raw trajectory or
offline dataset is shared across clients. Within each benchmark family the
observation and action spaces remain shared, while the reward, transition,
initial-state, or task distribution changes across clients.

\paragraph{Bandit2D.}
We use four clients and four reward modes. The Gaussian centers are placed at
$(1,0)$, $(0,1)$, $(-1,0)$, and $(0,-1)$ with $\sigma=0.2$. Each client is
assigned to one center and draws $1000$ offline samples from the annular sector
around it, with radius sampled uniformly by area over $r\in[0.7,1.3]$ and angle
uniform over a $120^\circ$ window, so adjacent clients overlap by $30\%$. In the
federated rollouts, each client sees dense behavior near one mode, while the
union of clients forms a multi-modal behavior distribution supported on the
whole ring, which the diffusion prior should preserve.

\paragraph{Reacher.}
The main Reacher setting uses $N=8$ clients. Client goals are placed in eight
angular sectors centered at
$0^\circ,45^\circ,\ldots,315^\circ$ on an annular region with radius $0.18$,
radial width $0.06$, and angular width $30^\circ$. This gives each client a
different target region while keeping the target inside the reachable workspace.
We additionally perturb each client by adding a fixed two-dimensional action
noise vector, scaling the reward, and randomizing the initial joint angle. In
the evaluated split, the action-noise norm ranges from $0.265$ to $1.173$, the
reward scale ranges from $0.621$ to $1.332$, and the angle-noise range is
$[-0.132,0.188]$.

\paragraph{Hopper and Walker2D.}
Hopper and Walker2D use the same metadata design with $N=8$ clients and
environment-specific MuJoCo backbones. The dynamics heterogeneity cycles through
four presets: nominal, heavy, weak-actuator, and high-damping/friction. These
presets are implemented by scaling body mass, joint damping, ground friction,
and action gain, with small random jitter around each preset. Reward
heterogeneity cycles through speed, efficiency, and stability preferences by
changing the forward reward weight, control cost weight, and an optional
instability penalty. For the evaluated clients, mass scales range from $0.972$
to $1.462$, damping from $0.779$ to $1.367$, ground friction from $0.887$ to
$1.267$, and action gain from $0.818$ to $1.002$. The forward reward weights
range from $0.906$ to $1.095$, control cost weights from $9.83{\times}10^{-4}$
to $1.74{\times}10^{-3}$, and instability weights from $0$ to $0.0234$.

\paragraph{HalfCheetah.}
The main HalfCheetah setting uses $N=8$ clients. The dynamics presets include
nominal, heavy-body, weak-actuator, and high-damping and friction clients, while the
reward preferences include speed, efficiency, and stability. Compared with
Hopper and Walker2D, the HalfCheetah reward uses a larger control cost scale, so
the reported control weights are the actual HalfCheetah-v4 coefficients. Across
the evaluated clients, mass scales range from $0.959$ to $1.097$, damping from
$0.963$ to $1.156$, ground friction from $0.940$ to $1.183$, and action gain
from $0.850$ to $1.023$. The forward-reward weights range from $0.948$ to
$1.077$, control-cost weights from $0.0899$ to $0.1232$, and reset-noise scales
from $0.0885$ to $0.1125$. This setting does not add an instability penalty.

\paragraph{MetaWorld10.}
MetaWorld10 uses one manipulation task per client. The 10 clients correspond
to \texttt{reach-v3}, \texttt{push-v3}, \texttt{pick-place-v3},
\texttt{door-open-v3}, \texttt{drawer-close-v3},
\texttt{button-press-topdown-v3}, \texttt{peg-insert-side-v3},
\texttt{window-open-v3}, \texttt{sweep-v3}, and \texttt{basketball-v3}. The
heterogeneity is therefore task-semantic: clients differ in object layout,
contact sequence, success condition, and reward geometry, while the loader keeps
a shared observation and action interface. This setting produces weaker overlap
between client behavior supports than the parameter-perturbation settings above.

\begin{figure*}[t]
    \centering
    \includegraphics[width=0.99\textwidth]{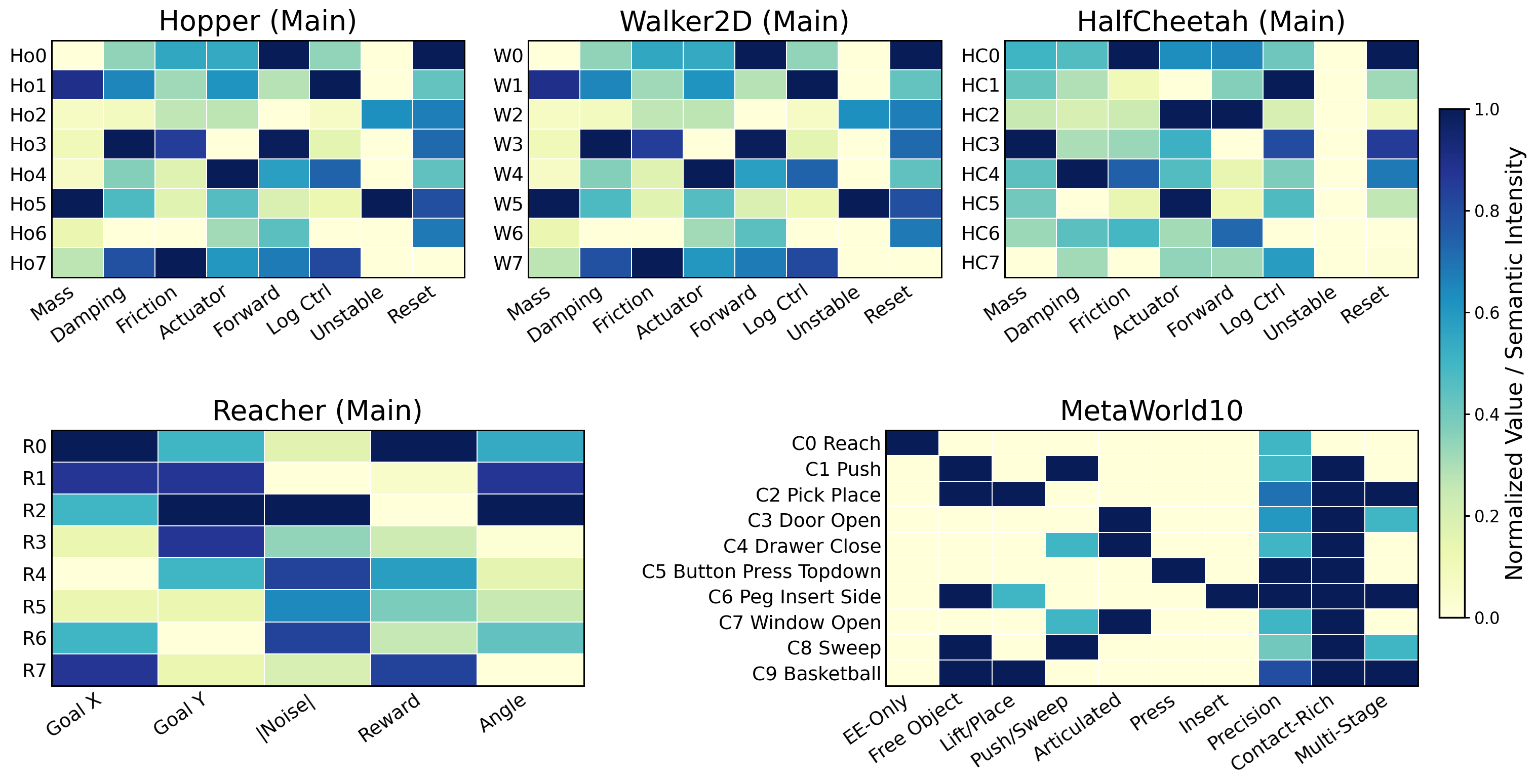}
    \caption{Client-level heterogeneity heatmaps for the main benchmark settings. Hopper, Walker2D, and HalfCheetah show normalized dynamics and reward metadata; Reacher shows normalized goal, action-noise, reward, and angle-shift metadata; MetaWorld10 shows task-semantic attributes.}
    \label{fig:app_heterogeneity_heatmaps}
\end{figure*}

Fig.~\ref{fig:app_heterogeneity_heatmaps} visualizes the source of heterogeneity in the main settings. For Hopper, Walker2D, HalfCheetah, and Reacher, each column is normalized within the corresponding environment so that the heatmap emphasizes relative client variation rather than raw physical units. For MetaWorld10, the heatmap uses semantic task attributes instead of continuous dynamics parameters, reflecting that its heterogeneity comes from manipulation task geometry, object interaction, contact mode, and multi-stage behavior. 

\subsection{Heterogeneity Setup for Stronger Environments}
\label{app:hard_heterogeneous_setup}

To further stress robustness, we construct harder environments for Reacher, Hopper,
Walker2D, and HalfCheetah. These ``Hard'' environments increase the separation
among clients by adding categorical reward preferences, structural perturbations,
or wider locomotion dynamics and reward-parameter ranges. The purpose is not
simply to make each individual MDP harder, but to increase cross-client
distribution mismatch. In these settings, the local policies are more likely to
drift toward incompatible regions of the policy space, making policy averaging
and shared parameter updates less reliable.

\paragraph{Reacher-Hard.}
Reacher-Hard keeps the eight annular goal sectors from the main Reacher setting,
but adds two stronger sources of heterogeneity. First, each client receives one
of four reward preferences: default, speed, stability, or efficiency. These
preferences change whether the local objective emphasizes reaching distance,
fingertip speed, joint stability, or control efficiency. Second, we introduce
structural heterogeneity by scaling the Reacher model mass, damping, and
actuator gear per client. In the evaluated split, mass ranges from $0.598$ to
$1.955$, damping from $0.568$ to $1.923$, and actuator gear from $0.587$ to
$1.954$. The action-noise norm ranges from $0.412$ to $1.178$, the reward scale
ranges from $0.698$ to $1.522$, and the angle-noise range is $[-0.192,0.074]$.

\paragraph{Hopper-Hard and Walker2D-Hard.}
Hopper-Hard and Walker2D-Hard use the hard locomotion setting. The four standard
dynamics presets are replaced by eight more separated regimes: nominal,
very-heavy, very-weak-actuator, extreme-damping/friction, low-friction,
heavy-low-damping, weak-low-friction, and inverted-dynamics. These regimes
jointly vary mass, damping, ground friction, and action gain, so clients require
different contact timing and actuation strategies. The reward preferences are
also pulled farther apart into sprint, miser, tightrope, and explorer modes.
Sprint increases the forward reward and lowers control cost, miser strongly
penalizes control, tightrope increases the instability penalty, and explorer
uses a moderate instability penalty with a higher forward-reward preference.
Across the evaluated clients, mass ranges from $0.862$ to $1.698$, damping from
$0.613$ to $1.767$, ground friction from $0.417$ to $1.476$, and action gain
from $0.620$ to $0.902$. The forward-reward weights range from $0.752$ to
$1.339$, control-cost weights from $6.95{\times}10^{-4}$ to
$3.83{\times}10^{-3}$, and instability weights from $0$ to $0.0339$.

\paragraph{HalfCheetah-Hard.}
HalfCheetah-Hard keeps the same categories of dynamics and reward preferences as
the main setting, but widens their perturbation ranges and introduces a nonzero
instability penalty. In the evaluated split, mass ranges from $0.943$ to
$1.078$, damping from $0.898$ to $1.204$, ground friction from $0.851$ to
$1.376$, and action gain from $0.670$ to $1.066$. The forward-reward weights
range from $0.906$ to $1.102$, control cost weights from $0.0919$ to $0.1391$,
instability weights from $0.0121$ to $0.0352$, and reset noise scales from
$0.0600$ to $0.1441$. Therefore, the Hard setting primarily increases
HalfCheetah's actuation, friction, reward-shaping, and stability mismatch.

\begin{figure*}[t]
    \centering
    \includegraphics[width=0.96\textwidth]{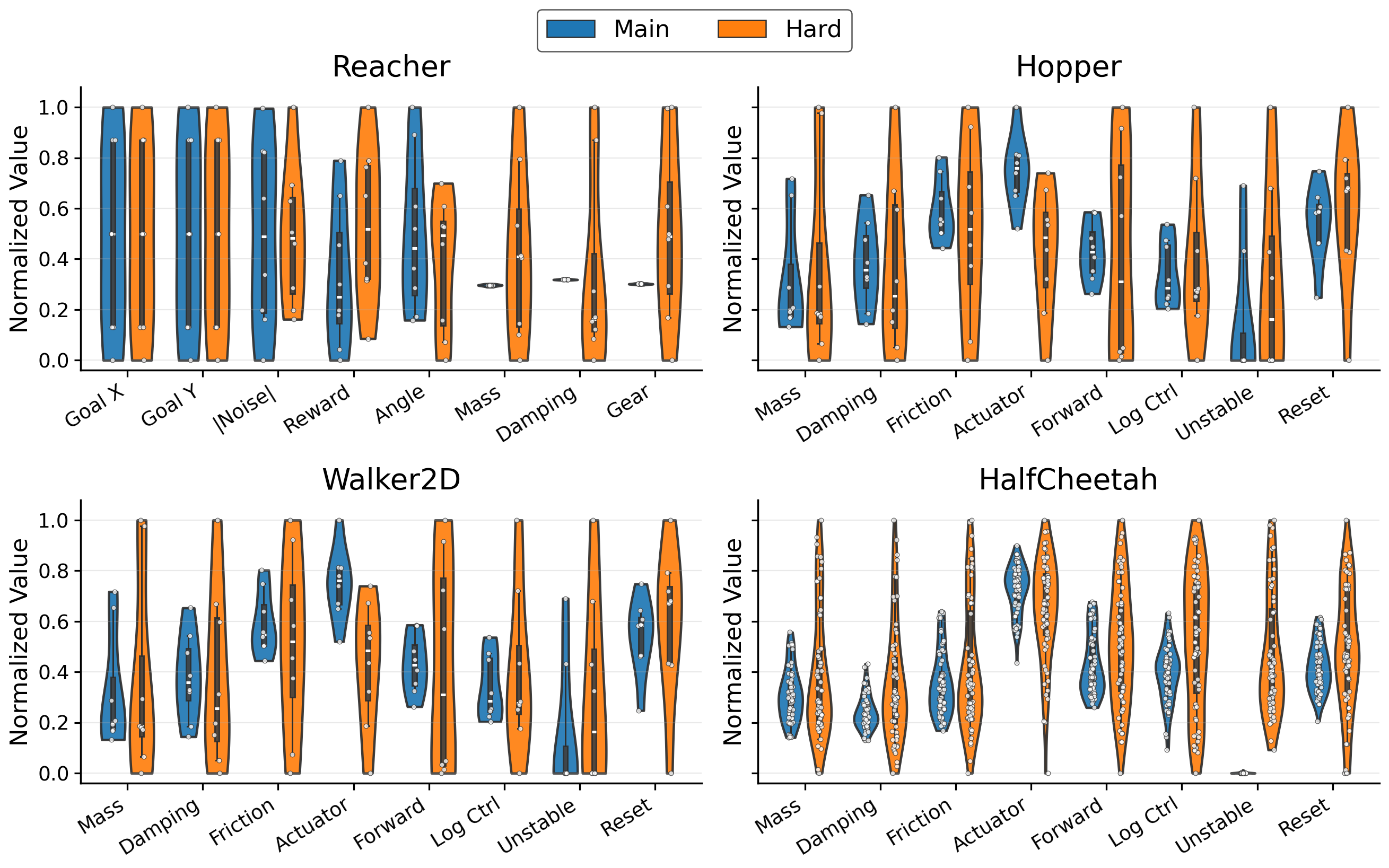}
    \caption{Main-to-Hard metadata range expansion. For each metadata dimension, client values are normalized by the pooled Main and Hard range and plotted as side-by-side Main and Hard distributions.}
    \label{fig:app_range_expansion}
    \vspace{-0.4cm}
\end{figure*}

Fig.~\ref{fig:app_range_expansion} compares each Hard setting against its corresponding main setting while preserving the identity of each metadata dimension. For every parameter, client values are normalized using the pooled Main and Hard range, so wider or more shifted Hard violins indicate expanded cross-client variation along that specific source of heterogeneity.

\section{Baseline Details}
\label{app:baseline_details}

We compare FedGuide with representative federated reinforcement learning baselines that resolve heterogeneity through parameter-space, policy-output, or policy-gradient alignment, rather than through distribution-space prior alignment. For a fair comparison, all methods use the same client splits, communication rounds, local rollout budget, evaluation protocol, and random seeds. No method shares raw trajectories or offline datasets across clients.

Tables~\ref{tab:method_attribute} and~\ref{tab:theory_mechanism} summarize the comparison at two levels. Table~\ref{tab:method_attribute} highlights the implementation-level mechanism: what each method aggregates, which space it aligns, and whether it preserves local policies and multi-modal support. Table~\ref{tab:theory_mechanism} connects these mechanisms to the type of theoretical or empirical guarantee associated with each method. The descriptions below unpack the rows of these two tables, while the optimizer-level hyperparameters are reported separately in Table~\ref{tab:app_baseline_hparams}.

\paragraph{FedAvg.}
FedAvg is the standard federated learning baseline based on iterative model averaging~\cite{mcmahan2017communication}. In each communication round, the server broadcasts the current global policy parameters $\bar{\theta}^{(h)}$ to all participating clients. Each client $i$ initializes from $\bar{\theta}^{(h)}$, collects local rollouts in its own MDP $\mathcal{M}_i$, and performs local policy optimization using the same PPO backbone as FedGuide but without any diffusion prior, OT-MoE aggregation, or DICE value baseline. After local training, the client uploads its updated policy parameters $\theta_i^{(h+1)}$ to the server, which computes
\begin{equation}
    \bar{\theta}^{(h+1)}
    =
    \sum_{i=1}^{N} \zeta_i \theta_i^{(h+1)},
    \quad
    \zeta_i = \frac{n_i}{\sum_{j=1}^{N} n_j},
\end{equation}
where $n_i$ is the number of local samples used by client $i$. In our experiments, all clients use the same local rollout budget, so $\zeta_i=1/N$ unless otherwise specified.

FedAvg is an important baseline because it represents the most direct policy-parameter aggregation strategy, corresponding to the ``Policy Params.'' and ``Parameter Space'' entries in Table~\ref{tab:method_attribute}. However, it does not explicitly account for policy-induced distribution mismatch across heterogeneous MDPs. When different clients learn policies adapted to different rewards, goals, dynamics, or task layouts, parameter averaging may combine incompatible local policies and produce a global policy that is not well supported by any client distribution.

\paragraph{FedKL.}
FedKL is a policy-gradient based FRL method designed to mitigate data heterogeneity by penalizing divergence in the policy-output distribution space~\cite{xie2023fedkl}. The original method shows that local policy improvement can benefit the global objective only when the local update is properly constrained with respect to the global policy. Instead of penalizing model parameters directly, FedKL constrains the divergence between policy distributions.

We instantiate FedKL with the same PPO local optimizer used by FedGuide. At the beginning of round $h$, the server broadcasts the global policy $\pi_{\bar{\theta}^{(h)}}$ to all clients. During local training, client $i$ optimizes a PPO-style objective augmented with two KL penalties:
\begin{equation}
    \mathcal{L}_{i}^{\mathrm{FedKL}}(\theta_i)
    =
    \mathcal{L}_{i}^{\mathrm{PPO}}(\theta_i)
    +
    \lambda_{\mathrm{g}}
    \widehat{D}_{\mathrm{KL}}
    \left(
        \pi_{\theta_i}(\cdot|s)
        \,\|\, 
        \pi_{\bar{\theta}^{(h)}}(\cdot|s)
    \right)
    +
    \lambda_{\mathrm{l}}
    \widehat{D}_{\mathrm{KL}}
    \left(
        \pi_{\theta_i}(\cdot|s)
        \,\|\,
        \pi_{\theta_i^{\mathrm{old}}}(\cdot|s)
    \right).
\end{equation}
The first term is the global KL penalty, which discourages the local policy from drifting too far away from the server policy. The second term is the local KL penalty, which constrains the policy change between consecutive local updates. Following the original FedKL formulation, these penalties act on the policy outputs rather than on the raw network parameters.

FedKL is stronger than FedAvg under heterogeneous clients because it explicitly limits policy drift, matching the ``Policy Output'' alignment entry in Table~\ref{tab:method_attribute}. However, FedKL still maintains a single aggregated global policy. It does not model heterogeneous client behavior distributions with a multi-modal prior, and therefore can still struggle when different clients require distinct behavior modes.

\begin{table*}[tb!]
\small
\newcolumntype{O}{>{          \arraybackslash}m{0.35cm}}
\newcolumntype{A}{>{          \arraybackslash}m{2.7cm}}
\newcolumntype{B}{>{\centering\arraybackslash}m{2.5cm}}
\newcolumntype{C}{>{\centering\arraybackslash}m{2.5cm}}
\newcolumntype{D}{>{\centering\arraybackslash}m{2.5cm}}
\newcolumntype{E}{>{\centering\arraybackslash}m{1.5cm}}
\newcolumntype{F}{>{\centering\arraybackslash}m{1.5cm}}
\newcolumntype{G}{>{\centering\arraybackslash}m{1.5cm}}
\setlength\tabcolsep{-2pt}
\renewcommand{\arraystretch}{1.5}
\newcommand\B{\rule[-0.9ex]{0pt}{0pt}}
\newcommand{\midsepremove}{\aboverulesep = 0mm \belowrulesep = 0mm}
\newcommand{\midsepdefault}{\aboverulesep = 0.605mm \belowrulesep = 0.984mm}
\begin{center}
\begin{tabular}{O|ABCDEFG}
\toprule
\multicolumn{2}{c}{\textbf{Method}} 
& {\textbf{Heterogeneity} \smash{$\hphantom{^{X}}$} \textbf{Type}}
& {\textbf{Aggregation} \smash{$\hphantom{^{X}}$} \textbf{Target}}
& {\textbf{Alignment} \smash{$\hphantom{^{X}}$} \textbf{Space}}
& {\textbf{Multi-Modal} \smash{$\hphantom{^{X}}$} \textbf{Support}}
& {\textbf{Local} \smash{$\hphantom{^{X}}$} \textbf{Policy}}
& {\textbf{Return-Aware} \newline \textbf{Baseline}}\\
\midrule

\parbox[t]{2mm}{\multirow{4}{*}{\rotatebox[origin=c]{90}
{\textbf{Baselines}\hspace{-0.8em}}}}
& ~~FedAvg~\cite{mcmahan2017communication}
& Data
& Policy Params.
& Parameter Space
& \xm
& \xm
& \xm\\

&~~FedKL~\cite{xie2023fedkl}
& Data
& Policy Params.
& Policy Output
& \xm
& \xm
& \pc\\

&~~FedRL~\cite{jin2022federated}
& MDP / Dynamics
& Actor + Critic
& Parameter Space
& \xm
& \xm
& \pc\\

&~~FedSVRPG-M~\cite{wang2024momentum}
& MDP / Dynamics
& Policy Grad.
& Gradient Space
& \xm
& \xm
& \pc\\

\midrule

\parbox[t]{2mm}{\multirow{3}{*}{\rotatebox[origin=c]{90}
{\textbf{Variants}\hspace{-0.5em}}}}
&~~FedGuide-P
& Data / MDP / Task
& Diffusion Prior
& Distribution Space
& \cm
& \cm
& \xm\\

&~~FedGuide-A
& Data / MDP / Task
& Prior + Policy
& Distribution Space
& \cm
& \xm
& \cm\\

\cmidrule{2-8}

&~~\textbf{FedGuide}
& Data / MDP / Task
& Diffusion Prior
& Distribution Space
& \cm
& \cm
& \cm\\

\bottomrule
\end{tabular}
\caption{Comparison of the baselines and FedGuide variants used in our experiments. FedAvg aggregates policy parameters, FedKL regularizes local policies through KL penalties in the policy-output space, FedRL aggregates actor-critic models, and FedSVRPG-M aggregates variance-reduced policy-gradient updates. FedGuide instead aggregates diffusion behavior priors in distribution space while keeping each client's policy local. FedGuide-P removes the DICE value baseline, and FedGuide-A additionally averages client policies. Here, \cm ~denotes explicit support, \xm ~denotes no explicit support, and \pc ~denotes partial or indirect support.}
\label{tab:method_attribute}
\end{center}
\vspace{-0.4cm}
\end{table*}

\begin{table*}[tb!]
\small
\newcolumntype{O}{>{          \arraybackslash}m{0.35cm}}
\newcolumntype{A}{>{          \arraybackslash}m{2.85cm}}
\newcolumntype{B}{>{\centering\arraybackslash}m{3cm}}
\newcolumntype{C}{>{\centering\arraybackslash}m{2.5cm}}
\newcolumntype{D}{>{\centering\arraybackslash}m{1.85cm}}
\newcolumntype{G}{>{\centering\arraybackslash}m{3.75cm}}
\setlength\tabcolsep{-2pt}
\renewcommand{\arraystretch}{1.5}
\newcommand\B{\rule[-0.9ex]{0pt}{0pt}}
\begin{center}
\begin{tabular}{O|ABCDG}
\toprule
\multicolumn{2}{c}{\textbf{Method}} 
& {\textbf{Convergence}}
& {\textbf{Communication} \smash{$\hphantom{^{X}}$} \textbf{Speedup}}
& {\textbf{Local Updates} \smash{$\hphantom{^{X}}$} \textbf{Allowed}}
& {\textbf{Distribution-Level} \smash{$\hphantom{^{X}}$} \textbf{Analysis}}\\
\midrule

\parbox[t]{2mm}{\multirow{4}{*}{\rotatebox[origin=c]{90}
{\textbf{Baselines}\hspace{-1em}}}}
&~~FedAvg~\cite{mcmahan2017communication}
& FL convergence
& Empirical
& \cm
& \xm \\

&~~FedKL~\cite{xie2023fedkl}
& Asymptotic
& Not claimed
& \pc
& Policy-output KL \\

&~~FedRL~\cite{jin2022federated}
& Finite to subopt.
& Not claimed
& \cm
& \xm \\

&~~FedSVRPG-M~\cite{wang2024momentum}
& Variance-reduced PG
& Claimed
& \cm
& Policy-gradient correction \\

\midrule

\parbox[t]{2mm}{\multirow{4}{*}{\rotatebox[origin=c]{90}
{\textbf{Variants}\hspace{-0.8em}}}}
&~~FedGuide-P
& Prior-only ablation
& Not isolated
& \cm
& OT-MoE diffusion prior \\

&~~FedGuide-A
& Controlled avg. error
& Not isolated
& \cm
& OT-MoE prior + policy avg. \\

\cmidrule{2-6}

&~~\textbf{FedGuide}
& Finite to controlled error
& Variance-controlled
& \cm
& OT-MoE diffusion prior \\

\bottomrule
\end{tabular}
\caption{Theoretical and mechanism-level comparison of the baselines and FedGuide variants. FedAvg provides the standard federated policy averaging baseline, FedKL provides convergence analysis for KL-regularized policy-output alignment, FedRL analyzes federated RL under environment heterogeneity with convergence to a heterogeneity dependent suboptimal solution, and FedSVRPG-M uses momentum-based variance-reduced policy-gradient correction for efficient heterogeneous FRL. FedGuide keeps policies local and aggregates only diffusion-prior heads through OT-MoE; its analysis separates the effective stochastic variance, OT errors, prior or value surrogate errors, and local KL drift, while the DICE value baseline reduces the variance of local policy updates.}
\label{tab:theory_mechanism}
\end{center}
\vspace{-0.4cm}
\end{table*}

\paragraph{FedRL.}
We use FedRL to denote the deep continuous-control federated RL baseline following the DDPGAvg extension of Jin et al.~\cite{jin2022federated}. The original work studies environment heterogeneity in FRL, where multiple agents interact with environments that share state and action spaces but differ in transition dynamics. It proposes QAvg and PAvg in the tabular setting and extends them to DQNAvg and DDPGAvg for deep RL. Since our benchmark tasks use continuous actions, we implement the DDPGAvg-style variant.

In each communication round, every client maintains a local DDPG agent with actor parameters $\theta_i^\mu$ and critic parameters $\theta_i^Q$. Client $i$ interacts with its own environment, stores transitions in a local replay buffer, and performs off-policy actor-critic updates. After local updates, the server aggregates both actor and critic parameters:
\begin{equation}
    \bar{\theta}^{\mu,(h+1)}
    =
    \sum_{i=1}^{N} \zeta_i \theta_i^{\mu,(h+1)},
    \quad
    \bar{\theta}^{Q,(h+1)}
    =
    \sum_{i=1}^{N} \zeta_i \theta_i^{Q,(h+1)}.
\end{equation}
The averaged actor and critic are then broadcast back to clients for the next round. No local replay buffer or trajectory data is shared.

FedRL provides an off-policy actor-critic counterpart to the on-policy FedAvg and FedKL baselines. This corresponds to the ``Actor + Critic'' aggregation target in Table~\ref{tab:method_attribute}. It can benefit from replay-buffer reuse and actor-critic learning, but it still relies on direct network averaging. Therefore, it does not explicitly preserve multiple heterogeneous behavior modes and can be sensitive when client dynamics, rewards, or task layouts induce incompatible local state-action distributions.

\paragraph{FedSVRPG-M.}
FedSVRPG-M is a momentum-based federated policy-gradient method for collaborative FRL across heterogeneous environments~\cite{wang2024momentum}. Instead of averaging local policy parameters after local training, each client estimates a stochastic variance-reduced policy-gradient update around the broadcast policy and sends the resulting update direction to the server. The server then updates a shared global policy using the aggregated direction, while the momentum term reduces the variance of local policy-gradient estimation.

In our experiments, FedSVRPG-M uses the same client splits, rollout budget, behavior-cloning warm-up, and evaluation protocol as the other baselines. The implementation follows the FedSVRPG-M communication pattern, but uses the practical PPO/SVRPG client-gradient path used in our continuous-control codebase. This keeps the comparison aligned with the common training backbone while still testing the main FedSVRPG-M design: collaboration through variance-reduced policy-gradient updates. Unlike FedGuide, FedSVRPG-M does not aggregate behavior priors, does not build a personalized distribution mixture for each client, and synchronizes learning through a shared policy-update direction. Thus, it can improve communication and interaction-level optimization, but it does not explicitly preserve separated state-action behavior modes.

\begin{table*}[t]
\centering
\small
\setlength{\tabcolsep}{3.5pt}
\renewcommand{\arraystretch}{1.08}
\begin{tabular}{p{0.23\linewidth}p{0.1\linewidth}p{0.13\linewidth}p{0.14\linewidth}p{0.16\linewidth}p{0.15\linewidth}}
\toprule
Parameter & Symbol & Bandit2D & Reacher & MuJoCo locomotion & MetaWorld10 \\
\midrule
Offline buffer size & $|\mathcal{D}_i|$ & $10^3$ & $2{\times}10^4$ & $2{\times}10^5$ cap & $5{\times}10^3$ \\
Offline source & -- & local modes & client rollouts & D4RL & scripted policy \\
Diffusion steps & $K$ & Gaussian & $1000$ & $1000$ & $1000$ \\
Prior width / horizon & $d_D,L_D$ & Gaussian & $64,64$ & $64,64$ & $64,64$ \\
Prior epochs / batch & $E_D,B_D$ & closed form & $40,512$ & $40,512$ & $40,512$ \\
Prior learning rate & $\eta_D$ & -- & $10^{-4}$ & $10^{-4}$ & $10^{-4}$ \\
DICE width & $d_{\phi}$ & $256$ & $256$ & $256$ & $256$ \\
DICE learning rates & $\eta_{\phi},\eta_w$ & $10^{-4},10^{-4}$ & $10^{-4},10^{-4}$ & $10^{-4},10^{-4}$ & $10^{-4},10^{-4}$ \\
DICE temperature & $\omega$ & $0.5$ & $0.5$ & $0.5$ & $0.5$ \\
DICE warm-up epochs & $E_{\phi}$ & $100$ & $80$ & $80$ & $80$ \\
BC epochs / batch & $E_{\mathrm{BC}},B_{\mathrm{BC}}$ & prior init & $40,512$ & $40,512$ & $40,512$ \\
BC width / learning rate & $d_{\mathrm{BC}},\eta_{\mathrm{BC}}$ & prior init & $256,3{\times}10^{-4}$ & $256,3{\times}10^{-4}$ & $256,3{\times}10^{-4}$ \\
\bottomrule
\end{tabular}
\caption{Offline pretraining hyperparameters.}
\label{tab:app_pretrain_hparams}
\end{table*}

\paragraph{FedGuide-P.}
FedGuide-P is an ablation of FedGuide that keeps the OT-MoE diffusion-prior aggregation but removes the DICE value baseline. Each client still receives a personalized diffusion prior $\bar{\pi}_{D,i}$ from the server and uses the prior regularizer in local policy learning:
\begin{equation}
    \lambda_2
    \widehat{\mathcal R}_{\mathrm{prior}}(\theta_i;\bar{\pi}_{D,i}).
\end{equation}
However, its advantage estimator uses only the online critic, which corresponds to setting $\beta=1$ in the blended value baseline. This variant tests whether distribution-space prior alignment alone is sufficient. This is why Table~\ref{tab:method_attribute} marks FedGuide-P as preserving multi-modal support and local policies, but not using the return-aware DICE baseline.

\paragraph{FedGuide-A.}
FedGuide-A keeps the OT-MoE diffusion prior and the local DICE value baseline, but additionally averages client policies at the server. This variant is included to evaluate whether policy averaging is harmful after client policies have already been aligned by personalized diffusion priors. Specifically, each client performs the same prior-regularized and value-baseline-blended local update as FedGuide, uploads its policy parameters, and the server computes an averaged policy:
\begin{equation}
    \bar{\theta}^{(h+1)}
    =
    \sum_{i=1}^{N} \zeta_i \theta_i^{(h+1)}.
\end{equation}
The averaged policy is then used to synchronize clients in the next communication round.

This ablation separates two design choices in FedGuide: distribution-space prior aggregation and keeping policies fully local. In Table~\ref{tab:method_attribute}, this is reflected by the ``Prior + Policy'' aggregation target and the absence of a fully local client policy. If FedGuide-A performs close to FedGuide, it suggests that OT-MoE prior alignment can reduce policy dispersion enough to make policy averaging less destructive. If it performs worse, it indicates that even with aligned priors, directly averaging client policies can still remove useful client-specific adaptation.

\paragraph{FedGuide.}
FedGuide is our proposed method. Unlike FedAvg, FedKL, FedRL, and FedSVRPG-M, FedGuide does not aggregate local policies as the main collaboration mechanism. Instead, each client keeps its policy $\pi_{\theta_i}$, online critic $V_{\theta,i}$, and frozen DICE value baseline $V_{\phi,i}$ local. Only the diffusion-prior head is uploaded to the server. The server aggregates these prior heads through OT-MoE and broadcasts a personalized distribution-mixture prior $\bar{\pi}_{D,i}$ back to each client. Local policy learning is then regularized toward this personalized behavior prior and uses the blended value baseline
\begin{equation}
    \mathbf{V}_i(s)
    =
    \beta V_{\theta,i}(s)
    +
    (1-\beta)V_{\phi,i}(s).
\end{equation}
Therefore, FedGuide differs from the baselines in two aspects: it aligns heterogeneous clients in distribution space rather than policy-parameter or policy-gradient space, and it uses a local DICE value baseline for return-aware policy improvement without sharing raw data, policies, or value functions. These two choices correspond to the full-support row in Table~\ref{tab:method_attribute} and to the controlled-error decomposition summarized in Table~\ref{tab:theory_mechanism}.

\section{Hyperparameters}
\label{app:hyperparameters}
We showcase the hyperparameters that we use for different environments on the main page.

\subsection{Offline Pretraining}
\label{app:pretrain_hparams}

Each client first trains its own behavior prior and DICE value baseline from a
local offline buffer. The prior models the support of $\mathcal{D}_i$, while
$V_{\phi,i}$ is frozen after pretraining and used only as the state-only
baseline in Eq.~\eqref{eq:Vblend}.

\subsection{Online FedGuide Settings}
\label{app:online_hparams}

The online stage uses PPO locally at each client. The policy, online critic, and
DICE value baseline stay local; only diffusion-prior heads are communicated for
FedGuide and FedGuide-P. FedGuide-A additionally averages policies as specified
in Table~\ref{tab:app_variant_hparams}.

\begin{table*}[t]
\centering
\small
\setlength{\tabcolsep}{4pt}
\renewcommand{\arraystretch}{1.08}
\begin{tabular}{p{0.25\linewidth}p{0.12\linewidth}p{0.13\linewidth}p{0.13\linewidth}p{0.13\linewidth}p{0.14\linewidth}}
\toprule
Parameter & Symbol & Bandit2D & Reacher & MuJoCo locomotion & MetaWorld10 \\
\midrule
Clients & $N$ & $4$ & $8$ & $8$ & $10$ \\
Prior experts & $M$ & $4$ & $8$ & $8$ & $8$ \\
Federated rounds & $H$ & $60$ & $100$ & $100$ & $100$ \\
Rollout size / round & $T_{\mathrm{roll}}$ & $200$ & $2048$ & $4096$ & $4096$ \\
Local PPO epochs & $U$ & $4$ & $4$ & $4$ & $4$ \\
PPO mini-batch & $B$ & $64$ & $64$ & $64$ & $64$ \\
Trust-region weight & $\lambda_1$ & $0.05$ & $0.05$ & $0.05$ & $0.05$ \\
Prior regularizer weight & $\lambda_2$ & $0.5$ & $0.5$ & $0.5$ & $0.5$ \\
Prior annealing horizon & $H_{\lambda_2}$ & $60$ & $50$ & $50$ & $50$ \\
Sinkhorn regularization & $\eta$ & $0.05$ & $0.05$ & $0.05$ & $0.05$ \\
Initial log std. & $\log \sigma_0$ & $-1$ & $0$ & $0$ & $0$ \\
Evaluation episodes & $n_{\mathrm{eval}}$ & grid / return & $10$ & $10$ & $10$ \\
Action range & $\mathcal{A}$ & $[-1.5,1.5]$ & $[-1,1]$ & $[-1,1]$ & $[-1,1]$ \\
\bottomrule
\end{tabular}
\caption{Shared online hyperparameters for FedGuide variants.}
\label{tab:app_online_hparams}
\end{table*}

\subsection{FedGuide Variants}
\label{app:variant_hparams}

The three FedGuide variants differ only in how they use the DICE baseline and
whether policy parameters are averaged. All three use the same prior
aggregation, client split, rollout budget, and local PPO optimizer.

\begin{table*}[t]
\centering
\small
\setlength{\tabcolsep}{2pt}
\renewcommand{\arraystretch}{1.08}
\begin{tabular}{p{0.14\linewidth}p{0.22\linewidth}p{0.15\linewidth}p{0.20\linewidth}p{0.20\linewidth}}
\toprule
Method & Prior aggregation & Value baseline & Policy aggregation & Key coefficients \\
\midrule
FedGuide & OT-MoE on $\{\psi_i\}_{i=1}^{N}$ & $\beta=0.5$ & none & $\lambda_1=0.05$, $\lambda_2=0.5$, $\eta=0.05$ \\
FedGuide-P & OT-MoE on $\{\psi_i\}_{i=1}^{N}$ & $\beta=1$ & none & $\lambda_1=0.05$, $\lambda_2=0.5$, $\eta=0.05$ \\
FedGuide-A & OT-MoE on $\{\psi_i\}_{i=1}^{N}$ & $\beta=0.5$ & every 5 rounds & $\lambda_1=0.05$, $\lambda_2=0.5$, $\eta=0.05$ \\
\bottomrule
\end{tabular}
\caption{FedGuide variant hyperparameters.}
\label{tab:app_variant_hparams}
\end{table*}

\subsection{Baseline Hyperparameters}
\label{app:baseline_hparams}

The baseline implementations follow the communication mechanisms summarized in
Tables~\ref{tab:method_attribute} and~\ref{tab:theory_mechanism}. FedAvg and
FedKL use the same PPO backbone, FedSVRPG-M uses the PPO/SVRPG client-gradient
path, and FedRL uses a DDPGAvg-style actor-critic backbone.

\begin{table*}[t]
\centering
\small
\setlength{\tabcolsep}{3pt}
\renewcommand{\arraystretch}{1.08}
\begin{tabular}{p{0.24\linewidth}p{0.10\linewidth}p{0.13\linewidth}p{0.14\linewidth}p{0.14\linewidth}p{0.14\linewidth}}
\toprule
Parameter & Symbol & FedAvg & FedKL & FedRL & FedSVRPG-M \\
\midrule
Local optimizer & -- & PPO & PPO & DDPG & PPO/SVRPG \\
Aggregated parameters & -- & $\theta_i$ & $\theta_i$ & $\theta_i^{\mu},\theta_i^{Q}$ & $\Delta\theta_i$ \\
Global KL weight & $\lambda_{\mathrm{g}}$ & $0$ & $0.1$ & -- & -- \\
Local KL weight & $\lambda_{\mathrm{l}}$ & $0$ & $0.05$ & -- & -- \\
Local PPO epochs & $U$ & $10$ & $10$ & -- & $4$ \\
Mini-batch size & $B$ & $64$ & $64$ & $64$ & $64$ \\
Discount factor & $\gamma$ & $0.99$ & $0.99$ & $0.99$ & $0.99$ \\
GAE coefficient & $\lambda_{\mathrm{GAE}}$ & $0.95$ & $0.95$ & -- & $0.95$ \\
PPO clipping radius & $\epsilon$ & $\{0.1,0.2\}$ & $\{0.1,0.2\}$ & -- & $\{0.1,0.2\}$ \\
Value loss weight & $c_V$ & $0.5$ & $0.5$ & -- & $0.5$ \\
Gradient clipping & $c_{\mathrm{grad}}$ & $0.5$ & $0.5$ & -- & $0.5$ \\
Learning rate & $\eta_{\mathrm{opt}}$ & $10^{-4}$ & $10^{-4}$ & $10^{-4}$ & $10^{-4}$ \\
Reference update freq. & $K_{\mathrm{ref}}$ & -- & -- & -- & $5$ \\
Momentum weight & $\beta_{\mathrm{mom}}$ & -- & -- & -- & $\{0.2,0.5\}$ \\
BC initialization weight & $\alpha_{\mathrm{BC}}$ & $0.3$ & $0.3$ & $0.3$ & $0.3$ \\
Replay capacity & $R$ & -- & -- & $10^5$ & -- \\
Replay warm-up & $R_0$ & -- & -- & $10^3$ & -- \\
Target update coefficient & $\tau_{\mathrm{DDPG}}$ & -- & -- & $0.001$ & -- \\
Exploration noise & $(\mu_{\mathrm{OU}},\theta_{\mathrm{OU}},\sigma_{\mathrm{OU}})$ & -- & -- & $(0,0.15,0.2)$ & -- \\
\bottomrule
\end{tabular}
\caption{Baseline hyperparameters.}
\label{tab:app_baseline_hparams}
\end{table*}

\section{Additional Analysis}
\label{app:additional}

\subsection{Sensitivity to $M$, $\eta$, and $\lambda_2$}
\label{app:sensitivity}

We use Bandit2D, where the number of behavior modes is known to be four, to isolate the
role of the three quantities that govern OT-MoE aggregation and prior guidance: the expert
count $M$, the Sinkhorn temperature $\eta$, and the prior weight $\lambda_2$. All runs use
the four-client setting of Appendix~\ref{app:heterogeneous_setup} and five seeds.

\begin{figure}[htb]
  \centering
  \begin{minipage}[b]{0.32\linewidth}
    \centering
    \includegraphics[width=\linewidth,trim=0 0 0 22,clip]{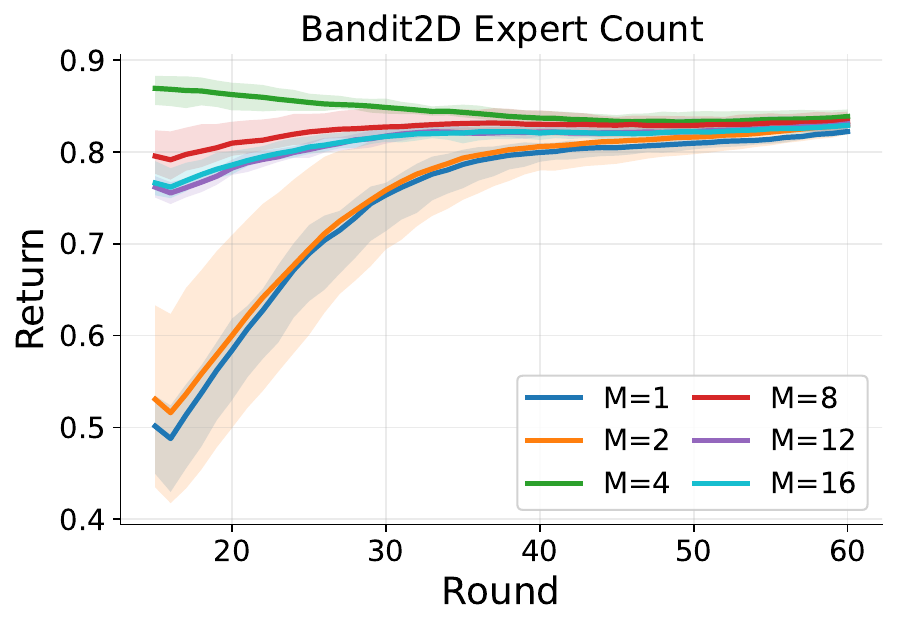}
    \\{\scriptsize (a) Expert count $M$}
  \end{minipage}\hfill%
  \begin{minipage}[b]{0.32\linewidth}
    \centering
    \includegraphics[width=\linewidth,trim=0 0 0 22,clip]{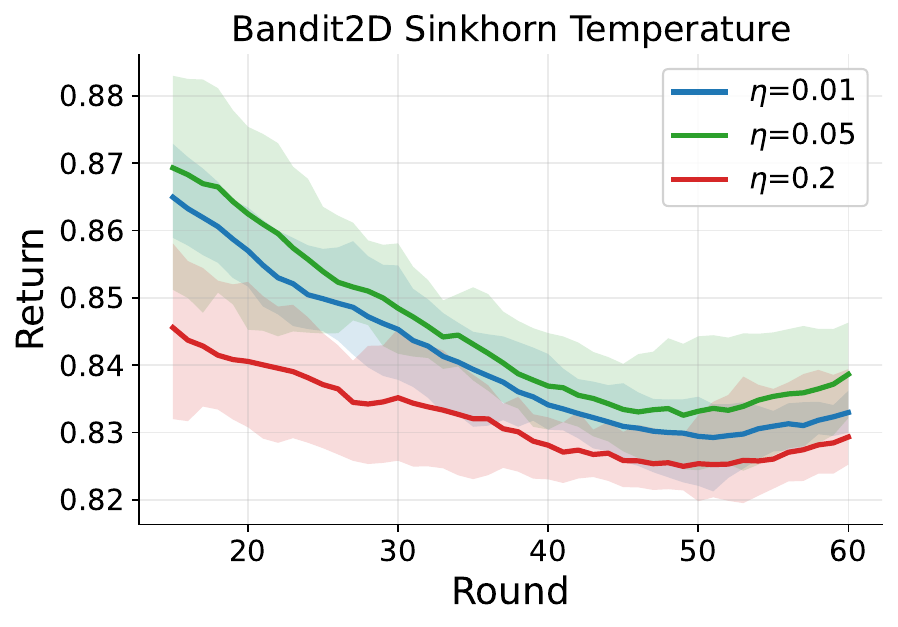}
    \\{\scriptsize (b) Sinkhorn $\eta$}
  \end{minipage}\hfill%
  \begin{minipage}[b]{0.32\linewidth}
    \centering
    \includegraphics[width=\linewidth,trim=0 0 0 22,clip]{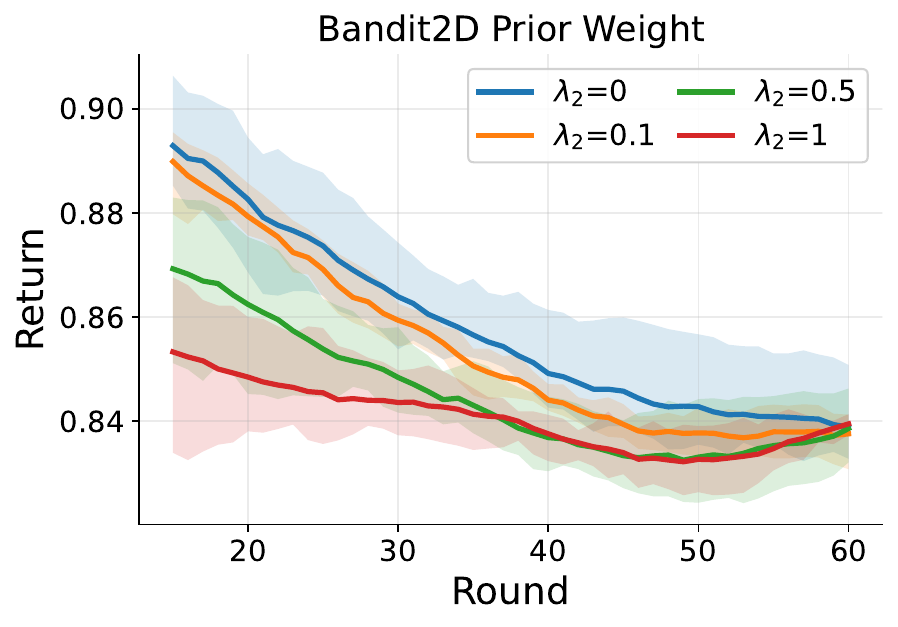}
    \\{\scriptsize (c) Prior weight $\lambda_2$}
  \end{minipage}
  \caption{Bandit2D sensitivity to the expert count $M$, the Sinkhorn temperature $\eta$,
  and the prior weight $\lambda_2$.}
  \label{fig:app_sensitivity}
\end{figure}

\paragraph{Expert count $M$.}
With four separated behavior modes, $M=1$ and $M=2$ under-resolve the support and adapt
substantially more slowly, the mode-matched $M=4$ is the most sample efficient, and larger
$M$ remains stable but redundant (Fig.~\ref{fig:app_sensitivity}a). The practical
implication is that $M$ should track the effective number of behavior modes rather than the
client count $N$; the default $M=N$ used elsewhere in this paper is therefore a conservative
choice that over-parameterizes the expert bank rather than under-resolving it. Estimating
the number of modes online, and adapting $M$ accordingly, is left to future work.

\paragraph{Sinkhorn temperature $\eta$.}
Varying $\eta$ over the tested range changes routing softness, moving the personalized prior
between nearly hard assignment and a broader mixture, but does not change the ordering of
methods (Fig.~\ref{fig:app_sensitivity}b).

\paragraph{Prior weight $\lambda_2$.}
Setting $\lambda_2=0$ removes prior guidance entirely and recovers purely local learning.
For positive values the curves converge (Fig.~\ref{fig:app_sensitivity}c), because the
implemented annealing reduces the guidance term over rounds and lets the locally rewarded
policy dominate later in training. Component roles are separately exposed by the variants:
FedGuide-P sets $\beta=1$ and removes the DICE baseline, while FedGuide-A adds policy
averaging and therefore removes fully local policy preservation. Behavior-cloning warm-up is
shared by all methods rather than specific to FedGuide, so it cannot account for the relative
gain.

\subsection{Behavior under Weak Overlap}
\label{app:weak_overlap}

We construct a controlled weak-overlap case to test what the personalized prior does when a
client's behavior support is not represented by any server expert. We add a fifth Bandit2D
client whose true target mode lies at the origin, equidistant from the four ring modes. Only
the four original clients update the experts; the fifth client participates in routing only,
since otherwise its mode would enter the expert bank and the out-of-distribution condition
would disappear.

\begin{figure}[htb]
  \centering
  \begin{minipage}[t]{0.62\linewidth}
    \centering
    \begin{minipage}[t]{0.48\linewidth}
      \centering
      \includegraphics[width=\linewidth]{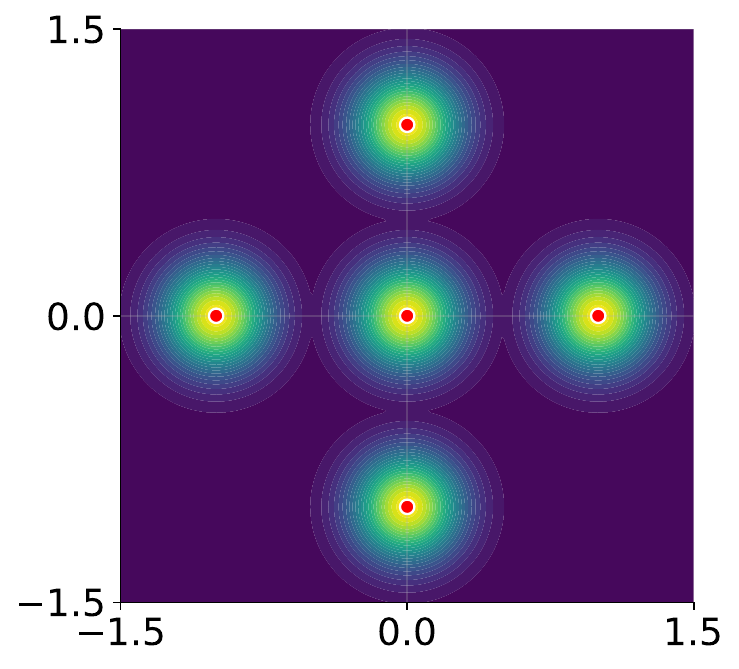}
    \end{minipage}\hfill%
    \begin{minipage}[t]{0.48\linewidth}
      \centering
      \includegraphics[width=\linewidth]{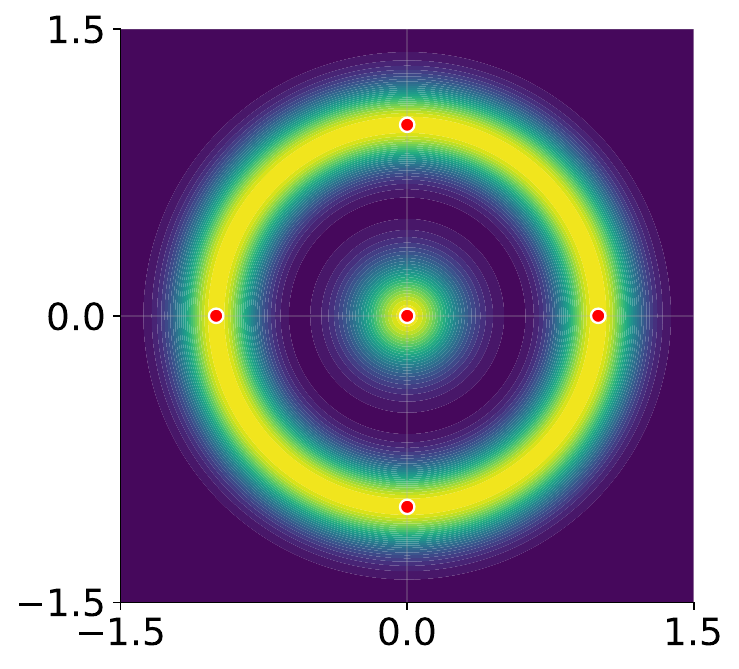}
    \end{minipage}\\
    {\scriptsize\textbf{Bandit2D Ground Truth}}
  \end{minipage}\hfill%
  \begin{minipage}[t]{0.35\linewidth}
    \centering
    \includegraphics[width=\linewidth]{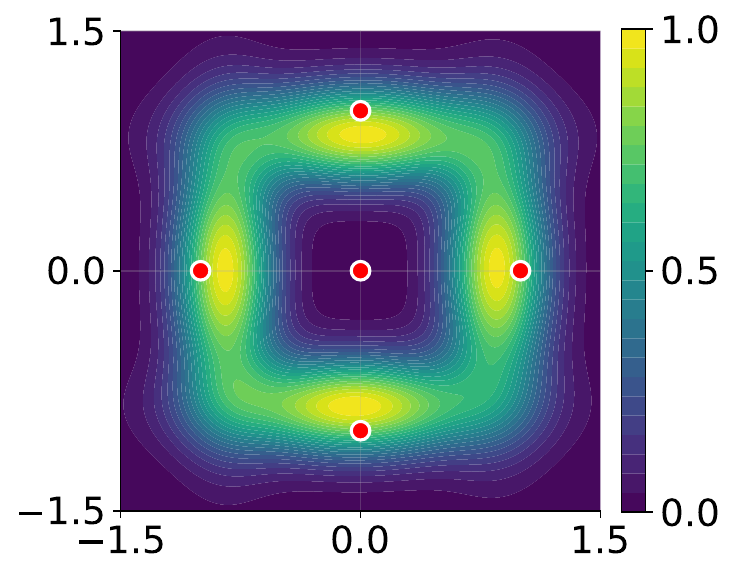}\\
    {\scriptsize\textbf{OT-MoE Global Prior}}
  \end{minipage}\\
  \includegraphics[width=\linewidth]{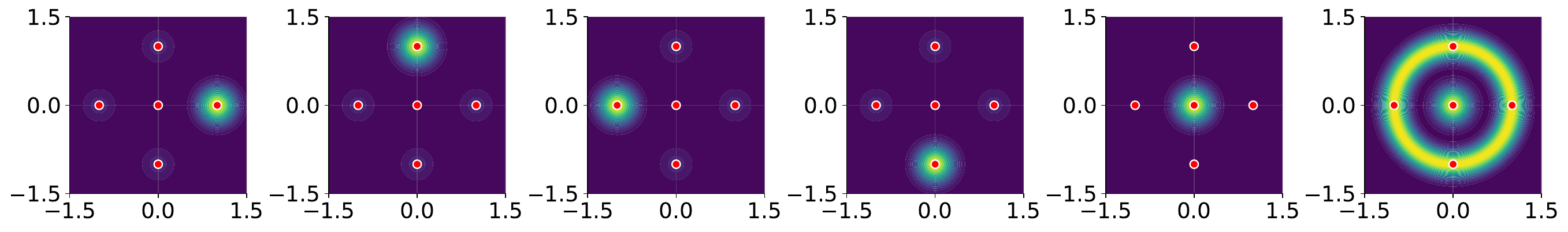}\\
  {\scriptsize
  \begin{tabular*}{\linewidth}{@{\extracolsep{\fill}}cccccc}
    Client 1 & Client 2 & Client 3 & Client 4 & Client 5 & Overall
  \end{tabular*}}
  \caption{Weak-overlap stress test. Client 5 has a true target mode at the origin, while the
  four experts contain only ring-shaped behavior support. OT-MoE routes Client 5 to a mixture
  of the existing experts, keeping its personalized prior on the ring rather than creating the
  unseen center mode.}
  \label{fig:app_weak_overlap}
\end{figure}

Because the fifth client is equidistant from all experts, the Sinkhorn plan assigns it a
near-uniform mixture, and its personalized prior remains supported on the ring instead of
forming a mode at the origin (Fig.~\ref{fig:app_weak_overlap}). This makes the operating
range of the method explicit: OT-MoE recombines behavior supports that the expert bank
already represents, and does not extrapolate a new mode outside them. A client whose support
no expert represents therefore receives guidance that is at best uninformative, which is the
mechanism behind the weaker gains on MetaWorld10 and the overlap assumption recorded in
Section~\ref{sec:limitations}.

\section{Visualizations}
\label{app:visualizations}
\begin{figure*}[tb!]
\centering
\includegraphics[width=\linewidth]{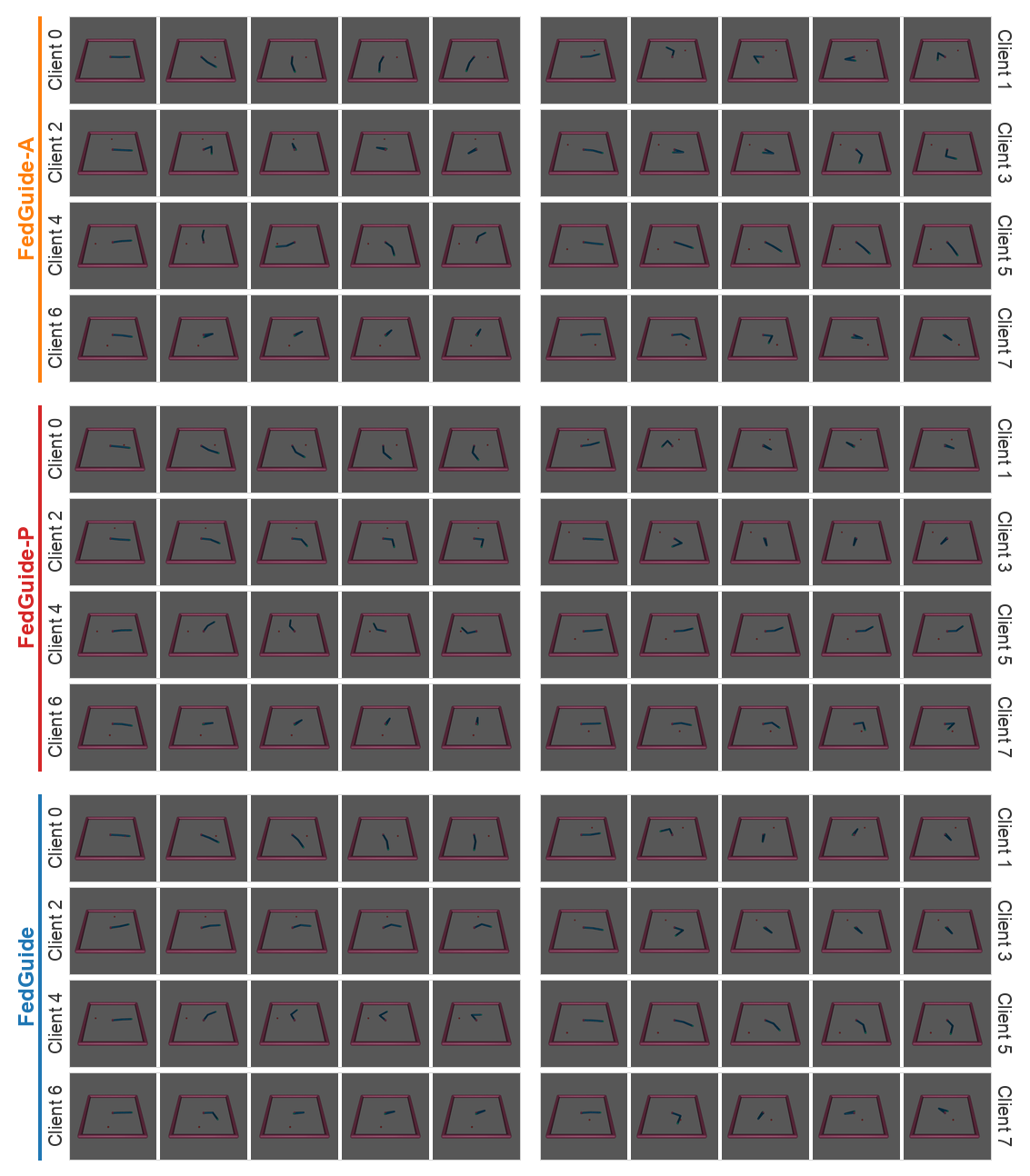}
\caption{FedGuide-family qualitative rollouts on Reacher, visualized with motion from a single final-round evaluation episode.}
\label{fig:app_video_reacher}
\end{figure*}

\begin{figure*}[tb!]
\centering
\includegraphics[width=\linewidth]{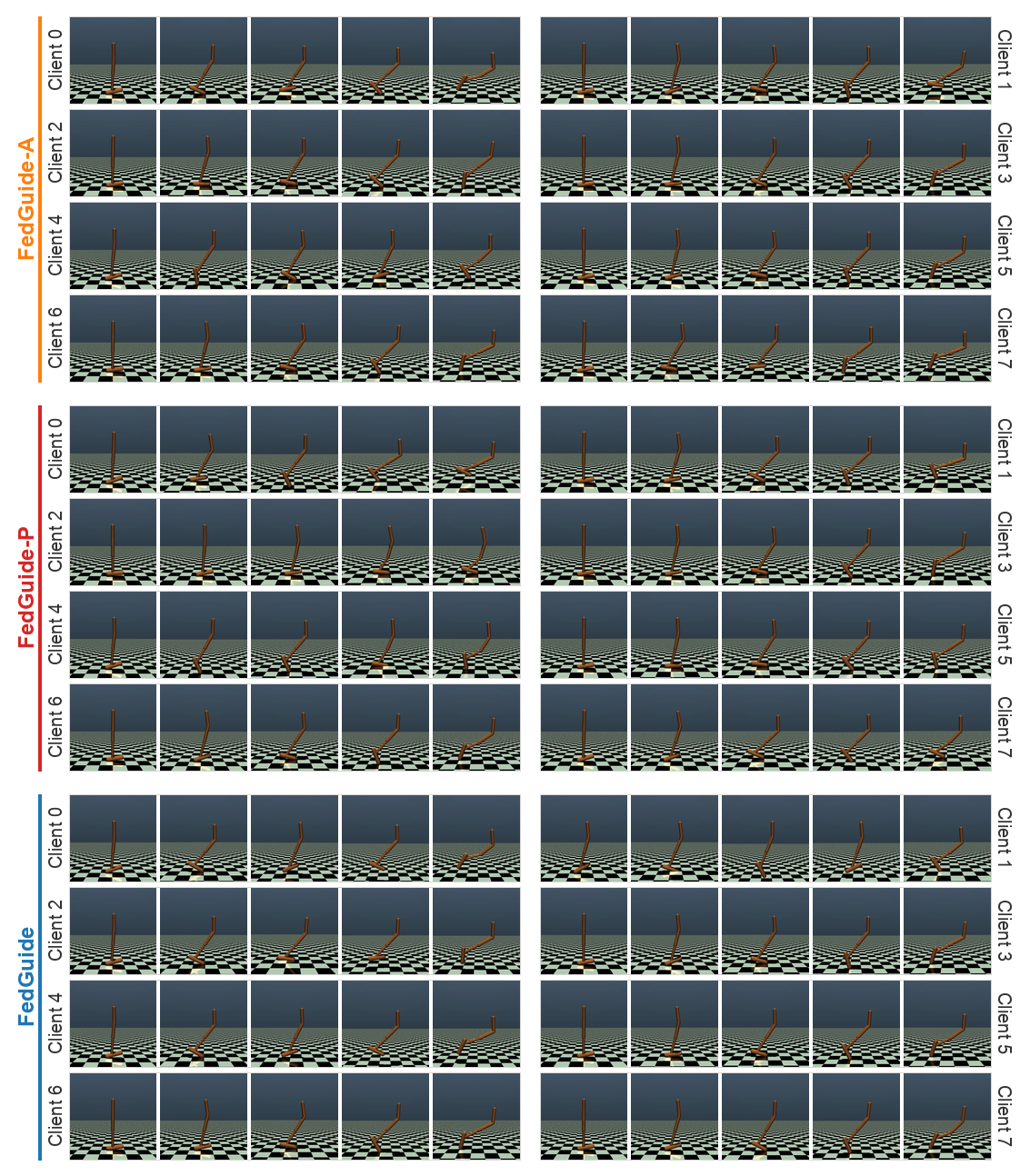}
\caption{FedGuide-family qualitative rollouts on Hopper. Each cell overlays
sampled frames from the final-round rollout to show the learned motion trace.}
\label{fig:app_video_hopper}
\end{figure*}

\begin{figure*}[tb!]
\centering
\includegraphics[width=\linewidth]{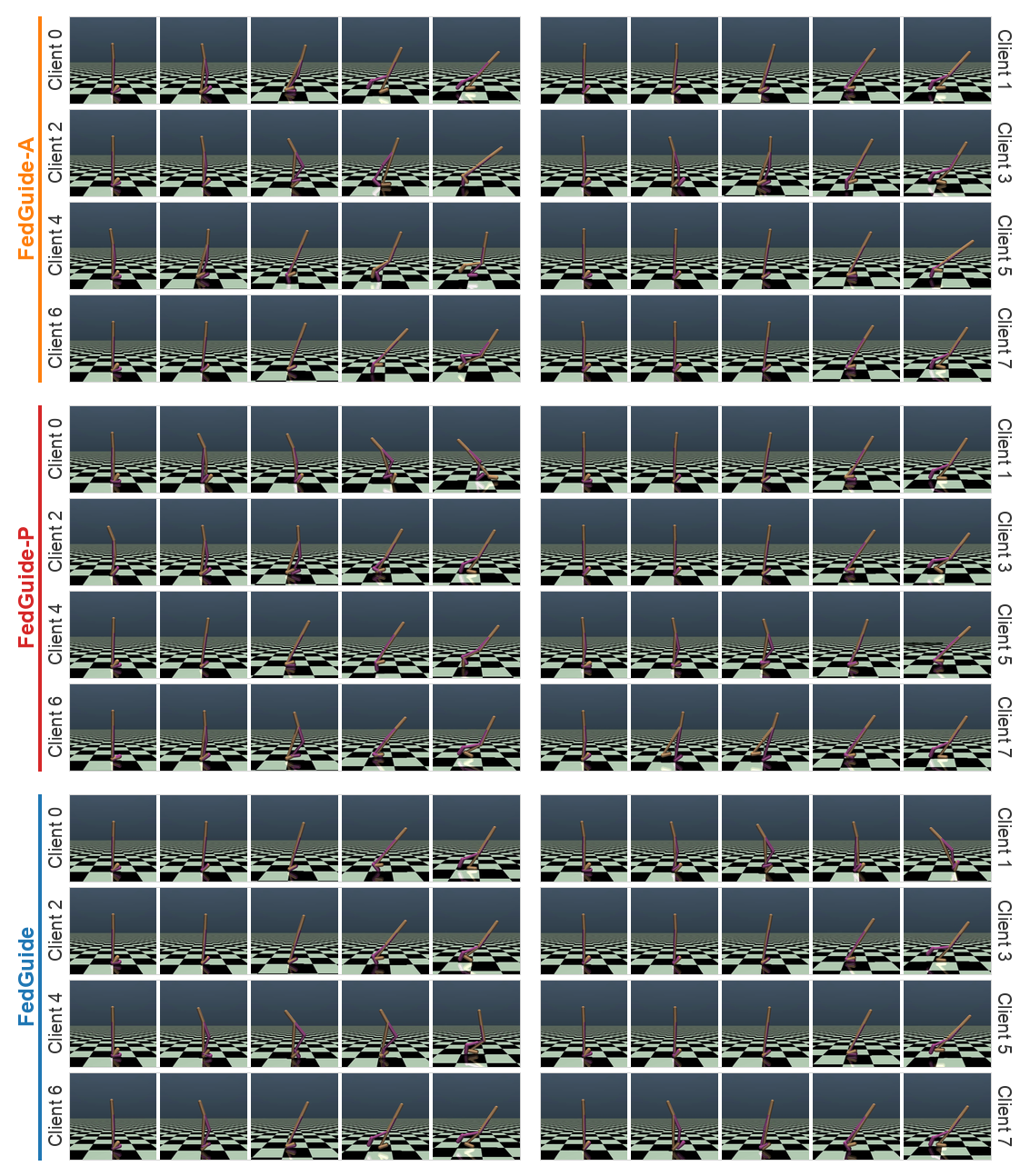}
\caption{FedGuide-family qualitative rollouts on Walker2D, visualized with
motion across heterogeneous clients.}
\label{fig:app_video_walker}
\end{figure*}

\begin{figure*}[tb!]
\centering
\includegraphics[width=\linewidth]{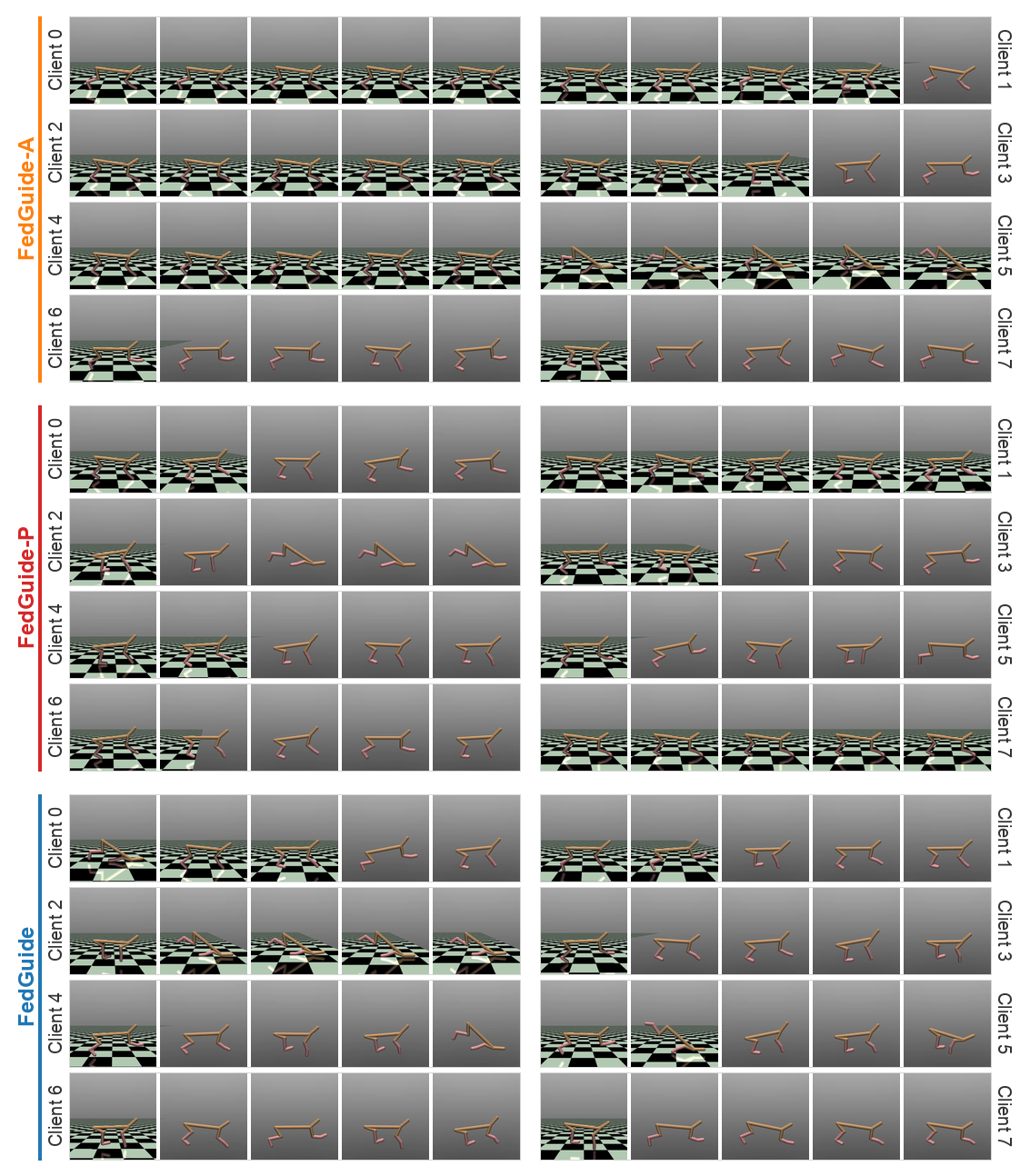}
\caption{FedGuide-family qualitative rollouts on HalfCheetah, visualized with
motion overlays across the first eight clients.}
\label{fig:app_video_halfcheetah}
\end{figure*}

\begin{figure*}[tb!]
\centering
\includegraphics[width=\linewidth]{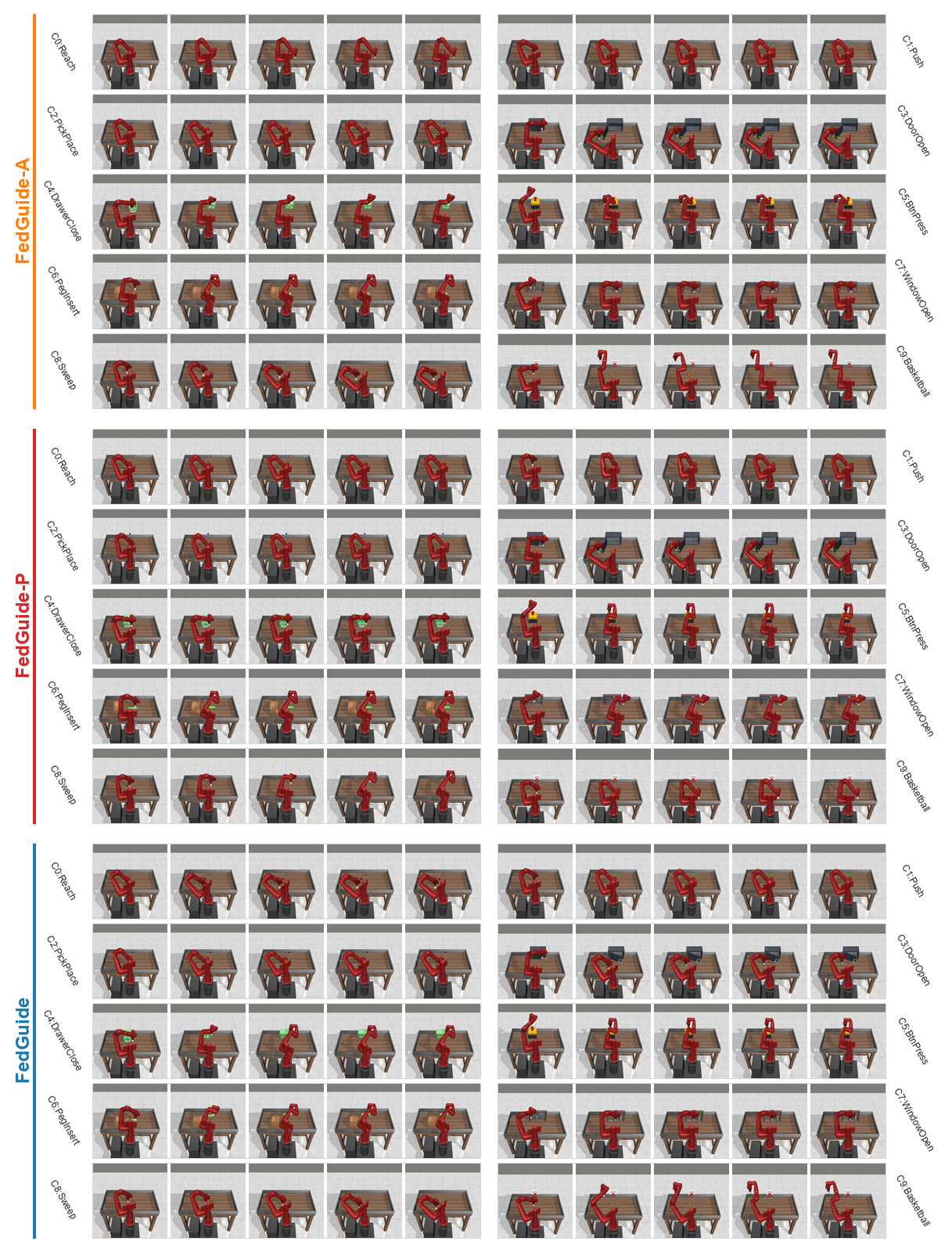}
\caption{FedGuide-family qualitative rollouts on MetaWorld10, visualized with across the ten task clients.}
\label{fig:app_video_metaworld}
\end{figure*}

We provide qualitative rollouts for the FedGuide family across heterogeneous
clients in Figures~\ref{fig:app_video_reacher}--\ref{fig:app_video_metaworld}.
Within each client block, rows follow FedGuide-A, FedGuide-P, and FedGuide. For
visualization only, each environment-method rollout is selected from the best
seed among five seeds according to final evaluation return; all
quantitative results in the paper use seed-averaged metrics. Each cell overlays
frames sampled from a single final-round evaluation episode, highlighting
reaching motion, gait, displacement, and object interaction without mixing
episode resets.

\end{document}